\documentclass{easychair}

\usepackage{doc}

\usepackage{pifont}
\usepackage{xcolor}
\definecolor{second}{HTML}{007000}

\usepackage{comment}

\usepackage{amssymb}
\usepackage{pifont}
\usepackage{booktabs}
\usepackage{longtable}
\usepackage{pgfplotstable}
\usepackage{titlesec}
\usepackage{multirow}
\usepackage{subcaption}
\usepackage{fancyvrb}
\usepackage{cleveref}
\usepackage{threeparttable}
\usepackage{wrapfig}
\usepackage{makecell}
\usepackage{hyperref}
\usepackage{tablefootnote}
\usepackage{float}

\pgfplotsset{compat=1.16}

\newcommand{\xmark}{\ding{55}}

\pgfkeysifdefined{/pgfplots/table/output empty row/.@cmd}{
    \pgfplotstableset{
        empty header/.style={
          every head row/.style={output empty row},
        }
    }
}{
    \pgfplotstableset{
        empty header/.style={
            typeset cell/.append code={%
                \ifnum\pgfplotstablerow=-1 %
                    \pgfkeyssetvalue{/pgfplots/table/@cell content}{}%
                \fi
            }
        }
    }
}

\titlespacing*{\paragraph}
{0pt}{2pt}{5pt}

\title{The 5th International Verification of Neural Networks Competition (VNN-COMP 2024): Summary and Results}%

\author{Christopher Brix\inst{1}
        \and Stanley Bak\inst{2}
        \and Taylor T. Johnson\inst{3}
        \and Haoze Wu \inst{4}
}

\institute{
    RWTH Aachen University, Aachen, Germany\\
    \email{brix@cs.rwth-aachen.de}
    \and
    Stony Brook University, Stony Brook, New York, USA\\
    \email{stanley.bak@stonybrook.edu}
    \and
    Vanderbilt University, Nashville, Tennessee, USA\\
    \email{taylor.johnson@vanderbilt.edu}
    \and
    Amherst College, Amherst, Massachussett, USA\\
    \email{hwu@amherst.edu}
 }

\authorrunning{C. Brix, S. Bak, T. Johnson, H. Wu}
\titlerunning{VNN-COMP 2024 Report}

\date{\phantom{date}}

\begin{document}

\maketitle

\begin{abstract}
This report summarizes the 5th International Verification of Neural Networks Competition (VNN-COMP 2024), held as a part of the The 7th International Symposium on AI Verification (SAIV), that was co-llocated with the 36th International Conference on Computer-Aided Verification (CAV). VNN-COMP is held annually to facilitate the fair and objective comparison of state-of-the-art neural network verification tools, encourage the standardization of tool interfaces, and bring together the neural network verification community. To this end, standardized formats for networks (ONNX) and specification (VNN-LIB) were defined, tools were evaluated on equal-cost hardware (using an automatic evaluation pipeline based on AWS instances), and tool parameters were chosen by the participants before the final test sets were made public. In the 2024 iteration, 8 teams participated on a diverse set of 12 regular and 8 extended benchmarks. This report summarizes the rules, benchmarks, participating tools, results, and lessons learned from this iteration of this competition.
\end{abstract}



\section{Introduction}
\label{sec:introduction}

Deep learning based systems are increasingly being deployed in a wide range of domains, including recommendation systems, computer vision, and autonomous driving. While the nominal performance of these methods has increased significantly over the last years, they largely lack formal guarantees on their behavior. 
However, in safety-critical applications, including autonomous systems, robotics, cybersecurity, and cyber-physical systems (CPS), such guarantees are essential for certification and reliability.

While the literature on the verification of traditionally designed systems is wide and successful, neural network verification remains an open problem, despite significant efforts over the last years. In 2020, the International Verification of Neural Networks Competition (VNN-COMP) was established to facilitate comparison between existing approaches, bring researchers working on this problem together, and help shape future directions of the field. VNN-COMP has been held annually since then~\cite{bak2021vnncomp,muller2022vnncomp,brix2023years,brix2023vnncomp}. In 2024, the 5th iteration of the annual VNN-COMP\footnote{\url{https://sites.google.com/view/vnn2024/home}} was held as a part of the 7th International Symposium on AI Verification (SAIV) that was collocated with the 36th International Conference on Computer-Aided Verification (CAV).

This 5th iteration of the VNN-COMP continues last year's trend of increasing standardization and automatization, aiming to enable a fair comparison between the participating tools and to simplify the evaluation of a large number of tools on a variety of (real-world) problems. 
As in the last iteration, VNN-COMP 2024 standardizes (1) neural network and specification formats, ONNX for neural networks and VNN-LIB \cite{vnnlib} for specifications, (2) evaluation hardware, providing participants the choice of a range of cost-equivalent AWS instances with different trade-offs between CPU and GPU performance, and (3) evaluation pipelines, enforcing a uniform interface for the installation and evaluation of tools. 

The competition was kicked off with the solicitation for participation in February 2024. Rule discussion started in April and rules are finalized in May 2024 (see an overview in \Cref{sec:rules}). From April to June 2024, benchmarks were proposed and discussed. Meanwhile, the organizing team decided to continue using AWS as the evaluation platform and started to implement an automated submission and testing system for both benchmarks and tools. By mid-July 2024, eight teams submitted their tools and the organizers evaluated all entrants to obtain the final results, discussed in \Cref{sec:results} and presented at SAIV on July 23, 2024. 
Discussions were structured into three issues on the official GitHub repository\footnote{\url{https://github.com/verivital/vnncomp2024/issues/}}: rules discussion, benchmarks discussion, and tool submission. All submitted benchmarks\footnote{\url{https://github.com/ChristopherBrix/vnncomp2024_benchmarks}} and final results\footnote{\url{https://github.com/ChristopherBrix/vnncomp2024_results}} were aggregated in separate GitHub repositories. 

The remainder of this report is organized as follows: \Cref{sec:rules} discusses the competition rules, \Cref{sec:participants} lists all participating tools, \Cref{sec:benchmarks} lists all benchmarks,  \Cref{sec:results} summarizes the results, and \Cref{sec:conclusion} concludes the report, discussing potential future improvements.

\newpage

\section{Rules}
\label{sec:rules}

\paragraph*{Terminology}
An \emph{instance} is defined by a property specification (pre- and post-condition), a network, and a timeout). 
For example, one instance might consist of an MNIST classifier with one input image, a given local robustness threshold $\epsilon$, and a specific timeout.
A \emph{benchmark} is defined as a set of related instances.
For example, one benchmark might consist of a specific MNIST classifier with 100 input images, potentially different robustness thresholds $\epsilon$, and one timeout per input.

\paragraph*{Run-time caps}
Run-times are capped on a per-instance basis, i.e., any verification instance will timeout (and be terminated) after at most X seconds, determined by the benchmark proposer. These can be different for each instance. 
The total per-benchmark runtime (sum of all per-instance timeouts) may not exceed 6 hours per benchmark. 
For example, a benchmark proposal could have six instances with a one-hour timeout, or 100 instances with a 3.6-minute timeout, each.
To enable a fair comparison, we measure the startup overhead for each tool by running it on a range of tiny networks and subtract the minimal overhead from the total runtime.

\paragraph*{Hardware}
To allow for comparability of results, all tools were evaluated on equal-cost hardware using  Amazon Web Services (AWS).
Each team could decide between a range of AWS instance types (see \Cref{tab:instances}) providing a CPU, GPU, or mixed focus.


\begin{table}[h]
\centering
\caption{Available AWS instances.}\label{tab:instances}
\renewcommand{\arraystretch}{1.1}
\scalebox{0.985}{
\begin{tabular}{lccc} \toprule
         &vCPUs & RAM [GB] & GPU \\ 
         \midrule
         p3.2xlarge  & 8 & 61 & V100 GPU with 16 GB memory \\
         m5.16xlarge  & 64 & 256 & \xmark \\
         g5.8xlarge  & 32 & 128 & A10G GPU with 24 GB memory \\
         \bottomrule
\end{tabular}
}
\end{table}

\paragraph{Scoring} 
The final score is aggregate as the sum of all benchmark scores.
Each benchmark score is the number of points (sum of instance scores discussed below) achieved by a given tool, normalized by the maximum number of points achieved by any tool on that benchmark. 
Thus, the tool with the highest sum of instance scores for a benchmark will get a benchmark score of 100, ensuring that all benchmarks are weighted equally, regardless of the number of constituting instances.

\paragraph*{Instance score}
\label{sec:scoring}
Each instance is scored is as follows: 
\begin{itemize}\setlength{\itemsep}{0pt}
    \item Correct hold (property proven): 10 points;
    \item Correct violated (counterexample found): 10 points;
    \item Incorrect result: -150 points (penalty increased compared to 2022);
    \item Timeout / Runtime Error / Unknown: 0 points.

\end{itemize}
However, the ground truth for any given instance is generally not known a priori. In the case of disagreement between tools, we, therefore, place the burden of proof on the tool claiming that a specification is violated, i.e. that a counterexample can be found, and deem it correct exactly if it produces a valid counterexample.

%
The provided counterexamples were supposed to define both the input and the resulting output of the networks.
However, for some tools and instances, the output definition was either missing or differed from the network output as computed by the onnxruntime package used to evaluate counterexamples (by performing inference given the inputs).
The competition rules were ambiguous how this would be handled. We decided to discard all outputs in the counterexample files and base the evaluation solely on the given inputs and their respective outputs as computed by the onnxruntime.
A ranking using the alternative evaluation, where incorrect or missing outputs result in a penalty can be found in Appendix~\ref{sec:alternative_ranking}.

\paragraph*{Time bonus}
As opposed to previous years, no time bonus was awarded.
Instead, all tools that are compute the correct result within the time limit receive the same amount of points.

\paragraph{Overhead Correction}
The overhead of tools was measured, but only used to adapt the timeouts. It did not influence the scores, as no time bonus was awarded.
To measure the tool-specific overhead, we created trivial network instances and included those in the measurements. We then observed the minimum verification time over all instances and considered that to be the overhead time for the tool.

\paragraph{Format}
As since 2021, we standardized neural networks to be in \texttt{onnx} format, specifications in \texttt{vnnlib} format, and counterexamples in a format similar to the \texttt{vnnlib} format.
Further, tool authors were required to provide scripts fully automating the installation process of their tool, including the acquisition of any licenses that might be needed. Similar to the previous year, a preparation and execution script had to be provided for running their tool on a specific instance consisting of a network file, specification file, and timeout.
The specifications are interpreted as definitions of counterexamples, meaning that a property is proven ``correct'' if the specification is shown to be unsatisfiable, conversely, the property is shown to be violated if a counterexample fulfilling the specification is found. 
Specifications consisted of disjunctions over conjunctions in both pre- and post-conditions, allowing a wide range of properties from adversarial robustness over multiple hyper-boxes to safety constraints to be encoded.
For example, robustness with respect to inputs in a hyper-box had to be encoded as disjunctive property, where any of the other classes is predicted.

\paragraph{Tracks}
For the first time, the competition was split into two tracks, both of them scored: A "regular" track, and an "extended" track.
The regular track consists of benchmarks selected by the tool participants based on a voting process, where a benchmark is included in the regular track if at least 50\% of tool participants voted for it to be scored. Benchmarks with at least 1 vote and not included in the regular track were scored in the extended track.
All benchmarks received at least one vote, so no benchmark was unscored.

\newpage

\section{Participants}
\label{sec:participants}
We list the tools and teams that participated in the VNN-COMP 2024 in \Cref{tab:tools} and reproduce their own descriptions of their tools below.



\begin{table*}[h]
\begin{center}
\begin{minipage}{\linewidth}
\caption{Summary of the key features of participating tools. The hardware column describes the used AWS instance with \texttt{p3} and \texttt{g5} making GPUs available, see \Cref{tab:instances} for more details. Licenses refer to the external licenses required to use the corresponding tool, not the licensing of the tool itself.} 
\label{tab:tools}%
\renewcommand{\arraystretch}{1.25}
\resizebox{\textwidth}{!}{ 
\begin{tabular}{lcm{5.0cm}ccc} \toprule
         Tool & References & Organizations & Place & Hardware & Licenses\\
         \midrule
         $\alpha$,$\beta$-CROWN & \cite{xu2020automatic,xu2021fast,wang2021betacrown,zhang2022general,shi2024genbab} & UIUC, UCLA, Drexel, Duke, RWTH Aachen & \textbf{1} & g5 & GUROBI\\
%
         CORA &  \cite{cora-website,kochdumper_et_al_2023,ladner_althoff_2023,koller_et_al_2024} & Technical University of Munich & 7 & g5 & MATLAB \\
         
%
         Marabou & \cite{katz2019marabou,wu2024marabou} & Hebrew University of Jerusalem, Stanford University, NRI Secure & 3 & m5 & GUROBI\\

         NeVer2 & \cite{demarchi2024never2, guidotti2021pynever} & Universit\`a degli Studi di Genova, Universit\`a degli Studi di Sassari, University of Kent & 6 & g5 & - \\
         NeuralSAT & \cite{duong2023dpllt, duong2024harnessing} & George Mason University & 8 & g5 & GUROBI\\
         nnenum & \cite{bak2020cav,bak2021nnenum} & Stony Brook University & 4 & m5 & - \\
         NNV & \cite{tran2020cav_tool,manzanas2023cav} & Vanderbilt University & 5 & m5 & MATLAB\\
         PyRAT &  \cite{pyrat2024, pyrat-website} & Universite Paris-Saclay, CEA, List & 2 & m5 & - \\

         \bottomrule
\end{tabular}
}
\end{minipage}
\end{center}
\end{table*}


\subsection{$\alpha,\!\beta$-CROWN}

\paragraph*{Team} Co-leaders: Huan Zhang (UIUC) and Zhouxing Shi (UCLA); 
Team members:  
Duo Zhou (UIUC),
Jorge Chavez (UIUC),
Xiangru Zhong (UIUC),
Hongji Xu (Duke, working as an intern supervised by Prof. Huan Zhang at UIUC),
Kaidi Xu (Drexel),
Hao Chen (UIUC)

The team \textbf{acknowledges} (ordered by last names) Christopher Brix (RWTH Aachen), Sanil Chawla (UIUC), Qirui Jin (University of Michigan), Suhas Kotha (CMU), and Zhuolin Yang (UIUC) who were involved into the development of the verifier during 2023 - 2024 but did not directly work on any benchmarks of the competition.

\paragraph*{Description} 
$\alpha,\!\beta$-CROWN (\texttt{alpha-beta-CROWN}) is an efficient neural network verifier based on the linear bound propagation framework and built on a series of works on bound-propagation-based neural network verifiers:  CROWN~\cite{zhang2018efficient}, auto\_LiRPA~\cite{xu2020automatic}, $\alpha$-CROWN~\cite{xu2021fast}, $\beta$-CROWN~\cite{wang2021betacrown}, GCP-CROWN~\cite{zhang2022general}, GenBaB~\cite{shi2024genbab}, BICCOS~\cite{zhou2024scalable}.
The core techniques in $\alpha,\!\beta$-CROWN combine the efficient and GPU-accelerated linear bound propagation method with branch-and-bound methods specialized for neural network verification. 

The linear bound propagation algorithms in $\alpha,\!\beta$-CROWN are based on our \texttt{auto\_LiRPA} library~\cite{xu2020automatic}, which supports general neural network architectures (including convolutional layers, pooling layers, residual connections, recurrent neural networks, and Transformers) and a wide range of nonlinear functions (e.g., ReLU, tanh, trigonometric functions, sigmoid, max pooling and average pooling), and is efficiently implemented on GPUs with Pytorch and CUDA. We jointly optimize intermediate layer bounds and final layer bounds using gradient ascent (referred to as $\alpha$-CROWN or optimized CROWN/LiRPA~\cite{xu2021fast}). Most importantly, we use branch and bound~\cite{bunelunified2018} (BaB) and incorporate split constraints in BaB into the bound propagation procedure efficiently via the $\beta$-CROWN algorithm~\cite{wang2021betacrown}, use cutting-plane method in GCP-CROWN~\cite{zhang2022general} and BICCOS~\cite{zhou2024scalable} to further tighten the bound, and support general nonlinearities in the branch-and-bound by GenBaB~\cite{shi2024genbab}.
For smaller networks, we also use a mixed integer programming (MIP) formulation~\cite{Tjeng2019EvaluatingRO} combined with tight intermediate layer bounds from $\alpha$-CROWN (referred to as $\alpha$-CROWN + MIP~\cite{zhang2022general}). The combination of efficient, optimizable and GPU-accelerated bound propagation with BaB produces a powerful and scalable neural network verifier.

New in this year, we have:
improved branch-and-bound for general nonlinear functions by GenBaB~\cite{shi2024genbab} which leverages the more flexible nature of general nonlinearities to make smarter branching decisions;
improved algorithms for input-space branch-and-bound;
developed new Branch-and-bound Inferred Cuts with COnstraint Strengthening (BICCOS)~\cite{zhou2024scalable}, a cutting plane (cut) approach that leverages inferred cuts from verified subproblems during branch-and-bound framework to achieve scalability for large networks without dependence on external MIP solvers;
and introduced new strategies for better time and GPU utilization.
Multiple papers are in progress or in submission.

\paragraph*{Link} \url{https://github.com/Verified-Intelligence/alpha-beta-CROWN} (main version)
\paragraph*{Competition submission} \url{https://github.com/Verified-Intelligence/alpha-beta-CROWN_vnncomp2024} (only for reproducing competition results; please use the main version for other purposes)
\paragraph*{Hardware and licenses} CPU and GPU with 32-bit or 64-bit floating point; Gurobi license required for certain benchmarks.
\paragraph*{Participated benchmarks} All benchmarks.

\subsection{CORA}
\paragraph*{Team} Lukas Koller, Tobias Ladner, Matthias Althoff (Technical University of Munich)
\paragraph*{Description} 
CORA~\cite{althoff_2015} enables the formal verification of neural networks, both in open-loop as well as in closed-loop scenarios. 
Open-loop verification refers to the task where properties of the output set of a neural network are verified, 
e.g. correctly classified images given noisy input, as also considered at VNN-COMP. 
In closed-loop scenarios, the neural network is used as a controller of a dynamic system, e.g., controlling a car while keeping a safe distance over some time horizon.

This is realized using reachability analysis, mainly using polynomial zonotopes~\cite{kochdumper_et_al_2023,ladner_althoff_2023}, 
allowing a non-convex enclosure of the output set of a neural network.
Moreover, CORA can also train robust neural networks~\cite{koller_et_al_2024}, 
which requires an efficient batch-wise propagation of zonotopes through a neural network on a GPU. 
This can also be used during verification by efficiently propagating the splitted sets batch-wise.

\paragraph*{Link} \url{https://github.com/kollerlukas/cora-vnncomp2024}

\paragraph*{Commit} 6f1923030baafadfadca3982b72fdea217a92479

\paragraph*{Hardware and licenses} GPU, MATLAB license.

\paragraph*{Participated Benchmarks} 
\texttt{acasxu}, \texttt{cifar100}, \texttt{collins-rul-cnn}, \texttt{cora}, \texttt{dist-shift}, \texttt{nn4sys}, \texttt{safenlp}, \texttt{tinyimagenet}, \texttt{tllverifybench}.

\subsection{Marabou}
\paragraph{Team} Haoze Wu (Stanford University), Clark Barrett (Stanford University), Guy Katz (Hebrew University of Jerusalem)
\paragraph{Description}  Marabou~\cite{katz2019marabou,wu2024marabou} is a user-friendly Neural Network Verification toolkit that can answer queries about a network’s properties by encoding and solving these queries as constraint satisfaction problems. It has both Python/C++ APIs through which users can load neural networks and define arbitrary linear properties over the neural network. Marabou supports many different linear, piecewise-linear, and non-linear~\cite{wu2022toward,wei2023convex} operations and architectures (e.g., FFNNs, CNNs, residual connections, Graph Neural Networks~\cite{vegas}). 

Under the hood, Marabou employs a uniform solving strategy for a given verification query. In particular, Marabou performs complete analysis that employs a specialized convex optimization procedure~\cite{wu2022efficient} and abstract interpretation~\cite{DeepPoly:19,vegas}. It also uses the Split-and-Conquer algorithm~\cite{wu2020parallelization} for parallelization.
\footnote{Thanks to the authors of the $\alpha-\beta$-CROWN team, an unsoundness issue of the competition version of Marabou on the \texttt{ViT} benchmarks was discovered. The networks in that benchmark contain bilinear and softmax connections. For this benchmark, the competition version of Marabou first performs DeepPoly-style abstract interpretation and then encodes the verification problem in the Gurobi optimizer. It turns out that Gurobi can report ``Infeasible'' on benchmarks where counter-examples are expected. The Marabou team is actively looking into resolving this issue. }

\paragraph{Link} \url{https://github.com/NeuralNetworkVerification/Marabou}
\paragraph*{Commit} 1a3ca6010b51bba792ef8ddd5e1ccf9119121bd8
\paragraph{Hardware and Licenses} CPU, no license required. Can also be accelerated with Gurobi (which requires a license)
\paragraph{Participated benchmarks} \texttt{acasxu}, \texttt{cgan}, \texttt{collins\_rul\_cnn}, \texttt{dist\_shift}, \texttt{linearizenn},
\texttt{metaroom},
\texttt{nn4sys}, 
\texttt{safenlp}, \texttt{tllverifybench}, \texttt{cifar100}, \texttt{tinyimagenet}.

\subsection{NeVer2}
\paragraph*{Team} Setefano Demarchi, Armando Tacchella (University of Genova), Elena Botoeva (University of Kent)

\paragraph*{Description}
\texttt{NeVer2}~\cite{demarchi2024never2} is an open-source, cross-platform tool aimed 
at designing, training, and verifying neural networks. 
It seamlessly integrates popular learning libraries with our verification backend, offering 
their functionalities also via a graphical interface.

\texttt{NeVer2} relies on the \texttt{pyNeVer}~\cite{guidotti2021pynever} Python API, which
provides the verification capability employing an abstraction-refinement algorithm, which
uses symbolic bounds propagation to compute stable and unstable neurons and an iterative 
refinement procedure to grow a search tree for proving the safety or the unsafety of a
verification query, which is expressed using star sets~\cite{tran2019fm}.

The algorithm propagates the abstraction provided by symbolic propagation and, if there
is an intersection with the unsafe post-conditions, checks whether there is a counter-example
to state the unsafety of the network, or to refine the abstraction branching on unstable
ReLU neurons. The next refinement target is decided based on the presence of unstable neurons
in early layers, or by the approximation area of the linear relaxation. 
In the form it is presented, this behaves as a complete algorithm. However, it can
be easily turned into an incomplete one by incorporating some early stopping
criteria, e.g., a timeout, or the maximum depth/number of refined neurons in a
branch. Currently, \texttt{NeVer2} supports only feed-forward architectures with
ReLU layers.

\paragraph*{Link}
\url{https://github.com/nevertools/pynever}

\paragraph*{Commit}
a7212f843a7e5137bdc54181b98b271f6b724747

\paragraph*{Hardware and licenses}
CPU, no license required.

\paragraph*{Participated benchmarks}
\texttt{acasxu}, \texttt{cora}, \texttt{safenlp}, \texttt{tllverifybench}.

\subsection{nnenum}
\paragraph*{Team} Ali Arjomandbigdeli (Student), Stanley Bak (Supervisor) (Stony Brook University)

\paragraph*{Description} 
The nnenum tool~\cite{bak2021nnenum} uses multiple levels of abstraction to achieve high-performance verification of ReLU networks without sacrificing completeness~\cite{bak2020vnn}. 
The core verification method is based on reachability analysis using star sets~\cite{tran2019fm}, combined with the ImageStar method~\cite{tran2020cav} to propagate stes through all linear layers supported by the ONNX runtime, such as convolutional layers with arbitrary parameters.  
The tool is written in Python 3 and uses GLPK for LP solving.
New this year, we added support for single lower and single upper bounds propagation in addition to zonotopes, similar to the DeepPoly method or the CROWN approach.
We also added an option to use Gurobi instead of GLPK for LP solving.

\paragraph*{Link}
\url{https://github.com/aliabigdeli/nnenum}

\paragraph*{Commit}
bcd65bcea050454b1a16a5fc5e8f94064af21085

\paragraph*{Hardware and licences}
CPU, Gurobi license (optional)

\paragraph*{Participated benchmarks}
\texttt{acasxu}, \texttt{cgan}, \texttt{collins-rul-cnn}, \texttt{cora}, \texttt{linearizenn}, \texttt{metaroom}, \texttt{safeNLP}, \texttt{nn4sys}, \texttt{tllverifybench}, \texttt{vggnet16}.

\subsection{NNV}
\paragraph*{Team} Diego Manzanas Lopez (Vanderbilt University), Samuel Sasaki (Vanderbilt University), Taylor T. Johnson (Vanderbilt University)
\paragraph*{Description} The Neural Network Verification (NNV) Tool~\cite{tran2020cav_tool,manzanas2023cav} is a formal verification software tool for deep learning models and cyber-physical systems with neural network components written in MATLAB and available at \url{https://github.com/verivital/nnv}. NNV uses a star-set state-space representation and reachability algorithm that allows for a layer-by-layer computation of exact or overapproximate reachable sets for feed-forward~\cite{tran2019fm}, convolutional~\cite{tran2020cav}, semantic segmentation (SSNN)~\cite{tran2021cav}, and recurrent (RNN)\cite{tran2023hscc} neural networks, as well as neural network control systems (NNCS)~\cite{tran2019emsoft,tran2020cav_tool} and neural ordinary differential equations (Neural ODEs)~\cite{manzanas2022formats}. 
The star-set based algorithm is naturally parallelizable, which allows NNV to be designed to perform efficiently on multi-core platforms. Additionally, if a particular safety property is violated, NNV can be used to construct and visualize the complete set of counterexample inputs for a neural network (exact-analysis). 
%
For this competition, updated from last year's, we tailor the solver approach depending on the benchmark at hand, although all follow a similar flow. First, we perform a simulation-guided search for counterexamples for a fixed number of samples. If no counterexamples are found (i.e., demonstrate that the property is SAT), then we utilize an iterative refinement approach using reachability analysis to verify the property (UNSAT). This consists of performing reachability analysis using a relax-approximation method~\cite{tran2021cav}, if not verified, then a less conservative approximation based on zonotope pre-filtering approach~\cite{tran2021fac}, and finally using the exact analysis when possible~\cite{tran2020cav} until the specification is verified or there is a timeout. Based on the benchmark to evaluate, the initial reachability analysis may be any of the overapproximation methods or the exact method, based on the complexity of the benchmarks (size of network, input, etc). 

\paragraph*{Link} \url{https://github.com/verivital/nnv}

\paragraph*{Commit} 50da012e7bf390788322329591e9edd3c45b4f0f

\paragraph*{Hardware and licenses} CPU, MATLAB license.

\paragraph*{Participated Benchmarks} 
Regular track except for LinearizeNN.

\subsection{NeuralSAT}
\paragraph*{Team} Hai Duong and Thanhvu Nguyen (George Mason).

\begin{wrapfigure}{r}{0.25\textwidth}
  \centering
  \vspace{-0.1in}
  \includegraphics[width=1\linewidth]{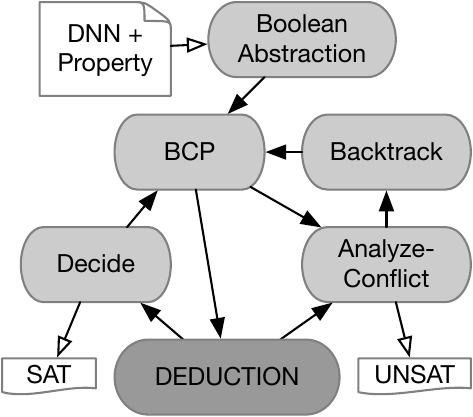}
  \vspace{-0.1in}  
\end{wrapfigure}
\paragraph*{Description} 
NeuralSAT~\cite{duong2023dpllt, duong2024harnessing} integrates conflict-driven clause learning (CDCL) in SAT/SMT-solving with an DNN abstraction-based theory solver for infeasibility checking.
The figure on the right gives an overview of NeuralSAT, which implements the DPLL(T) framework used in modern SMT solvers such as  Z3 and CVC. 
The design of NeuralSAT is inspired by the core algorithms used in SMT solvers such as CDCL components (light shades) and theory solving (dark shade). 
The tool is written in Python and uses Gurobi for LP solving. 
Unlike many other modern VNN verification tools,  NeuralSAT \emph{does not} require parameter tuning and works out of the box, e.g., the rool runs on the wide-range of benchmarks in VNN-COMPs without any tuning.

\emph{New:} In VNN-COMP'24, NeuralSAT has been improved with neuron stability, parallel beam search, and restarting strategies, which improved its performance significantly~\cite{duong2024harnessing}.

\paragraph*{Link}
\url{https://github.com/dynaroars/neuralsat}

\paragraph*{Commit}
5693c3da130942283744ce56c2e74ac6c16eef94

\paragraph*{Hardware and licences}
GPU, Gurobi License

\paragraph*{Participated benchmarks}
\texttt{acasxu}, 
\texttt{cgan}, 
\texttt{collins-rul-cnn}, 
\texttt{dist-shift}, 
\texttt{nn4sys}, 
\texttt{vggnet16},
\texttt{tllverifybench}, 
\texttt{trafic-signs-recognition},
\texttt{reach-prob-density},
\texttt{metaroom}.

\subsection{PyRAT}

\paragraph*{Team} Augustin Lemesle, Julien Lehmann, Tristan Le Gall (CEA-List)
\paragraph*{Description} 
PyRAT (Python Reachability Assessment Tool) \cite{pyrat2024} is an abstract interpretation based tool to verify the safety and robustness of neural networks. PyRAT implements several abstract domains such as Intervals, Zonotopes, Constrained Zonotopes, or Polyhedras to efficiently compute the reachable states for different architectures of neural networks such as dense, convolutional, residual or recurrent neural networks. It supports multiple non-linear activation functions (ReLU, Sigmoid, Softmax, Floor, ...) with precise abstractions while maintaining floating-point soudness. PyRAT is correct in that the output bounds reached will always be a sound over-approximation of the results and complete for ReLU based networks in that it will always give a true or false result given enough time.
Depending on the benchmark, different verification modes and domains can be selected. For smaller networks and problems, PyRAT can leverage branch and bound strategies on the inputs with heuristics like ReCIPH~\cite{durand2022reciph}. While on larger ReLU networks, PyRAT will use branch and bound strategies on the ReLU neurons in conjunction with fast GPU computation. 

\paragraph*{Link} \href{https://git.frama-c.com/pub/pyrat/}{https://git.frama-c.com/pub/pyrat}
\paragraph*{Commit} be7352ca5628669ae6a2ae5149c52a99fe86ed6a 
\paragraph*{Hardware and licenses} CPU and GPU, closed source CEA licence.
\paragraph*{Participated benchmarks} All benchmarks.

\newpage
\section{Benchmarks}
\label{sec:benchmarks}

In this section, we provide an overview of all scored benchmarks, reproducing the benchmark proposers' descriptions.
Artifacts for all benchmarks are available in the repository\footnote{\url{https://github.com/ChristopherBrix/vnncomp2024_benchmarks/tree/main/benchmarks}}.

               
\begin{table}[h]
    \centering
    \caption{Overview of all scored benchmarks. }
    \label{tab:my_label}
    \resizebox{\textwidth}{!}{
    \renewcommand{\arraystretch}{1.4}
    \begin{tabular}{ccccccc}
    \toprule
    Category &
    Benchmark &
    Application &
    Network Types &
    \# Params &
    Effective Input Dim &
    Track
    \\
    \midrule
    \multirow{8}{*}{Complex} 
    & cGAN & \makecell{Image Generation \\ \& Image Prediction} & Conv. + Vision Transformer & 500k - 68M & 5 & regular \\
    & NN4Sys & \makecell{Dataset Indexing \\ \&  Cardinality Prediction}   & ReLU + Sigmoid & 33k - 37M & 1-308 & regular \\
    & LinearizeNN & NN controller approximation & FC. + Conv. + Vision Transformer + Residual + ReLU & 203k & 4 & regular \\
    & ml4acopf & Power System & Complex (ReLU + Trigonometric + Sigmoid) & 4k-680k & 22 - 402 & extended \\
    & ViT & Vision & Conv. + Residual + Softmax + BatchNorm & 68k - 76k & 3072 & extended \\
    & Collins Aerospace & - & FC + Conv. + Residual, LeakyReLU + MaxPool + Square & 1.8M & 1.2M & extended \\
    & LSNC & Lyapunov stability of NN controllers & FC + Residual, ReLU + Sin + Cos & 210, 406 & 8 & extended \\
    & CCTSDB & - & FC + Conv. + Residual, ReLU +  MaxPool + Clip & 100k & 2 & extended \\
    \cmidrule(lr){1-7}
    \multirow{7}{*}{\makecell{CNN \\ \& ResNet}} 
    & Collins RUL CNN & Condition Based Maintenance & Conv. + ReLU, Dropout  & 60k - 262k   & 400 - 800 & regular \\
    & VGGNet16 & Image Classification & Conv. + ReLU + MaxPool    & 138M & 150k & extended \\
    & Traffic Signs Recognition & Image Classification & Conv. + Sign + MakPool + BatchNorm & 905k - 1.7M & 2.7k - 12k & extended \\
    & cifar100 & Image Classification & FC + Conv. + Residual, ReLU + BatchNorm & 2.5M - 3.8M & 3072 & regular \\
    & tinyimagenet & Image Classification & FC + Conv. + Residual, ReLU + BatchNorm & 3.6M & 9408 & regular \\
    & Metaroom & - & Conv. + FC, ReLU & 466k - 7.4M & 5376 & regular \\
    & Yolo & - & FC + Conv. + Residual, ReLU + Sigmoid & 22k - 37M & 1 - 308 & extended \\
    \cmidrule(lr){1-7} %
    \multirow{5.5}{*}{\makecell{FC}}
    & TLL Verify Bench & Two-Level Lattice NN & \makecell{Two-Level Lattice NN \\(FC. + ReLU)}  & 17k - 67M & 2 & regular \\
    & Acas XU & Collision Detection & FC. + ReLU & 13k & 5 & regular \\
    & Dist Shift & Distribution Shift Detection & FC. + ReLU + Sigmoid & 342k - 855k & 792 & regular \\
    & safeNLP & Sentence classification & FC. + ReLU & 4k & 30 & regular \\
    & CORA & Image Classification & FC. + ReLU & 575k, 1.1M & 784, 3072 & regular \\
    \bottomrule
    \end{tabular}
    }
\end{table}

\subsection{cGAN}
\paragraph*{Proposed by} Feiyang Cai, Ali Arjomandbigdeli, Stanley Bak (Stony Brook University)
\paragraph*{Motivation}
While existing neural network verification benchmarks focus on discriminative models, the exploration of practical and widely used generative networks remains neglected in terms of robustness assessment.
This benchmark introduces a set of image generation networks specifically designed for verifying the robustness of the generative networks.
\paragraph*{Networks}
The generative networks are trained using conditional generative adversarial networks (cGAN), whose objective is to generate camera images that contain a vehicle obstacle located at a specific distance in front of the ego vehicle, where the distance is controlled by the input distance condition.
The network to be verified is the concatenation of a generator and a discriminator.  The generator takes two inputs: 1) a distance condition (1D scalar) and 2) a noise vector controlling the environment (4D vector). The output of the generator is the generated image. The discriminator takes the generated image as input and outputs two values: 1) a real/fake score (1D scalar) and 2) a predicted distance (1D scalar).
Several different models with varying architectures (CNN and vision transformer) and image sizes (32x32, 64x64) are provided for different difficulty levels.
\paragraph*{Specifications}
The verification task is to check whether the generated image aligns with the input distance condition, or in other words, verify whether the input distance condition matches the predicted distance of the generated image.
In each specification, the inputs (condition distance and latent variables) are constrained in small ranges, and the output is the predicted distance with the same center as the condition distance but with slightly larger range.
\paragraph*{Link} \url{https://github.com/feiyang-cai/cgan_benchmark2023}

\pagebreak
\subsection{NN4Sys}
\paragraph*{Proposed by} the $\alpha,\!\beta$-CROWN team with collaborations with Cheng Tan, Haoyu He and Shuyi Lin at Northeastern University.
\paragraph*{Application}
The benchmark contains networks for database learned index, video streaming learned adaptive bitrate, and learned cardinality
estimation which map inputs from various dimensions to 1-dimension outputs.

\begin{itemize}

\item \textit{Background}: learned index, learned cardinality, and learned
    adaptive bitrate are all instances in neural networks for computer systems
        (NN4Sys), which are neural network based methods performing system
        operations. These classes of methods show great potential but have one
        drawback---the outputs of an NN4Sys model (a neural network) can be
        arbitrary, which may lead to unexpected issues in systems.

\item \textit{What to verify}: our benchmark provides multiple pairs of (1) trained NN4Sys model
and (2) corresponding specifications. We design these pairs with different parameters such
that they cover a variety of user needs and have varied difficulties for verifiers. 
We describe benchmark details in our NN4SysBench report:
        \url{http://naizhengtan.github.io/doc/papers/nn4sys23lin.pdf}.

\item \textit{Translating NN4Sys applications to a VNN benchmark}: 
the original NN4Sys applications have some sophisticated structures that are hard to verify.
We tailored the neural networks and their specifications to be suitable for VNN-COMP.
For example, learned index~\cite{kraska18case} contains multiple NNs in a tree structure that together serve one purpose.
However, this cascading structure is inconvenient/unsupported to verify
because there is a ``switch" operation---choosing one NN in the second stage
based on the prediction of the first stage's NN.
To convert learned indexes to a standard form, we train a monolithic (larger) NN.

\item \textit{A note on broader impact}: using NNs for systems is a broad topic, but many existing works
lack strict safety guarantees. We believe that NN Verification can help system developers gain confidence
to apply NNs to critical systems. We hope our benchmark can be an early step toward this vision.

\end{itemize}

\paragraph*{Networks}
This benchmark has twelve networks with different parameters: two for learned
indexes, four for learned cardinality estimation and six for learned adaptive bitrate.
The learned index uses fully-connected feed-forward neural networks. The other
two---the learned cardinality and the learned adaptive bitrate---has a
relatively sophisticated internal structure. Please see our NN4SysBench report
(URL listed above) for details

\paragraph*{Specifications}
For learned indexes,
the specification aims to check if the prediction error is bounded.
The specification is a collection of pairs of input and output intervals such that
any input in the input interval should be mapped to the corresponding output interval.
For learned cardinality estimation and learned adaptive bitrate,
the specifications check the prediction error bounds (similar to the learned indexes)
and monotonicity of the networks.
By monotonicity specifications, we mean that for two inputs, the network should produce a larger
output for the larger input, which is required by cardinality estimation or adaptive bitrate.

\paragraph{Link:} \url{https://github.com/Khoury-srg/VNNComp23_NN4Sys}

\subsection{LinearizeNN}
\paragraph*{Proposed by}  Ali Arjomandbigdeli, Stanley Bak (Stony Brook University).
\paragraph*{Motivation}
Assuming having a neural network controller approximation with a piecewise linear model in the form of a set of linear models with added noise to account for local linearization error. The objective of this benchmark is to investigate the neural network output falls within the range we obtain from our linear model output plus some uncertainty.

The idea of this benchmark came from one of our recent paper~\cite{ArjomandBigdeli2024} in which we approximated the NN controller with a piecewise linear model, and we wanted to check if the neural network output falls within the range we obtained from our linear model output plus some uncertainty.
\paragraph*{Networks} The neural network controller we used in this benchmark is an image-based controller for an Autonomous Aircraft Taxiing System whose goal is to control an aircraft's taxiing at a steady speed on a taxiway. This network was introduced  in the paper "Verification of Image-based Neural Network Controllers Using Generative Models"~\cite{katz2021veri}. The neural network integrates a concatenation of the cGAN (conditional GAN) and controller, resulting in a unified neural network controller with low-dimensional state inputs. In this problem, the inputs to the neural network consist of two state variables and two latent variables. The aircraft's state is determined by its crosstrack position (p) and heading angle error ($\theta$) with respect to the taxiway center line. Two latent variables with a range of -0.8 to 0.8 are introduced to account for environmental changes.

Because in this case the output spec depends on both the input and output and considering the VNN-LIB limitation, we added a skip-connection layer to the neural network to have the input values present in the output space. We also added one linear layer after that to create a linear equation for each local model.
\paragraph*{Specifications} As mentioned earlier, the aim of this benchmark is to examine whether the neural network output stays within the range defined by the linear model's output, including a margin for uncertainty.Given input $x \in X$ and output $Y = f_{NN}(x)$, the query is of the form: $A_{mat}\times X + b + U_{lb} \leq Y \leq A_{mat}\times X + b + U_{ub}$ for each linear model in its abstraction region.
\paragraph*{Link} \url{https://github.com/aliabigdeli/LinearizeNN_benchmark2024}

\pagebreak





\subsection{ml4acopf}
\paragraph*{Proposed by} Haoruo Zhao, Michael Klamkin, Mathieu Tanneau, Wenbo Chen, and Pascal Van Hentenryck (Georgia Institute of Technology), and Hassan Hijazi, Juston Moore, and Haydn Jones (Los Alamos National Laboratory).

\paragraph*{Motivation}
Machine learning models are utilized to predict solutions for an optimization model known as AC Optimal Power Flow (ACOPF) in the power system. Since the solutions are continuous, a regression model is employed. The objective is to evaluate the quality of these machine learning model predictions, specifically by determining whether they satisfy the constraints of the optimization model. Given the challenges in meeting some constraints, the goal is to verify whether the worst-case violations of these constraints are within an acceptable tolerance level.

\paragraph*{Networks}
The neural network designed comprises two components. The first component predicts the solutions of the optimization model, while the second evaluates the violation of each constraint that needs checking. The first component consists solely of general matrix multiplication (GEMM) and rectified linear unit (ReLU) operators. However, the second component has a more complex structure, as it involves evaluating the violation of AC constraints using nonlinear functions, including sigmoid, quadratic, and trigonometric functions such as sine and cosine. This complex evaluation component is incorporated into the network due to a limitation of the VNNLIB format, which does not support trigonometric functions. Therefore, these constraints violation evaluation are included in the neural network.

\paragraph*{Specifications}
In this benchmark, four different properties are checked, each corresponding to a type of constraint violation:
\begin{enumerate}
    \item Power balance constraints: the net power at each bus node is equal to the sum of the power flows in the branches connected to that node.
    \item Thermal limit constraints: power flow on a transmission line is within its maximum and minimum limits.
    \item Generation bounds: a generator's active and reactive power output is within its maximum and minimum limits.
    \item Voltage magnitude bounds: a voltage's magnitude output is within its maximum and minimum limits.
\end{enumerate}

The input to the model is the active and reactive load. The chosen input point for perturbation is a load profile for which a corresponding feasible solution to the ACOPF problem is known to exist. For the feasibility check, the input load undergoes perturbation. Although this perturbation does not exactly match physical laws, the objective is to ascertain whether a machine learning-predicted solution with the perturbation can produce a solution that does not significantly violate the constraints.

The scale of the perturbation and the violation threshold are altered by testing whether an adversarial example can be easily found using projected gradient descent with the given perturbation. The benchmark, provided with a fixed random seed, is robust against the simple projected gradient descent that is implemented.

\paragraph*{Link} \url{https://github.com/AI4OPT/ml4acopf_benchmark}

\subsection{ViT}
\paragraph*{Proposed by} the $\alpha,\!\beta$-CROWN team.
\paragraph*{Motivation}
Transformers~\cite{vaswani2017attention} based on the self-attention mechanism have much more complicated architectures and contain more kinds of nonlinerities, compared to simple feedforward networks with relatively simple activation functions. 
It makes verifying Transformers challenging. We aim to encourage the development of verification techniques for Transformer-based models, and we also aim to benchmark neural network verifiers on relatively complicated neural network architectures and more general nonlinearities. Therefore, we propose a new benchmark with Vision Transformers (ViTs)~\cite{dosovitskiy2020image}. This benchmark is developed based on our work on neural network verification for models with general nonlinearities~\cite{shi2024genbab}.

\paragraph*{Networks}
The benchmark contains two ViTs, as shown in \Cref{tab:vits}.
Considering the difficulty of verifying ViTs, we modify the ViTs and make the models relatively shallow and narrow, with significantly reduced number of layers and attention heads.
Following \cite{shi2019robustness}, we also replace the layer normalization with batch normalization.
The models are mainly trained with PGD training~\cite{madry2017towards}, and we also add a weighted IBP~\cite{gowal2018effectiveness,shi2021fast} loss for one of the models as a regularization.

\begin{table}[ht]
\centering
\caption{Networks in the ViT benchmark.}
\label{tab:vits}
\begin{tabular}{ccc}
\toprule 
Model & \texttt{PGD\_2\_3\_16} & \texttt{IBP\_3\_3\_8} \\
\midrule
Layers & 2 & 3\\
Attention heads & 3 & 3\\
Patch size & 16 & 8\\
Weight of IBP loss & 0 & 0.01\\
Training $\epsilon$ & $\frac{2}{255}$ & $\frac{1}{255}$\\
Clean accuracy & 59.78\% & 62.21\%\\
\bottomrule
\end{tabular}
\end{table}

\paragraph*{Specifications} 
The specifications are generated from the robustness verification problem with $\ell_\infty$ perturbation. 
We use the CIFAR-10 dataset with perturbation size $\epsilon=\frac{1}{255}$ at test time.
We have filtered the CIFAR-10 test set to exclude instances where either adversarial examples can be found (by PGD attack~\cite{madry2017towards} with 100 steps and 1000 restarts) or the vanilla CROWN-like method~\cite{zhang2018efficient,shi2019robustness} can already easily verify. 
We randomly keep 100 instances for each model, with a timeout threshold of 100 seconds. 
Note that since instances with adversarial examples have mostly been excluded during the filtering process, this version of the benchmark may not be able to reflect soundness issues in verifiers, and we refer readers to \cite{zhou2024testing} for discussions on testing soundness with models including ViT.

\paragraph*{Link} \url{https://github.com/shizhouxing/ViT_vnncomp2023}

\subsection{LSNC}

\paragraph*{Proposed by} the $\alpha,\!\beta$-CROWN team.
\paragraph*{Motivation}
We develop a benchmark for the problem of verifying the Lyapunov stability of NN controllers in nonlinear dynamical systems within a region-of-intrest and a region-of-attraction. This is important for providing stability guarantees that are essential for safety-critical applications with NN controllers. It is also a useful application of neural network verification as recently demonstrated in \cite{yang2024lyapunov,shi2024certified}, and we refer readers to those works for more details on the problem.
\paragraph*{Networks and Specifications}
Models are adopted from \cite{yang2024lyapunov}. 
We adopt two models for the 2D quadrotor dynamical system with state feedback and output feedback, respectively. Each model consists of a controller which is a shallow ReLU network, a Lyapunov function which is a quadratic function, and nonlinear operators modelling the dynamics of a 2D quadrotor. The model for output feedback further consists of a shallow LeakyReLU network as the observer. The verification objective of the Lyapunov stability has been encoded in the ONNX graphs and VNNLIB specifications. Specifications for the benchmark are randomly generated and consist of random sub-regions within the original region-of-interest. The size of the random sub-regions is controlled by a factor $\epsilon~(0<\epsilon\leq 1)$ which is applied to each input dimension, and it has been adjusted for a suitable difficulty given the timeout. For the state feedback model, we set $\epsilon=0.5$ and the timeout is 100s; for the output feedback model, we set $\epsilon=0.3$ and timeout is 200s. For each of the two models, we randomly generate 20 instances. 
\paragraph*{Link} \url{https://github.com/shizhouxing/LSNC_VNNCOMP2024}

\subsection{Collins-RUL-CNN}
\paragraph*{Proposed by} Collins Aerospace, Applied Research \& Technology (\href{https://www.collinsaerospace.com/what-we-do/capabilities/technology-and-innovation/applied-research-and-technology}{website}).

\paragraph*{Motivation} Machine Learning (ML) is a disruptive technology for the aviation industry. This particularly concerns safety-critical aircraft functions, where high-assurance design and verification methods have to be used in order to obtain approval from certification authorities for the new ML-based products. Assessment of correctness and robustness of trained models, such as neural networks, is a crucial step for demonstrating the absence of unintended functionalities~\cite{ForMuLA, kirov2023formal}. The key motivation for providing this benchmark is to strengthen the interaction between the VNN community and the aerospace industry by providing a realistic use case for neural networks in future avionics systems~\cite{kirov2023benchmark}.

\paragraph*{Application} Remaining Useful Life (RUL) is a widely used metric in Prognostics and Health Management (PHM) that manifests the remaining lifetime of a component (e.g., mechanical bearing, hydraulic pump, aircraft engine). RUL is used for Condition-Based Maintenance (CBM) to support aircraft maintenance and flight preparation. It contributes to such tasks as augmented manual inspection of components and scheduling of maintenance cycles for components, such as repair or replacement, thus moving from preventive maintenance to \emph{predictive} maintenance (do maintenance only when needed, based on component’s current condition and estimated future condition). This could allow to eliminate or extend service operations and inspection periods, optimize component servicing (e.g., lubricant replacement), generate inspection and maintenance schedules, and obtain significant cost savings. Finally, RUL function can also be used in airborne (in-flight) applications to dynamically inform pilots on the health state of aircraft components during flight. Multivariate time series data is often used as RUL function input, for example, measurements from a set of sensors monitoring the component state, taken at several subsequent time steps (within a time window). Additional inputs may include information about the current flight phase, mission, and environment. Such highly multi-dimensional input space motivates the use of Deep Learning (DL) solutions with their capabilities of performing automatic feature extraction from raw data.

\paragraph*{Networks} The benchmark includes 3 convolutional neural networks (CNNs) of different complexity: different numbers of filters and different sizes of the input space. All networks contain only convolutional and fully connected layers with ReLU activations. All CNNs perform the regression function. They have been trained on the same dataset (time series data for mechanical component degradation during flight).

\paragraph*{Specifications} We propose 3 properties for the NN-based RUL estimation function. First, two properties (robustness and monotonicity) are local, i.e., defined around a given point. We provide a script with an adjustable random seed that can generate these properties around input points randomly picked from a test dataset. For robustness properties, the input perturbation (delta) is varied between 5\% and 40\%, while the number of perturbed inputs varies between 2 and 16. For monotonicity properties, monotonic shifts between 5\% and 20\% from a given point are considered. Properties of the last type ("if-then") require the output (RUL) to be in an expected value range given certain input ranges. Several if-then properties of different complexity are provided (depending on range widths).

\paragraph*{Link} \url{https://github.com/loonwerks/vnncomp2022}

\paragraph*{Paper} Available in~\cite{kirov2023benchmark} or on request.

\subsection{VGGNET16}
\paragraph*{Proposed by} Stanley Bak, Stony Brook University

\paragraph*{Motivation} This benchmark tries to scale up the size of networks being analyzed by using the well-studied VGGNET-16 architecture~\cite{simonyan2014very} that runs on ImageNet. Input-output properties are proposed on pixel-level perturbations that can lead to image misclassification. 

\paragraph*{Networks} All properties are run on the same network, which includes 138 million parameters. The network features convolution layers, ReLU activation functions, as well as max pooling layers.

\paragraph*{Specifications} Properties analyzed ranged from single-pixel perturbations to perturbations on all 150528 pixles (L-infinity perturbations). A subset of the images was used to create the specifications, one from each category, which was randomly chosen to attack. Pixels to perturb were also randomly selected according to a random seed.

\paragraph*{Link} \url{https://github.com/stanleybak/vggnet16_benchmark2022/}

\subsection{Traffic Signs Recognition}
\paragraph*{Proposed by} M\u{a}d\u{a}lina Era\c{s}cu and Andreea Postovan (West University of Timisoara, Romania)
\paragraph*{Motivation} Traffic signs play a crucial role in ensuring road safety and managing traffic flow in both city and highway driving. The recognition of these signs, a vital component of autonomous driving vision systems, faces challenges such as susceptibility to adversarial examples~\cite{szegedy2013intriguing} and occlusions~\cite{zhang2020lightweight}, stemming from diverse traffic scene conditions.

\paragraph*{Networks} Binary neural networks (BNNs) show promise in computationally limited and energy-constrained environments within the realm of autonomous driving~\cite{hubara2016binarized}. BNNs, where weights and/or activations are binarized to $\pm 1$, offer reduced model size and simplified convolution operations for image recognition compared to traditional neural networks (NNs).

We trained and tested various BNN architectures using the German Traffic Sign Recognition Benchmark (GTSRB) dataset~\cite{GTSRB}. This multi-class dataset, containing images of German road signs across 43 classes, poses challenges for both humans and models due to factors like perspective change, shade, color degradation, and lighting conditions. The dataset was also tested using the Belgian Traffic Signs \cite{BelgianTrafficSignDatabase} and Chinese Traffic Signs \cite{ChineseTrafficSignDatabase} datasets. The Belgium Traffic Signs dataset, with 62 classes, had 23 overlapping classes with GTSRB. The Chinese Traffic Signs dataset, with 58 classes, shared 15 classes with GTSRB. Pre-processing steps involved relabeling classes in the Belgium and Chinese datasets to match those in GTSRB and eliminating non-overlapping classes (see \cite{postovan2023architecturing} for details).

\begin{figure}[h]
  \centering
    \includegraphics[width=0.7\textwidth]{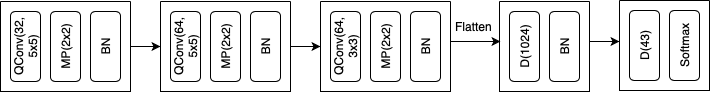}
    \caption{Accuracy Efficient Architecture for GTSRB and Belgium dataset}
    \label{fig:Acc-Efficient-Arch-GTSRB-Belgium}
\end{figure}

\begin{figure}[h]
  \centering
    \includegraphics[width=0.7\textwidth]{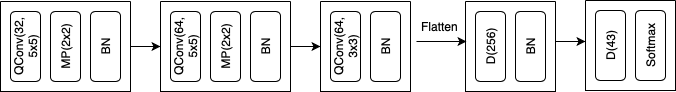}
    \caption{Accuracy Efficient Architecture for Chinese dataset}
    \label{fig:Acc-Efficient-Arch-Chinese}
\end{figure}

\begin{figure}[h]
  \centering
    \includegraphics[width=0.3\textwidth]{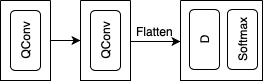}
    \caption{XNOR(QConv) architecture}
    \label{fig:XNOR(QConv)-arch}
\end{figure}

We provide three models with the structure in Figures \ref{fig:Acc-Efficient-Arch-GTSRB-Belgium}, \ref{fig:Acc-Efficient-Arch-Chinese}, and \ref{fig:XNOR(QConv)-arch}. They contain QConv, Batch Normalization (BN), Max Pooling (ML), Fully Connected/Dense (D) layers.  Note that the QConv layer binarizes the corresponding convolutional layer. All models were trained for 30 epochs. The model from Figure \ref{fig:Acc-Efficient-Arch-GTSRB-Belgium} was trained with images having the dimension 64px x 64 px, the one from Figure \ref{fig:Acc-Efficient-Arch-Chinese} with 48px x 48 px and the one from Figure \ref{fig:XNOR(QConv)-arch} with 30px x 30 px. The two models involving Batch Normalization layers introduce real valued parameters besides the binary ones, while the third one contains only binary parameters (see Table \ref{tab:stats}) for statistics.

\begin{table}[h]
\caption{Training and Testing Statistics}
\label{tab:stats}
\centering
\scriptsize
\begin{tabular}{|c|c|ccc|ccc|}
\hline
\multirow{2}{*}{\textbf{Input size}} & \multirow{2}{*}{\textbf{Model name}} & \multicolumn{3}{c|}{\textbf{Accuracy}}                                      & \multicolumn{3}{c|}{\textbf{\#Params}}                                      \\ \cline{3-8} 
                            &                             & \multicolumn{1}{c|}{\textbf{German}} & \multicolumn{1}{c|}{\textbf{China}} & \textbf{Belgium} & \multicolumn{1}{c|}{\textbf{Binary}}  & \multicolumn{1}{c|}{\textbf{Real}} & \textbf{Total}   \\ \hline
64px $\times$ 64px          & Figure \ref{fig:Acc-Efficient-Arch-GTSRB-Belgium}                  & \multicolumn{1}{c|}{96.45}  & \multicolumn{1}{c|}{81.50} & 88.17   & \multicolumn{1}{c|}{1772896} & \multicolumn{1}{c|}{2368} & 1775264 \\ \hline
48px $\times$ 48px          & Figure \ref{fig:Acc-Efficient-Arch-Chinese}                  & \multicolumn{1}{c|}{95.28}  & \multicolumn{1}{c|}{83.90} & 87.78   & \multicolumn{1}{c|}{904288}  & \multicolumn{1}{c|}{832}  & 905120  \\ \hline
30px $\times$ 30px          & Figure \ref{fig:XNOR(QConv)-arch}                  & \multicolumn{1}{c|}{81.54}  & \multicolumn{1}{c|}{N/A}   & N/A     & \multicolumn{1}{c|}{1005584} & \multicolumn{1}{c|}{0}    & 1005584 \\ \hline
\end{tabular}
\end{table}
\paragraph*{Specifications} To evaluate the \emph{adversarial robustness} of the networks above, we assessed perturbations within the infinity norm around zero, with the radius denoted as $\epsilon = \{1, 3, 5, 10, 15\}$. This involved randomly selecting three distinct images from the GTSRB dataset's test set for each model and generating \textsc{VNNLIB} files for each epsilon in the set. In total, we created 45 \textsc{VNNLIB} files. Due to a 6-hour total timeout constraint for solving all instances, each instance had a maximum timeout of 480 seconds. To review the generated \textsc{VNNLIB} specification files submitted to VNNCOMP 2023, as well as to generate new ones, please refer to \url{https://github.com/apostovan21/vnncomp2023}.

\paragraph*{Link} \url{https://github.com/apostovan21/vnncomp2023}

\subsection{CIFAR100}

\paragraph*{Proposed by} the $\alpha,\!\beta$-CROWN team.
\paragraph*{Motivation} This benchmark is reused from VNN-COMP 2022 with a reduced complexity (only two out of the four models with medium sizes are retained). 
See details in Section 4.5 of the report of VNN-COMP 2022~\cite{muller2022vnncomp}.

\paragraph*{Networks} We provide two ResNet models on CIFAR-100 with different model widths and depths (input dimension $32 \times 32 \times 3$, 100 classes):
\begin{itemize}
    \item \texttt{CIFAR100-ResNet-medium}: 8 residual blocks, 17 convolutional layers + 2 linear layers
    \item \texttt{CIFAR100-ResNet-large}: 8 residual blocks, 19 convolutional layers + 2 linear layers (almost identical to standard ResNet-18 architecture)
\end{itemize}

\paragraph*{Specifications} 
We randomly select 100 images from the CIFAR-100 test set with a verification timeout of 100 seconds for each of the two models. 
We filtered out the samples which can be verified by vanilla CROWN (which is used during training) to make the benchmark more challenging. The filtering process is done offline on a machine with a GPU due to the large sizes of these models. 
A small proportion of instances (around 18\%) with adversarial examples have been retained for potentially identifying unsound results. 

\paragraph*{Link} \url{https://github.com/huanzhang12/vnncomp2024_cifar100_benchmark}

\subsection{TinyImagenet}

\paragraph*{Proposed by} the $\alpha,\!\beta$-CROWN team.
\paragraph*{Motivation} This benchmark is reused from VNN-COMP 2022. See details in Section 4.5 of the report of VNN-COMP 2022~\cite{muller2022vnncomp}.

\paragraph*{Networks} We provide a ResNet for TinyImageNet (input dimension $64 \times 64 \times 3$, 200 classes):
\begin{itemize}
    \item \texttt{TinyImageNet-ResNet-medium}: 8 residual blocks, 17 convolutional layers + 2 linear layers
\end{itemize}

\paragraph*{Specifications} 
We randomly select 200 images from the TinyImageNet test set with a verification timeout of 100 seconds for each of the two models. A filtering procedure has been adopted similar to the CIFAR100 benchmark.

\paragraph*{Link} \url{https://github.com/huanzhang12/vnncomp2024_tinyimagenet_benchmark}

\subsection{TLL Verify Bench}
\paragraph*{Proposed by} James Ferlez (University of California, Irvine)

\paragraph*{Motivation} This benchmark consists of Two-Level Lattice (TLL) NNs, which have been shown to be amenable to fast verification algorithms (e.g. \cite{FerlezKS22}). Thus, this benchmark was proposed as a means of comparing TLL-specific verification algorithms with general-purpose NN verification algorithms (i.e. algorithms that can verify arbitrary deep, fully-connected ReLU NNs).

\paragraph*{Networks}  The networks in this benchmark are a subset of the ones used in \cite[Experiment 3]{FerlezKS22}. Each of these TLL NNs has $n=2$ inputs and $m=1$ output. The architecture of a TLL NN is further specified by two parameters: $N$, the number of local linear functions, and $M$, the number of selector sets. This benchmark contains TLLs of sizes $N = M = 8, 16, 24, 32, 40, 48, 56, 64$, with $30$ randomly generated examples of each (the generation procedure is described in \cite[Section 6.1.1]{FerlezKS22}). At runtime, the specified verification timeout determines how many of these networks are included in the benchmark so as to achieve an overall 6-hour run time; this selection process is deterministic. Finally, a TLL NN has a natural representation using multiple computation paths \cite[Figure 1]{FerlezKS22}, but many tools are only compatible with fully-connected networks. Hence, the ONNX models in this benchmark implement TLL NNs by ``stacking'' these computation paths to make a fully connected NN (leading to sparse weight matrices: i.e. with many zero weights and biases). The \texttt{TLLnet} class (\url{https://github.com/jferlez/TLLnet}) contains the code necessary to generate these implementations via the \texttt{exportONNX} method.

\paragraph*{Specifications}  All specifications have as input constraints the hypercube $[-2,2]^2$. Since all networks have only a single output, the output properties consist of a randomly generated real number and a randomly generated inequality direction. Random output samples from the network are used to roughly ensure that the real number property has an equal likelihood of being within the output range of the NN and being outside of it (either above or below all NN outputs on the input constraint set). The inequality direction is generated independently and with each direction having an equal probability. This scheme biases the benchmark towards verification problems for which counterexamples exist. 

\paragraph*{Link} \url{https://github.com/jferlez/TLLVerifyBench}
\paragraph*{Commit}
199d2c26d0ec456e62906366b694a875a21ff7ef

\subsection{ACAS Xu}
\paragraph{Networks} The ACASXu benchmark consists of ten properties defined over 45 neural networks used to issue turn advisories to aircraft to avoid collisions. The neural networks have 300 neurons arranged in 6 layers, with ReLU activation functions. There are five inputs corresponding to the aircraft states, and five network outputs, where the minimum output is used as the turn advisory the system ultimately produces.

\paragraph{Specifications} We use the original 10 properties~\cite{katz2017reluplex}, where properties 1-4 are checked on all 45 networks as was done in later work by the original authors~\cite{katz2019marabou}. Properties 5-10 are checked on a single network. The total number of benchmarks is therefore 186. The original verification times ranged from seconds to days---including some benchmark instances that did not finish. This year we used a timeout of around two minutes (116 seconds) for each property, in order to fit within a total maximum runtime of six hours.





\subsection{safeNLP}

\begin{figure}[htbp]
    \centering
    \includegraphics[width=0.9\columnwidth]{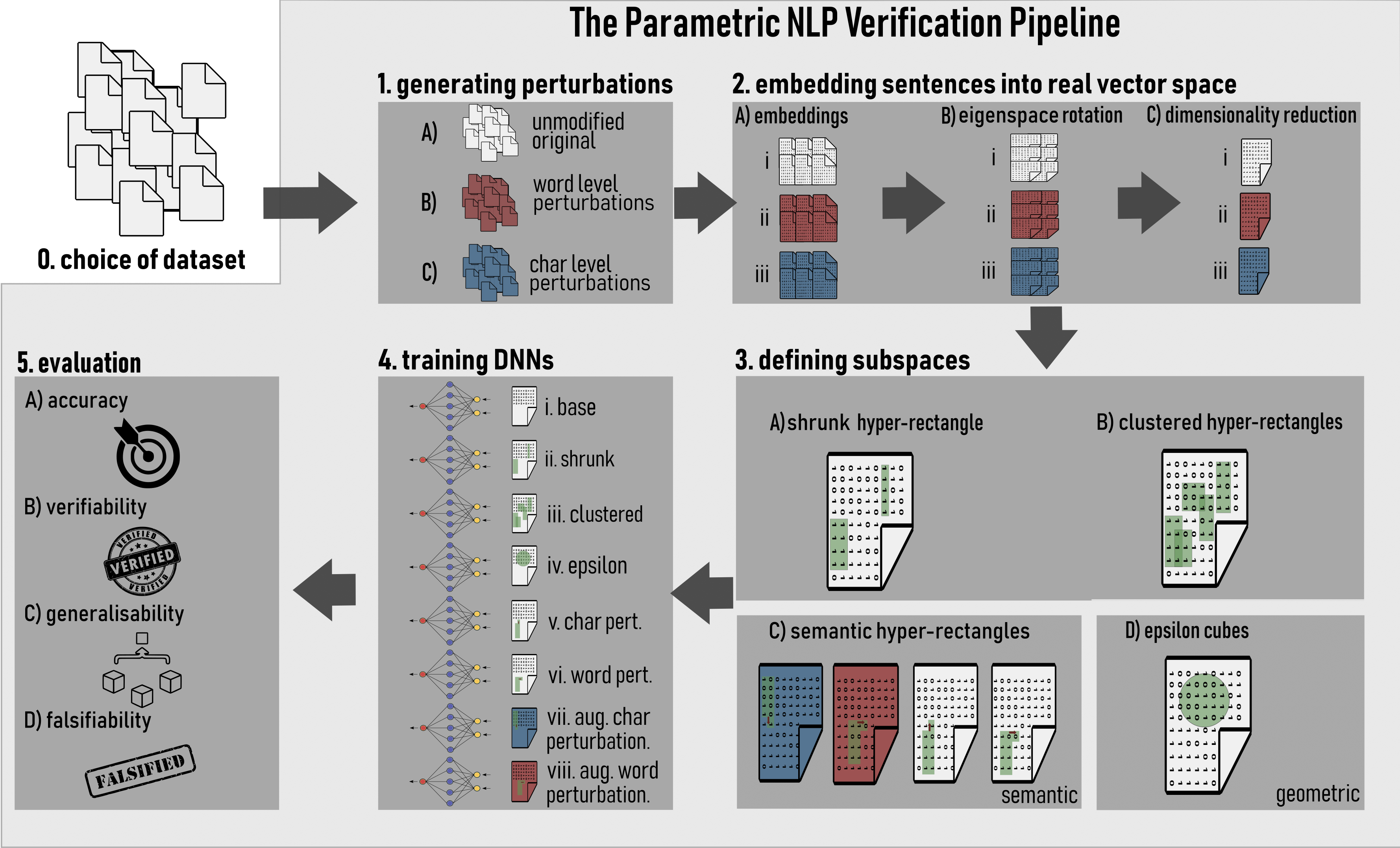}
    \caption{\small\emph{Generic approach to generating the NLP verification pipelines~\cite{casadio2023antonio,casadio2024nlp} deployed to obtain the safeNLP benchmark.}}
   \label{fig:antonio}
\end{figure}

\paragraph*{Proposed by} Marco Casadio, Ekaterina Komendantskaya, Luca Arnaboldi, Tanvi Dinkar.

\paragraph*{Motivation}
While considerable research has been dedicated to the verification of DNN-based systems in domains such as computer vision, there has been a notable lack of focus on the verification of natural language processing (NLP) systems. This is particularly critical given the rise of conversational agents across various domains, where inaccurate or misleading responses can cause real-world harm. For example, recent EU legislation~\cite{EUlaw} requires chatbots to disclose their non-human nature when queried, and developers of the chatbots should provide firm, and if possible, formal, guarantees that such disclosure will be given in an accurate manner. Medical assistants give another example where formal guarantees about the conversational agent responses are needed in order to safeguard against chatbots  generating harmful medical advice~\cite{bickmore2018patient}. While some initial work has been done in this area of NLP verification~\cite{jia2019certified,huang2019achieving,welbl2020towards,zhang2021certified,wang2023robustness,ko2019popqorn,du2021cert,shi2020robustness,bonaert2021fast}, no agreement on commonly  accepted benchmarks has been reached in this domain. To address this gap, we introduce safeNLP, the first such benchmark. 

\paragraph*{Application}
In~\cite{casadio2024nlp}, we have undertaken a large-scale study of the existing literature on  NLP verification, and distilled common patterns among the existing approaches. Usually, given a dataset consisting of sentences divided into classes, Large Language Models (LLMs) are used to embed these sentences into real-vector spaces, after which smaller neural networks are trained to classify the embedded vectors (relative to the originally given classes). For verification, one can generate meaning-preserving sentence perturbations, again embed them into vector spaces, and verify that subspaces that contain the (embeddings of) the perturbed sentences are classified correctly. Also, in line with classical verification pipelines~\cite{CasadioKDKKAR22}, one can use these input subspaces to train the neural network to be robust on them.  The problem was that each of the existing approaches~\cite{jia2019certified,huang2019achieving,welbl2020towards,zhang2021certified,wang2023robustness,ko2019popqorn,du2021cert,shi2020robustness,bonaert2021fast} used parts of this pipeline in different ways, which made it difficult to compare or audit the results. In~\cite{casadio2023antonio,casadio2024nlp}, we made a generic implementation of this pipeline, 
where each of the components of the pipeline is implemented in a modular and transparent way. For example, we can choose  and vary embedding functions, training modes, algorithms for sentence perturbations and algorithms for robust training, independently and modularly; as shown in Figure~\ref{fig:antonio}. This implementation was used to generate the presented VNNCOMP benchmark.

\begin{itemize}
    \item \emph{Datasets:} Although there was no clear consensus in~\cite{jia2019certified,huang2019achieving,welbl2020towards,zhang2021certified,wang2023robustness,ko2019popqorn,du2021cert,shi2020robustness,bonaert2021fast}, the most frequently used dataset in prior works was the IMDB dataset containing film reviews. Its disadvantage is unclear relation to safety critical domains that usually motivate verification efforts. On the other hand, none of the previously used datasets concerned safety-critical applications of NLP. We decided to address this problem, and therefore applied our generic NLP verification pipeline on two safety-critical datasets:  R-U-A-Robot~\cite{gros2021ruarobot}, which focuses on the chatbot disclosure problem, and Medical~\cite{abercrombie2022risk}, which addresses the issue of harmful advice provided by medical chatbots. Both datasets are pre-processed into two classes, positive and negative, to simplify the verification task. For further details on the pre-processing steps and datasets, see~\cite{casadio2024nlp} and the \href{https://github.com/ANTONIONLP/safeNLP}{benchmark GitHub repository}.
    \item \emph{Input Space:} In both datasets, sentences are transformed into fixed-size vector representations, i.e. embeddings, which serve as the inputs to the neural networks. For this VNNCOMP benchmark, we used Sentence-BERT~\cite{reimers-gurevych-2019-sentence}. 
    \item \emph{What to Verify:} For each dataset, we generated meaning-preseving sentence perturbations at character and word level as in Moradi et al.~\cite{moradi2021evaluating} and at sentence level with Vicuna~\cite{vicuna2023}. For each positive sentence in the dataset, the smallest hypercube containing the embeddings of all of its obtained perturbations formed one input subspace for verification. Such subspaces were obtained for all positive sentences from the given data set, and were subject to VNCCOMP verification challenge. 
    \item \emph{A note on broader impact:} Verified models can serve as filters for larger NLP systems: e.g. to screen inputs to ensure they meet safety criteria before being passed on to more complex models.
\end{itemize}

\paragraph*{Networks} The safeNLP benchmark includes two neural networks, each corresponding to a different dataset (R-U-A-Robot and Medical). Both networks share the same architecture, consisting of two fully-connected layers. The hidden layer has 128 units with a ReLU activation function, while the output layer has 2 units representing the two classification classes (positive/negative). To enhance the robustness of the networks to the specified safety requirements, they are trained using a custom PGD (Projected Gradient Descent)~\cite{madry2018towards} adversarial training technique. In particular, the PGD attack explores the above-mentioned subspaces of the input space (cf. also Figure~\ref{fig:antonio}). 

\paragraph*{Specifications} The benchmark uses hyper-rectangles in the 30-dimensional embedding space as the subspaces of choice, offering a computationally efficient way to define more precise and adaptable regions compared to the traditional $\epsilon$-cubes. 
The specifications require verifying that, for a given network and hyper-rectangle, every point within the hyper-rectangle is classified as the positive class by the network. To meet time constraints, we randomly select 1,080 such specifications, each linked to one of the two networks and a corresponding hyper-rectangle, with a timeout of 20 seconds per specification.

\paragraph*{Link} \url{https://github.com/ANTONIONLP/safeNLP}

\subsection{Real-world distribution shifts}
\paragraph*{Proposed by} the Marabou team.
\paragraph*{Motivation}
While robustness against handcrafted perturbations (e.g., norm-bounded) for perception networks are more commonly investigated, robustness against real-world distribution shifts~\cite{wu2022toward} are less studied but of practical interests. This benchmark set contains queries for verifying the latter type of robustness.  
\paragraph*{Networks} The network is a concatenation of a generative model and a MNIST classifier. The generative model is trained to take in an unperturbed image and an embedding of a particular type of distribution shifts in latent space, and produce a perturbed image. The distribution shift captured in this case is the "shear" perturbation. 
\paragraph*{Specifications} The verification task is to certify that a classifier correctly classifies all images in a perturbation set, which is a set of images generated by the generative model given a fixed image and a ball centering the mean perturbations on this image (in the latent space). This mean perturbation is computed by a prior network.
\paragraph*{Link} \url{https://github.com/wu-haoze/dist-shift-vnn-comp}

\subsection{CORA Benchmark}
\paragraph*{Proposed by} the CORA team.
\paragraph*{Motivation}
The verification of neural networks can be quite slow, i.e., the verification of a single instance can take multiple days -- which is often hard to justify, particularly in safety-critical scenarios. To encourage the fast verification of neural networks, our benchmark focuses on the verification time by setting a small timeout and testing three different (adversarial) training techniques that aim to ease the verifiability.
\paragraph*{Networks} The benchmark consists of one ReLU-neural network architecture (7x250 + ReLU), which was trained on three datasets, (MNIST, SVHN, and CIFAR10), using three different (adversarial) training methods, i.e., standard (point), interval-bound propagation, and set-based. Both interval-bound propagation and set-based training are training methods that improve the robustness of the trained neural network and aim to ease later verification. The neural networks are taken from the first evaluation run of~\cite{koller_et_al_2024}; please refer to~\cite{koller_et_al_2024} for the training details.
\paragraph*{Specifications} All networks are trained on classification tasks. The goal is to verify that no image within a given input set is incorrectly classified.
\paragraph*{Link} \url{https://github.com/kollerlukas/cora-vnncomp2024-benchmark}

\subsection{Additional Benchmarks}

We have not yet obtained benchmark descriptions for the following benchmarks: Collins Aerospace, CCTSDB, Metaroom, and yolo. We will update the report when these descriptions are available.
Artifacts of the benchmarks are available in the repository\footnote{\url{https://github.com/ChristopherBrix/vnncomp2024_benchmarks/tree/main/benchmarks}}.

\newpage
\section{Results}
\label{sec:results}

Each tool was run on each of the benchmarks and produced a \texttt{csv} result file, that was provided as feedback to the tool authors using the online execution platform.
The final \texttt{csv} files for each tool as well as scoring scripts are available online: \url{https://github.com/ChristopherBrix/vnncomp2024_results}.
The results were automatically analyzed to compute scores and create the statistics presented in this section.

\paragraph{\textbf{Update:}} \emph{Note that the two tools NeuralSAT and CORA were penalized for instances where they returned the correct result, but in a format that did not match the expected one.
We report the updated results in Section~\ref{sec:alternative_ranking}}.

%
%

\subsection{Regular Track}
The regular track contained all the benchmarks that were voted for by at least half of all participants.


\begin{table}[h]
\begin{center}
\caption{Overall Score} \label{tab:score}
{\setlength{\tabcolsep}{2pt}
\begin{tabular}[h]{@{}lll@{}}
\toprule
\textbf{\# ~} & \textbf{Tool} & \textbf{Score}\\
\midrule
1 & $\alpha$-$\beta$-CROWN & 1200.0 \\
2 & PyRAT & 1000.8 \\
3 & Marabou & 751.0 \\
4 & nnenum & 572.5 \\
5 & NNV & 530.0 \\
6 & NeVer2 & 262.3 \\
7 & CORA & 251.7 \\
8 & NeuralSAT & 0 \\
\bottomrule
\end{tabular}
}
\end{center}
\end{table}

\begin{figure}[h]
\centerline{\includegraphics[width=\textwidth]{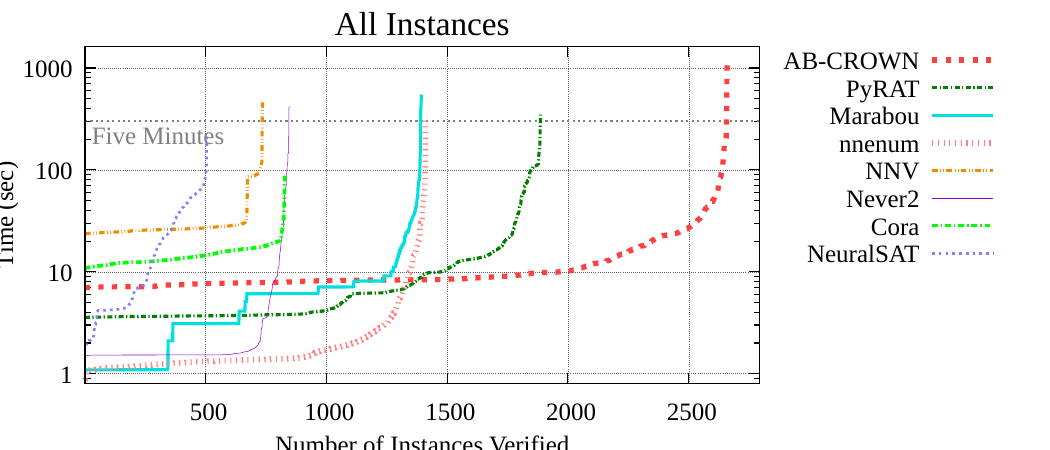}}
\caption{Cactus Plot for All Instances (Regular Track).}
\label{fig:quantPic}
\end{figure}

\subsection{Extended Track}
All benchmarks that were voted for by at least one team and did not make it into the regular track were part of the extended track.
Every benchmark was voted for at least once, so no benchmark was unscored.


\begin{table}[h]
\begin{center}
\caption{Overall Score} \label{tab:score_extended}
{\setlength{\tabcolsep}{2pt}
\begin{tabular}[h]{@{}lll@{}}
\toprule
\textbf{\# ~} & \textbf{Tool} & \textbf{Score}\\
\midrule
1 & $\alpha$-$\beta$-CROWN & 900.0 \\
2 & PyRAT & 398.5 \\
3 & nnenum & 72.2 \\
4 & NeuralSAT & 46.4 \\
\bottomrule
\end{tabular}
}
\end{center}
\end{table}

\begin{figure}[h]
\centerline{\includegraphics[width=\textwidth]{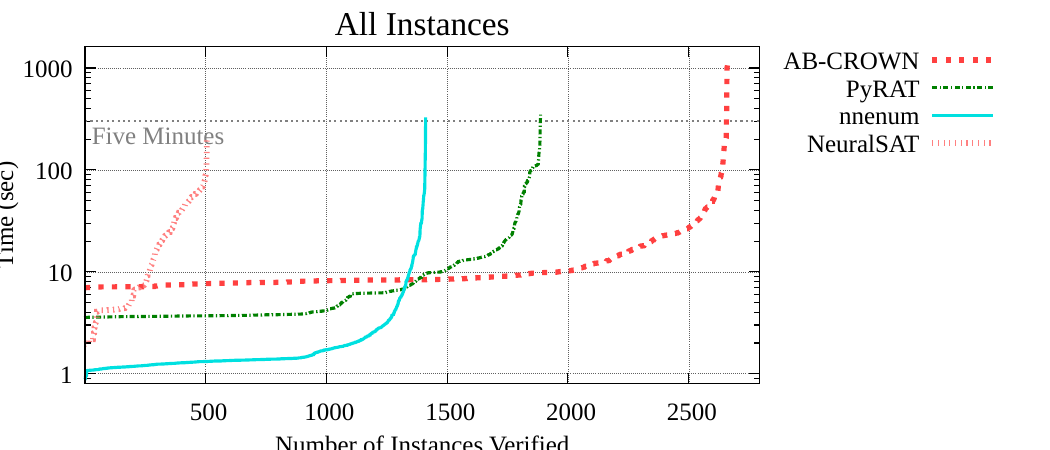}}
\caption{Cactus Plot for All Instances (Extended Track).
}
\label{fig:quantPic_extended}
\end{figure}

\clearpage
\subsection{Other Stats}

This section presents other statistics related to the measurements that are interesting but did not play a direct role in scoring this year.
The results reflect the tool performance on the regular track.



\begin{table}[h]
\begin{center}
\caption{Overhead} \label{tab:overhead}
{\setlength{\tabcolsep}{2pt}
\begin{tabular}[h]{@{}llr@{}}
\toprule
\textbf{\# ~} & \textbf{Tool} & \textbf{Seconds}\\
\midrule
1 & NeuralSAT & 0.3 \\
2 & nnenum & 0.9 \\
3 & Marabou & 1.1 \\
4 & NeVer2 & 1.5 \\
5 & PyRAT & 3.5 \\
6 & $\alpha$-$\beta$-CROWN & 7.0 \\
7 & CORA & 9.9 \\
8 & NNV & 23.5 \\
\bottomrule
\end{tabular}
}
\end{center}
\end{table}


\begin{table}[h]
\begin{center}
\caption{Num Benchmarks Participated} \label{tab:stats0}
{\setlength{\tabcolsep}{2pt}
\begin{tabular}[h]{@{}llr@{}}
\toprule
\textbf{\# ~} & \textbf{Tool} & \textbf{Count}\\
\midrule
1 & PyRAT & 12 \\
2 & Marabou & 12 \\
3 & $\alpha$-$\beta$-CROWN & 12 \\
4 & nnenum & 9 \\
5 & NNV & 9 \\
6 & NeuralSAT & 8 \\
7 & CORA & 8 \\
8 & NeVer2 & 4 \\
\bottomrule
\end{tabular}
}
\end{center}
\end{table}


\begin{table}[h]
\begin{center}
\caption{Num Instances Verified} \label{tab:stats1}
{\setlength{\tabcolsep}{2pt}
\begin{tabular}[h]{@{}llr@{}}
\toprule
\textbf{\# ~} & \textbf{Tool} & \textbf{Count}\\
\midrule
1 & $\alpha$-$\beta$-CROWN & 2285 \\
2 & PyRAT & 1760 \\
3 & nnenum & 1398 \\
4 & Marabou & 1394 \\
5 & NeVer2 & 848 \\
6 & CORA & 828 \\
7 & NNV & 736 \\
8 & NeuralSAT & 489 \\
\bottomrule
\end{tabular}
}
\end{center}
\end{table}


\begin{table}[h]
\begin{center}
\caption{Num SAT} \label{tab:stats2}
{\setlength{\tabcolsep}{2pt}
\begin{tabular}[h]{@{}llr@{}}
\toprule
\textbf{\# ~} & \textbf{Tool} & \textbf{Count}\\
\midrule
1 & $\alpha$-$\beta$-CROWN & 993 \\
2 & PyRAT & 890 \\
3 & nnenum & 755 \\
4 & Marabou & 704 \\
5 & NeVer2 & 526 \\
6 & CORA & 327 \\
7 & NNV & 308 \\
\bottomrule
\end{tabular}
}
\end{center}
\end{table}


\begin{table}[h]
\begin{center}
\caption{Num UNSAT} \label{tab:stats3}
{\setlength{\tabcolsep}{2pt}
\begin{tabular}[h]{@{}llr@{}}
\toprule
\textbf{\# ~} & \textbf{Tool} & \textbf{Count}\\
\midrule
1 & $\alpha$-$\beta$-CROWN & 1292 \\
2 & PyRAT & 870 \\
3 & Marabou & 690 \\
4 & nnenum & 643 \\
5 & CORA & 501 \\
6 & NeuralSAT & 489 \\
7 & NNV & 428 \\
8 & NeVer2 & 322 \\
\bottomrule
\end{tabular}
}
\end{center}
\end{table}


\begin{table}[h]
\begin{center}
\caption{Incorrect Results (or Missing CE)} \label{tab:stats4}
{\setlength{\tabcolsep}{2pt}
\begin{tabular}[h]{@{}llr@{}}
\toprule
\textbf{\# ~} & \textbf{Tool} & \textbf{Count}\\
\midrule
1 & Marabou & 403 \\
2 & NeuralSAT & 281 \\
3 & CORA & 101 \\
4 & nnenum & 2 \\
\bottomrule
\end{tabular}
}
\end{center}
\end{table}

\section{Conclusion and Ideas for Future Competitions}
\label{sec:conclusion}
This report summarizes the 5$^\text{th}$ Verification of Neural Networks Competition (VNN-COMP), held in 2024.
While we observed a significant increase in the diversity, complexity, and scale of the proposed benchmarks, the best-performing tools seem to converge to GPU-enabled linear bound propagation methods using a branch-and-bound framework.
In addition to the standardization of input formats (\texttt{onnx} and \texttt{vnnlib}) and evaluation hardware, introduced for VNN-COMP 2021, VNN-COMP 2024 also continued the standardized format for counter-examples and fully automated evaluation pipeline introduced in VNN-COMP 2022, requiring authors to provide complete installation scripts.
We hope that this increased standardization and automatization does not only simplify the evaluation during the competition but also enables practitioners and researchers to more easily apply a range of state-of-the-art verification methods to their individual problems.

VNN-COMP 2024, successfully implemented a range of improvement opportunities identified during the previous iteration. These included requiring witnesses of found counter-examples to disambiguate tool disagreement, increasing automatization to enable a smoother final evaluation, and making a broader range of AWS instances available to allow for a better fit with tools' requirements. 
However, some issues were identified in the evaluation particularly with respect to clarity of some output formats and parsing for computing scores, and further improvements will be made to avoid such issues in the future.
Further ideas for future competitions include the use of scored benchmarks specifically designed for year-on-year progress tracking, the reduction of tool tuning, a batch-processing mode, and more rigorous soundness evaluation.

\section*{Acknowledgements}
%

The 2024 competition was supported by CEA-List, which provided funds for computation costs.

Additionally, this research was supported in part by the Air Force Research Laboratory Information Directorate, through the Air Force Office of Scientific Research Summer Faculty Fellowship
Program, Contract Numbers FA8750-15-3-6003, FA9550-15-0001 and FA9550-20-F-0005.
This material is based upon work supported by the Air Force Office of Scientific Research under award numbers FA9550-19-1-0288, FA9550-21-1-0121, and FA9550-22-1-0019, the National Science Foundation (NSF) under grant numbers 2107035, 2220401, 2220418, 2220426, and 2238133, and the Defense Advanced Research Projects Agency (DARPA) Assured Neuro Symbolic Learning and Reasoning (ANSR)
programs through contract number and FA8750-23-C-0518.
Any opinions, findings, and conclusions or recommendations expressed in this material are those of the author(s) and do not necessarily reflect the views of the United States Air Force, DARPA, nor NSF.

Tool and benchmark authors listed in \Cref{sec:participants} and \Cref{sec:benchmarks} participated in the preparation and review of this report.

\clearpage
\label{sect:bib}
\bibliographystyle{plain}
\bibliography{bib/nnv, bib/nnenum, bib/peregriNN, bib/verinet, bib/oval,bib/venus,bib/MIPVerify, bib/mnbab, bib/alpha-beta-CROWN, bib/collins,bib/dnnf,bib/nvjl,bib/nn4sys,bib/Marabou,bib/RPM,bib/AVeriNN,bib/VeRAPAk,bib/general,bib/traffic-signs-recognition,bib/neuralsat, bib/pyrat, bib/vit, bib/cora, bib/never2, bib/safeNLP}

\begin{thebibliography}{10}

\bibitem{BelgianTrafficSignDatabase}
{Belgian Traffic Sign Database}.
\newblock \url{https://www.kaggle.com/datasets/shazaelmorsh/trafficsigns}.
\newblock Accessed: March 25th, 2023.

\bibitem{ChineseTrafficSignDatabase}
{Chinese Traffic Sign Database}.
\newblock \url{https://www.kaggle.com/datasets/dmitryyemelyanov/chinese-traffic-signs}.
\newblock Accessed: March 25th, 2023.

\bibitem{cora-website}
{CORA: A Tool for Continuous Reachability Analysis }.
\newblock \url{https://https://cora.in.tum.de//}.
\newblock Accessed: October 15th, 2024.

\bibitem{GTSRB}
{German Traffic Sign Recognition Benchmark}.
\newblock \url{https://www.kaggle.com/datasets/meowmeowmeowmeowmeow/gtsrb-german-traffic-sign?datasetId=82373&language=Python}.
\newblock Accessed: March 25th, 2023.

\bibitem{pyrat-website}
{PyRAT Analyzer website}.
\newblock \url{https://pyrat-analyzer.com/}.
\newblock Accessed: December 20th, 2024.

\bibitem{abercrombie2022risk}
Gavin Abercrombie and Verena Rieser.
\newblock Risk-graded safety for handling medical queries in conversational ai.
\newblock In {\em Proceedings of the 2nd Conference of the Asia-Pacific Chapter of the Association for Computational Linguistics and the 12th International Joint Conference on Natural Language Processing}, pages 234--243, 2022.

\bibitem{althoff_2015}
Matthias Althoff.
\newblock An introduction to {CORA} 2015.
\newblock In {\em Proc. of the Workshop on Applied Verification for Continuous and Hybrid Systems (ARCH)}, pages 120--151, 2015.

\bibitem{ArjomandBigdeli2024}
Ali ArjomandBigdeli, Andrew Mata, and Stanley Bak.
\newblock {\em Verification of Neural Network Control Systems in Continuous Time}, page 100–115.
\newblock Springer Nature Switzerland, 2024.

\bibitem{bak2020vnn}
Stanley Bak.
\newblock Execution-guided overapproximation (ego) for improving scalability of neural network verification, 2020.

\bibitem{bak2021nnenum}
Stanley Bak.
\newblock nnenum: Verification of relu neural networks with optimized abstraction refinement.
\newblock In {\em NASA Formal Methods Symposium}, pages 19--36. Springer, 2021.

\bibitem{bak2021vnncomp}
Stanley Bak, Changliu Liu, and Taylor Johnson.
\newblock The second international verification of neural networks competition (vnn-comp 2021): Summary and results, 2021.

\bibitem{bak2020cav}
Stanley Bak, Hoang-Dung Tran, Kerianne Hobbs, and Taylor~T. Johnson.
\newblock Improved geometric path enumeration for verifying {ReLU} neural networks.
\newblock In {\em 32nd International Conference on Computer-Aided Verification (CAV)}, July 2020.

\bibitem{bickmore2018patient}
Timothy~W Bickmore, Ha~Trinh, Stefan Olafsson, Teresa~K O'Leary, Reza Asadi, Nathaniel~M Rickles, and Ricardo Cruz.
\newblock Patient and consumer safety risks when using conversational assistants for medical information: an observational study of siri, alexa, and google assistant.
\newblock {\em Journal of medical Internet research}, 20(9):e11510, 2018.

\bibitem{bonaert2021fast}
Gregory Bonaert, Dimitar~I Dimitrov, Maximilian Baader, and Martin Vechev.
\newblock Fast and precise certification of transformers.
\newblock In {\em Proceedings of the 42nd ACM SIGPLAN International Conference on Programming Language Design and Implementation}, pages 466--481, 2021.

\bibitem{brix2023vnncomp}
Christopher Brix, Stanley Bak, Changliu Liu, and Taylor~T. Johnson.
\newblock The fourth international verification of neural networks competition (vnn-comp 2023): Summary and results.
\newblock 2023.

\bibitem{brix2023years}
Christopher Brix, Mark~Niklas Müller, Stanley Bak, Taylor~T. Johnson, and Changliu Liu.
\newblock First three years of the international verification of neural networks competition (vnn-comp), 2023.

\bibitem{bunelunified2018}
Rudy Bunel, Ilker Turkaslan, Philip~HS Torr, Pushmeet Kohli, and M~Pawan Kumar.
\newblock A unified view of piecewise linear neural network verification.
\newblock {\em Advances in Neural Information Processing Systems}, 2018.

\bibitem{casadio2023antonio}
Marco Casadio, Luca Arnaboldi, Matthew~L Daggitt, Omri Isac, Tanvi Dinkar, Daniel Kienitz, Verena Rieser, and Ekaterina Komendantskaya.
\newblock Antonio: Towards a systematic method for generating nlp benchmarks for verification.
\newblock In {\em Proceedings of the 6th Workshop on Formal}, volume~16, pages 59--70, 2023.

\bibitem{casadio2024nlp}
Marco Casadio, Tanvi Dinkar, Ekaterina Komendantskaya, Luca Arnaboldi, Omri Isac, Matthew~L Daggitt, Guy Katz, Verena Rieser, and Oliver Lemon.
\newblock Nlp verification: Towards a general methodology for certifying robustness.
\newblock {\em arXiv preprint arXiv:2403.10144}, 2024.

\bibitem{CasadioKDKKAR22}
Marco Casadio, Ekaterina Komendantskaya, Matthew~L. Daggitt, Wen Kokke, Guy Katz, Guy Amir, and Idan Refaeli.
\newblock Neural network robustness as a verification property: {A} principled case study.
\newblock In Sharon Shoham and Yakir Vizel, editors, {\em Computer Aided Verification - 34th International Conference, {CAV} 2022, Haifa, Israel, August 7-10, 2022, Proceedings, Part {I}}, volume 13371 of {\em Lecture Notes in Computer Science}, pages 219--231. Springer, 2022.

\bibitem{vicuna2023}
Wei-Lin Chiang, Zhuohan Li, Zi~Lin, Ying Sheng, Zhanghao Wu, Hao Zhang, Lianmin Zheng, Siyuan Zhuang, Yonghao Zhuang, Joseph~E. Gonzalez, Ion Stoica, and Eric~P. Xing.
\newblock Vicuna: An open-source chatbot impressing gpt-4 with 90\%* chatgpt quality, March 2023.

\bibitem{vnnlib}
Stefano Demarchi, Dario Guidotti, Luca Pulina, and Armando Tacchella.
\newblock Supporting standardization of neural networks verification with vnnlib and coconet.
\newblock In Nina Narodytska, Guy Amir, Guy Katz, and Omri Isac, editors, {\em Proceedings of the 6th Workshop on Formal Methods for ML-Enabled Autonomous Systems}, volume~16 of {\em Kalpa Publications in Computing}, pages 47--58. EasyChair, 2023.

\bibitem{demarchi2024never2}
Stefano Demarchi, Dario Guidotti, Luca Pulina, and Armando Tacchella.
\newblock Never2: Learning and verification of neural networks.
\newblock {\em Soft Computing}, 2024.

\bibitem{dosovitskiy2020image}
Alexey Dosovitskiy, Lucas Beyer, Alexander Kolesnikov, Dirk Weissenborn, Xiaohua Zhai, Thomas Unterthiner, Mostafa Dehghani, Matthias Minderer, Georg Heigold, Sylvain Gelly, et~al.
\newblock An image is worth 16x16 words: Transformers for image recognition at scale.
\newblock In {\em International Conference on Learning Representations}, 2020.

\bibitem{du2021cert}
Tianyu Du, Shouling Ji, Lujia Shen, Yao Zhang, Jinfeng Li, Jie Shi, Chengfang Fang, Jianwei Yin, Raheem Beyah, and Ting Wang.
\newblock Cert-rnn: Towards certifying the robustness of recurrent neural networks.
\newblock {\em CCS}, 21(2021):15--19, 2021.

\bibitem{duong2023dpllt}
Hai Duong, Linhan Li, ThanhVu Nguyen, and Matthew Dwyer.
\newblock {A DPLL(T) Framework for Verifying Deep Neural Networks}, 2023.
\newblock arXiv, 25 pages.

\bibitem{duong2024harnessing}
Hai Duong, Dong Xu, ThanhVu Nguyen, and Matthew~B Dwyer.
\newblock Harnessing neuron stability to improve dnn verification.
\newblock {\em Proceedings of the ACM on Software Engineering}, 1(FSE):859--881, 2024.

\bibitem{durand2022reciph}
Serge Durand, Augustin Lemesle, Zakaria Chihani, Caterina Urban, and Fran{\c{c}}ois Terrier.
\newblock Reciph: Relational coefficients for input partitioning heuristic.
\newblock In {\em 1st Workshop on Formal Verification of Machine Learning (WFVML 2022)}, 2022.

\bibitem{ForMuLA}
{EASA and Collins Aerospace}.
\newblock {Formal Methods use for Learning Assurance (ForMuLA)}.
\newblock Technical report, April 2023.

\bibitem{FerlezKS22}
James Ferlez, Haitham Khedr, and Yasser Shoukry.
\newblock Fast {BATLLNN:} fast box analysis of two-level lattice neural networks.
\newblock In Ezio Bartocci and Sylvie Putot, editors, {\em {HSCC} '22: 25th {ACM} International Conference on Hybrid Systems: Computation and Control, Milan, Italy, May 4 - 6, 2022}, pages 23:1--23:11. {ACM}, 2022.

\bibitem{gowal2018effectiveness}
Sven Gowal, Krishnamurthy Dvijotham, Robert Stanforth, Rudy Bunel, Chongli Qin, Jonathan Uesato, Relja Arandjelovic, Timothy Mann, and Pushmeet Kohli.
\newblock On the effectiveness of interval bound propagation for training verifiably robust models.
\newblock {\em arXiv preprint arXiv:1810.12715}, 2018.

\bibitem{gros2021ruarobot}
David Gros, Yu~Li, and Zhou Yu.
\newblock The rua-robot dataset: Helping avoid chatbot deception by detecting user questions about human or non-human identity.
\newblock In {\em Proceedings of the 59th Annual Meeting of the Association for Computational Linguistics and the 11th International Joint Conference on Natural Language Processing (Volume 1: Long Papers)}, pages 6999--7013, 2021.

\bibitem{guidotti2021pynever}
Dario Guidotti, Luca Pulina, and Armando Tacchella.
\newblock pynever: A framework for learning and verification of neural networks.
\newblock In {\em Automated Technology for Verification and Analysis: 19th International Symposium, ATVA 2021, Gold Coast, QLD, Australia, October 18--22, 2021, Proceedings 19}, pages 357--363. Springer, 2021.

\bibitem{huang2019achieving}
Po-Sen Huang, Robert Stanforth, Johannes Welbl, Chris Dyer, Dani Yogatama, Sven Gowal, Krishnamurthy Dvijotham, and Pushmeet Kohli.
\newblock Achieving verified robustness to symbol substitutions via interval bound propagation.
\newblock In {\em Proceedings of the 2019 Conference on Empirical Methods in Natural Language Processing and the 9th International Joint Conference on Natural Language Processing (EMNLP-IJCNLP)}, pages 4083--4093, 2019.

\bibitem{hubara2016binarized}
Itay Hubara, Matthieu Courbariaux, Daniel Soudry, Ran El-Yaniv, and Yoshua Bengio.
\newblock {Binarized Neural Networks}.
\newblock {\em Advances in Neural Information Processing Systems}, 29, 2016.

\bibitem{jia2019certified}
Robin Jia, Aditi Raghunathan, Kerem G{\"o}ksel, and Percy Liang.
\newblock Certified robustness to adversarial word substitutions.
\newblock In {\em Proceedings of the 2019 Conference on Empirical Methods in Natural Language Processing and the 9th International Joint Conference on Natural Language Processing (EMNLP-IJCNLP)}, pages 4129--4142, 2019.

\bibitem{katz2017reluplex}
Guy Katz, Clark Barrett, David~L Dill, Kyle Julian, and Mykel~J Kochenderfer.
\newblock Reluplex: An efficient smt solver for verifying deep neural networks.
\newblock In {\em International Conference on Computer Aided Verification}, pages 97--117. Springer, 2017.

\bibitem{katz2019marabou}
Guy Katz, Derek~A Huang, Duligur Ibeling, Kyle Julian, Christopher Lazarus, Rachel Lim, Parth Shah, Shantanu Thakoor, Haoze Wu, Aleksandar Zelji{\'c}, et~al.
\newblock The marabou framework for verification and analysis of deep neural networks.
\newblock In {\em International Conference on Computer Aided Verification}, pages 443--452. Springer, 2019.

\bibitem{katz2021veri}
Sydney~M. Katz, Anthony~L. Corso, Christopher~A. Strong, and Mykel~J. Kochenderfer.
\newblock Verification of image-based neural network controllers using generative models, 2021.

\bibitem{kirov2023benchmark}
Dmitrii Kirov and Simone~Fulvio Rollini.
\newblock Benchmark: remaining useful life predictor for aircraft equipment.
\newblock In {\em International Conference on Bridging the Gap between AI and Reality}, pages 299--304. Springer, 2023.

\bibitem{kirov2023formal}
Dmitrii Kirov, Simone~Fulvio Rollini, Luigi Di~Guglielmo, and Darren Cofer.
\newblock Formal verification of a neural network based prognostics system for aircraft equipment.
\newblock In {\em International Conference on Bridging the Gap between AI and Reality}, pages 225--240. Springer, 2023.

\bibitem{ko2019popqorn}
Ching-Yun Ko, Zhaoyang Lyu, Lily Weng, Luca Daniel, Ngai Wong, and Dahua Lin.
\newblock Popqorn: Quantifying robustness of recurrent neural networks.
\newblock In {\em International Conference on Machine Learning}, pages 3468--3477. PMLR, 2019.

\bibitem{kochdumper_et_al_2023}
Niklas Kochdumper, Christian Schilling, Matthias Althoff, and Stanley Bak.
\newblock Open- and closed-loop neural network verification using polynomial zonotopes.
\newblock In {\em NASA Formal Methods}, pages 16--36, 2023.

\bibitem{koller_et_al_2024}
Lukas Koller, Tobias Ladner, and Matthias Althoff.
\newblock Set-based training for neural network verification.
\newblock {\em arXiv preprint arXiv:2401.14961}, 2024.

\bibitem{EUlaw}
Mauritz Kop.
\newblock Eu artificial intelligence act: The european approach to ai, 2021.

\bibitem{kraska18case}
Tim Kraska, Alex Beutel, Ed~H Chi, Jeffrey Dean, and Neoklis Polyzotis.
\newblock The case for learned index structures.
\newblock In {\em Proceedings of the 2018 International Conference on Management of Data}, 2018.

\bibitem{ladner_althoff_2023}
Tobias Ladner and Matthias Althoff.
\newblock Automatic abstraction refinement in neural network verification using sensitivity analysis.
\newblock In {\em Proc. of the Int. Conf. on Hybrid Systems: Computation and Control (HSCC)}, pages 1--13, 2023.

\bibitem{pyrat2024}
Augustin Lemesle, Julien Lehmann, and Le~Gall Tristan.
\newblock {Neural Network Verification with PyRAT}.
\newblock {\em arXiv preprint arXiv:2410.23903}, 2024.

\bibitem{manzanas2023cav}
Diego~Manzanas Lopez, Sung~Woo Choi, Hoang-Dung Tran, and Taylor~T. Johnson.
\newblock {NNV 2.0}: The neural network verification tool.
\newblock In {\em 35th International Conference on Computer-Aided Verification (CAV)}, July 2023.

\bibitem{madry2017towards}
Aleksander Madry, Aleksandar Makelov, Ludwig Schmidt, Dimitris Tsipras, and Adrian Vladu.
\newblock Towards deep learning models resistant to adversarial attacks.
\newblock {\em arXiv preprint arXiv:1706.06083}, 2017.

\bibitem{madry2018towards}
Aleksander Madry, Aleksandar Makelov, Ludwig Schmidt, Dimitris Tsipras, and Adrian Vladu.
\newblock Towards deep learning models resistant to adversarial attacks.
\newblock In {\em International Conference on Learning Representations}, 2018.

\bibitem{manzanas2022formats}
Diego Manzanas~Lopez, Patrick Musau, Nathaniel Hamilton, and Taylor Johnson.
\newblock Reachability analysis of a general class of neural ordinary differential equation.
\newblock In {\em Proceedings of the 20th International Conference on Formal Modeling and Analysis of Timed Systems (FORMATS 2022), Co-Located with CONCUR, FMICS, and QEST as part of CONFEST 2022.}, Warsaw, Poland, September 2022.

\bibitem{moradi2021evaluating}
Milad Moradi and Matthias Samwald.
\newblock Evaluating the robustness of neural language models to input perturbations.
\newblock In {\em Proceedings of the 2021 Conference on Empirical Methods in Natural Language Processing}, pages 1558--1570, 2021.

\bibitem{muller2022vnncomp}
Mark~Niklas M{\"u}ller, Christopher Brix, Stanley Bak, Changliu Liu, and Taylor~T Johnson.
\newblock The third international verification of neural networks competition (vnn-comp 2022): Summary and results.
\newblock {\em arXiv preprint arXiv:2212.10376}, 2022.

\bibitem{postovan2023architecturing}
Andreea Postovan and M{\u{a}}d{\u{a}}lina Era{\c{s}}cu.
\newblock Architecturing binarized neural networks for traffic sign recognition.
\newblock {\em arXiv preprint arXiv:2303.15005}, 2023.

\bibitem{reimers-gurevych-2019-sentence}
Nils Reimers and Iryna Gurevych.
\newblock Sentence-{BERT}: Sentence embeddings using {S}iamese {BERT}-networks.
\newblock In {\em Proceedings of the 2019 Conference on Empirical Methods in Natural Language Processing and the 9th International Joint Conference on Natural Language Processing (EMNLP-IJCNLP)}, pages 3982--3992, Hong Kong, China, November 2019. Association for Computational Linguistics.

\bibitem{shi2024certified}
Zhouxing Shi, Cho-Jui Hsieh, and Huan Zhang.
\newblock Certified training with branch-and-bound: A case study on lyapunov-stable neural control.
\newblock {\em arXiv preprint arXiv:2411.18235}, 2024.

\bibitem{shi2024genbab}
Zhouxing Shi, Qirui Jin, Zico Kolter, Suman Jana, Cho-Jui Hsieh, and Huan Zhang.
\newblock Neural network verification with branch-and-bound for general nonlinearities.
\newblock {\em arXiv preprint arXiv:2405.21063}, 2024.

\bibitem{shi2021fast}
Zhouxing Shi, Yihan Wang, Huan Zhang, Jinfeng Yi, and Cho-Jui Hsieh.
\newblock Fast certified robust training with short warmup.
\newblock {\em Advances in Neural Information Processing Systems}, 34:18335--18349, 2021.

\bibitem{shi2019robustness}
Zhouxing Shi, Huan Zhang, Kai-Wei Chang, Minlie Huang, and Cho-Jui Hsieh.
\newblock Robustness verification for transformers.
\newblock In {\em International Conference on Learning Representations}, 2019.

\bibitem{shi2020robustness}
Zhouxing Shi, Huan Zhang, Kai-Wei Chang, Minlie Huang, and Cho-Jui Hsieh.
\newblock Robustness verification for transformers, 2020.

\bibitem{simonyan2014very}
Karen Simonyan and Andrew Zisserman.
\newblock Very deep convolutional networks for large-scale image recognition.
\newblock {\em arXiv preprint arXiv:1409.1556}, 2014.

\bibitem{DeepPoly:19}
Gagandeep Singh, Timon Gehr, Markus P{\"{u}}schel, and Martin~T. Vechev.
\newblock An abstract domain for certifying neural networks.
\newblock {\em Proc. {ACM} Program. Lang.}, 3({POPL}):41:1--41:30, 2019.

\bibitem{szegedy2013intriguing}
Christian Szegedy, Wojciech Zaremba, Ilya Sutskever, Joan Bruna, Dumitru Erhan, Ian Goodfellow, and Rob Fergus.
\newblock {Intriguing Properties of Neural Networks}.
\newblock {\em arXiv preprint arXiv:1312.6199}, 2013.

\bibitem{Tjeng2019EvaluatingRO}
Vincent Tjeng, Kai~Y. Xiao, and Russ Tedrake.
\newblock Evaluating robustness of neural networks with mixed integer programming.
\newblock In {\em ICLR}, 2019.

\bibitem{tran2021fac}
H.~D. Tran, N.~Pal, D.~Lopez, P.~Musau, X.~Yang, W.~Xiang L.~Nguyen, S.~Bak, , and T.~T. Johnson.
\newblock Verification of piecewise deep neural networks: A star set approach with zonotope pre-filter.
\newblock {\em Formal aspects of computing}, 2021.

\bibitem{tran2020cav}
Hoang-Dung Tran, Stanley Bak, Weiming Xiang, and Taylor~T. Johnson.
\newblock Verification of deep convolutional neural networks using imagestars.
\newblock In {\em 32nd International Conference on Computer-Aided Verification (CAV)}. Springer, July 2020.

\bibitem{tran2019emsoft}
Hoang-Dung Tran, Feiyang Cei, Diego~Manzanas Lopez, Taylor~T. Johnson, and Xenofon Koutsoukos.
\newblock Safety verification of cyber-physical systems with reinforcement learning control.
\newblock In {\em ACM SIGBED International Conference on Embedded Software (EMSOFT'19)}. ACM, October 2019.

\bibitem{tran2023hscc}
Hoang~Dung Tran, SungWoo Choi, Tomoya Yamaguchi, Bardh Hoxha, and Danil Prokhorov.
\newblock Verification of recurrent neural networks using star reachability.
\newblock In {\em The 26th ACM International Conference on Hybrid Systems: Computation and Control (HSCC)}, May 2023.

\bibitem{tran2019fm}
Hoang-Dung Tran, Patrick Musau, Diego~Manzanas Lopez, Xiaodong Yang, Luan~Viet Nguyen, Weiming Xiang, and Taylor~T. Johnson.
\newblock Star-based reachability analysis for deep neural networks.
\newblock In {\em 23rd International Symposium on Formal Methods (FM'19)}. Springer International Publishing, October 2019.

\bibitem{tran2021cav}
Hoang-Dung Tran, Neelanjana Pal, Patrick Musau, Xiaodong Yang, Nathaniel~P. Hamilton, Diego~Manzanas Lopez, Stanley Bak, and Taylor~T. Johnson.
\newblock Robustness verification of semantic segmentation neural networks using relaxed reachability.
\newblock In {\em 33rd International Conference on Computer-Aided Verification (CAV)}. Springer, July 2021.

\bibitem{tran2020cav_tool}
Hoang-Dung Tran, Xiaodong Yang, Diego~Manzanas Lopez, Patrick Musau, Luan~Viet Nguyen, Weiming Xiang, Stanley Bak, and Taylor~T. Johnson.
\newblock {NNV}: The neural network verification tool for deep neural networks and learning-enabled cyber-physical systems.
\newblock In {\em 32nd International Conference on Computer-Aided Verification (CAV)}, July 2020.

\bibitem{vaswani2017attention}
Ashish Vaswani, Noam Shazeer, Niki Parmar, Jakob Uszkoreit, Llion Jones, Aidan~N Gomez, {\L}ukasz Kaiser, and Illia Polosukhin.
\newblock Attention is all you need.
\newblock {\em Advances in neural information processing systems}, 30, 2017.

\bibitem{wang2021betacrown}
Shiqi Wang, Huan Zhang, Kaidi Xu, Xue Lin, Suman Jana, Cho-Jui Hsieh, and Zico Kolter.
\newblock {Beta-CROWN}: Efficient bound propagation with per-neuron split constraints for complete and incomplete neural network verification.
\newblock {\em arXiv preprint arXiv:2103.06624}, 2021.

\bibitem{wang2023robustness}
Yibin Wang, Yichen Yang, Di~He, and Kun He.
\newblock Robustness-aware word embedding improves certified robustness to adversarial word substitutions.
\newblock In {\em Findings of the Association for Computational Linguistics: ACL 2023}, pages 673--687, 2023.

\bibitem{wei2023convex}
Dennis Wei, Haoze Wu, Min Wu, Pin-Yu Chen, Clark Barrett, and Eitan Farchi.
\newblock Convex bounds on the softmax function with applications to robustness verification.
\newblock In {\em International Conference on Artificial Intelligence and Statistics}, pages 6853--6878. PMLR, 2023.

\bibitem{welbl2020towards}
Johannes Welbl, Po-Sen Huang, Robert Stanforth, Sven Gowal, Krishnamurthy~Dj Dvijotham, Martin Szummer, and Pushmeet Kohli.
\newblock Towards verified robustness under text deletion interventions.
\newblock 2020.

\bibitem{vegas}
Haoze Wu, Clark Barrett, Mahmood Sharif, Nina Narodytska, and Gagandeep Singh.
\newblock Scalable verification of gnn-based job schedulers.
\newblock 6(OOPSLA2), oct 2022.

\bibitem{wu2024marabou}
Haoze Wu, Omri Isac, Aleksandar Zelji{\'c}, Teruhiro Tagomori, Matthew Daggitt, Wen Kokke, Idan Refaeli, Guy Amir, Kyle Julian, Shahaf Bassan, et~al.
\newblock Marabou 2.0: a versatile formal analyzer of neural networks.
\newblock In {\em International Conference on Computer Aided Verification}, pages 249--264. Springer, 2024.

\bibitem{wu2020parallelization}
Haoze Wu, Alex Ozdemir, Aleksandar Zeljic, Kyle Julian, Ahmed Irfan, Divya Gopinath, Sadjad Fouladi, Guy Katz, Corina Pasareanu, and Clark Barrett.
\newblock Parallelization techniques for verifying neural networks.
\newblock In {\em \# PLACEHOLDER\_PARENT\_METADATA\_VALUE\#}, volume~1, pages 128--137. TU Wien Academic Press, 2020.

\bibitem{wu2022toward}
Haoze Wu, Teruhiro Tagomori, Alexander Robey, Fengjun Yang, Nikolai Matni, George Pappas, Hamed Hassani, Corina Pasareanu, and Clark Barrett.
\newblock Toward certified robustness against real-world distribution shifts.
\newblock {\em arXiv preprint arXiv:2206.03669}, 2022.

\bibitem{wu2022efficient}
Haoze Wu, Aleksandar Zelji{\'c}, Guy Katz, and Clark Barrett.
\newblock Efficient neural network analysis with sum-of-infeasibilities.
\newblock In {\em International Conference on Tools and Algorithms for the Construction and Analysis of Systems}, pages 143--163. Springer, 2022.

\bibitem{xu2020automatic}
Kaidi Xu, Zhouxing Shi, Huan Zhang, Yihan Wang, Kai-Wei Chang, Minlie Huang, Bhavya Kailkhura, Xue Lin, and Cho-Jui Hsieh.
\newblock Automatic perturbation analysis for scalable certified robustness and beyond.
\newblock {\em Advances in Neural Information Processing Systems}, 33, 2020.

\bibitem{xu2021fast}
Kaidi Xu, Huan Zhang, Shiqi Wang, Yihan Wang, Suman Jana, Xue Lin, and Cho-Jui Hsieh.
\newblock {Fast and Complete}: Enabling complete neural network verification with rapid and massively parallel incomplete verifiers.
\newblock In {\em International Conference on Learning Representations}, 2021.

\bibitem{yang2024lyapunov}
Lujie Yang, Hongkai Dai, Zhouxing Shi, Cho-Jui Hsieh, Russ Tedrake, and Huan Zhang.
\newblock Lyapunov-stable neural control for state and output feedback: A novel formulation.
\newblock In {\em Forty-first International Conference on Machine Learning}, 2024.

\bibitem{zhang2022general}
Huan Zhang*, Shiqi Wang*, Kaidi Xu*, Linyi Li, Bo~Li, Suman Jana, Cho-Jui Hsieh, and J~Zico Kolter.
\newblock General cutting planes for bound-propagation-based neural network verification.
\newblock {\em Advances in Neural Information Processing Systems (NeurIPS)}, 2022.

\bibitem{zhang2018efficient}
Huan Zhang, Tsui-Wei Weng, Pin-Yu Chen, Cho-Jui Hsieh, and Luca Daniel.
\newblock Efficient neural network robustness certification with general activation functions.
\newblock {\em Advances in Neural Information Processing Systems}, 31:4939--4948, 2018.

\bibitem{zhang2020lightweight}
Jianming Zhang, Wei Wang, Chaoquan Lu, Jin Wang, and Arun~Kumar Sangaiah.
\newblock {Lightweight Deep Network for Traffic Sign Classification}.
\newblock {\em Annals of Telecommunications}, 75:369--379, 2020.

\bibitem{zhang2021certified}
Yuhao Zhang, Aws Albarghouthi, and Loris D’Antoni.
\newblock Certified robustness to programmable transformations in lstms.
\newblock In {\em Proceedings of the 2021 Conference on Empirical Methods in Natural Language Processing}, pages 1068--1083, 2021.

\bibitem{zhou2024scalable}
Duo Zhou, Christopher Brix, Grani~A Hanasusanto, and Huan Zhang.
\newblock Scalable neural network verification with branch-and-bound inferred cutting planes.
\newblock In {\em The Thirty-eighth Annual Conference on Neural Information Processing Systems}, 2024.

\bibitem{zhou2024testing}
Xingjian Zhou, Hongji Xu, Andy Xu, Zhouxing Shi, Cho-Jui Hsieh, and Huan Zhang.
\newblock Testing neural network verifiers: A soundness benchmark with hidden counterexamples.
\newblock {\em arXiv preprint arXiv:2412.03154}, 2024.

\end{thebibliography}


\appendix

\clearpage
\section{Detailed Results}
In this section, we provide more fine-grained results.

\subsection{Regular Track}
\label{sec:benchmark_results_regular}


\begin{table}[h]
\begin{center}
\caption{Benchmark \texttt{2024-acasxu-2023}} \label{tab:cat_acas}
{\setlength{\tabcolsep}{2pt}
\begin{tabular}[h]{@{}llllllrrr@{}}
\toprule
\textbf{\# ~} & \textbf{Tool} & \textbf{Verified} & \textbf{Falsified} & \textbf{Fastest} & \textbf{Penalty} & \textbf{Points} & \textbf{Score} & \textbf{Solved}\\
\midrule
1 & $\alpha$-$\beta$-CROWN & 139 & 47 & 0 & 0 & 1860 & 100.0 & 100.0\% \\
2 & nnenum & 139 & 46 & 0 & 0 & 1850 & 99.5 & 99.5\% \\
3 & PyRAT & 137 & 47 & 0 & 0 & 1840 & 98.9 & 98.9\% \\
4 & Marabou & 134 & 45 & 0 & 1 & 1640 & 88.2 & 96.2\% \\
5 & NeVer2 & 121 & 40 & 0 & 0 & 1610 & 86.6 & 86.6\% \\
6 & NNV & 70 & 27 & 0 & 0 & 970 & 52.2 & 52.2\% \\
7 & CORA & 134 & 7 & 0 & 36 & -3990 & 0 & 75.8\% \\
8 & NeuralSAT & 138 & 0 & 0 & 46 & -5520 & 0 & 74.2\% \\
\bottomrule
\end{tabular}
}
\end{center}
\end{table}

\begin{figure}[h]
\centerline{\includegraphics[width=\textwidth]{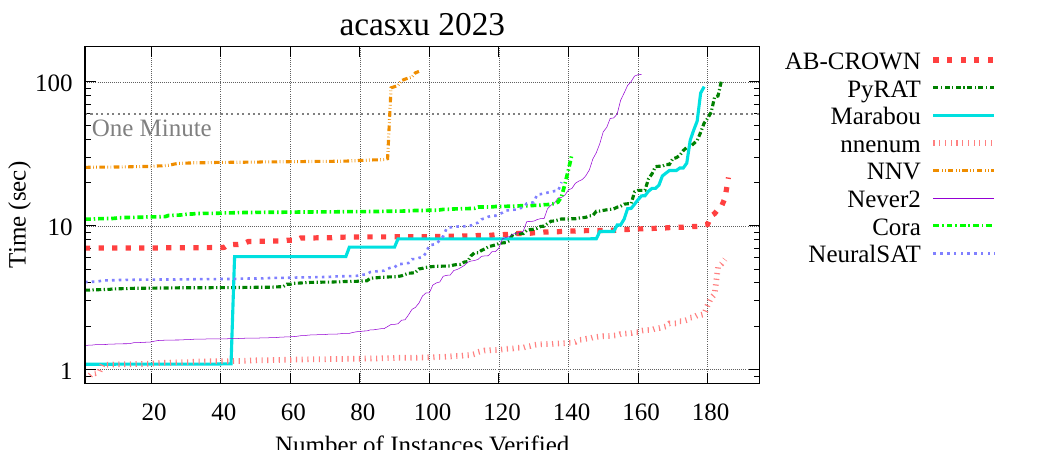}}
\caption{Cactus Plot for acasxu 2023.}
\label{fig:quantPic_acas}
\end{figure}

\clearpage

\begin{table}[h]
\begin{center}
\caption{Benchmark \texttt{2024-cgan-2023}} \label{tab:cat_cgan}
{\setlength{\tabcolsep}{2pt}
\begin{tabular}[h]{@{}llllllrrr@{}}
\toprule
\textbf{\# ~} & \textbf{Tool} & \textbf{Verified} & \textbf{Falsified} & \textbf{Fastest} & \textbf{Penalty} & \textbf{Points} & \textbf{Score} & \textbf{Solved}\\
\midrule
1 & PyRAT & 8 & 13 & 0 & 0 & 210 & 100.0 & 100.0\% \\
2 & $\alpha$-$\beta$-CROWN & 8 & 13 & 0 & 0 & 210 & 100.0 & 100.0\% \\
3 & nnenum & 6 & 11 & 0 & 0 & 170 & 81.0 & 81.0\% \\
4 & NNV & 6 & 11 & 0 & 0 & 170 & 81.0 & 81.0\% \\
5 & Marabou & 0 & 13 & 0 & 0 & 130 & 61.9 & 61.9\% \\
6 & NeuralSAT & 8 & 0 & 0 & 11 & -1570 & 0 & 38.1\% \\
\bottomrule
\end{tabular}
}
\end{center}
\end{table}

\begin{figure}[h]
\centerline{\includegraphics[width=\textwidth]{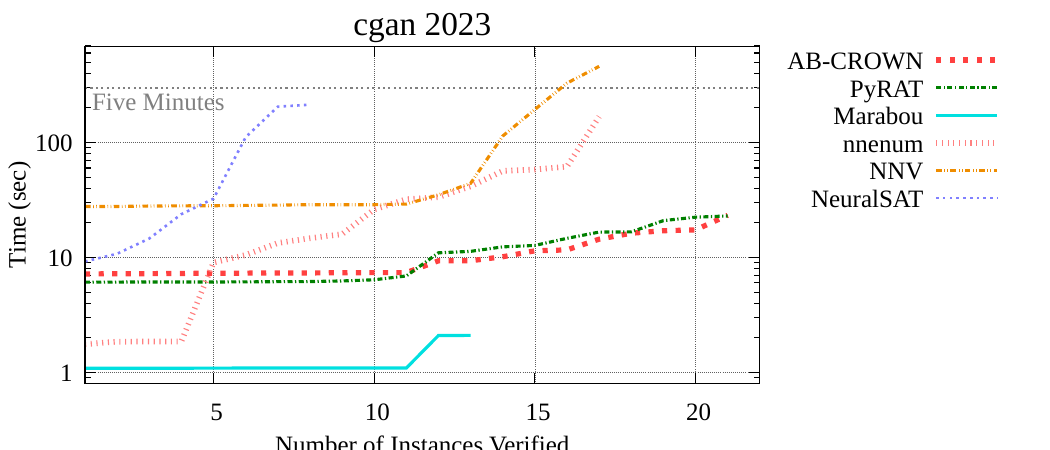}}
\caption{Cactus Plot for cgan 2023.}
\label{fig:quantPic_cgan}
\end{figure}

\clearpage

\begin{table}[h]
\begin{center}
\caption{Benchmark \texttt{2024-cifar100}} \label{tab:cat_cifar}
{\setlength{\tabcolsep}{2pt}
\begin{tabular}[h]{@{}llllllrrr@{}}
\toprule
\textbf{\# ~} & \textbf{Tool} & \textbf{Verified} & \textbf{Falsified} & \textbf{Fastest} & \textbf{Penalty} & \textbf{Points} & \textbf{Score} & \textbf{Solved}\\
\midrule
1 & $\alpha$-$\beta$-CROWN & 117 & 32 & 0 & 0 & 1490 & 100.0 & 74.5\% \\
2 & PyRAT & 67 & 25 & 0 & 0 & 920 & 61.7 & 46.0\% \\
3 & Marabou & 0 & 30 & 0 & 0 & 300 & 20.1 & 15.0\% \\
4 & NeuralSAT & 89 & 0 & 0 & 23 & -2560 & 0 & 44.5\% \\
\bottomrule
\end{tabular}
}
\end{center}
\end{table}

\begin{figure}[h]
\centerline{\includegraphics[width=\textwidth]{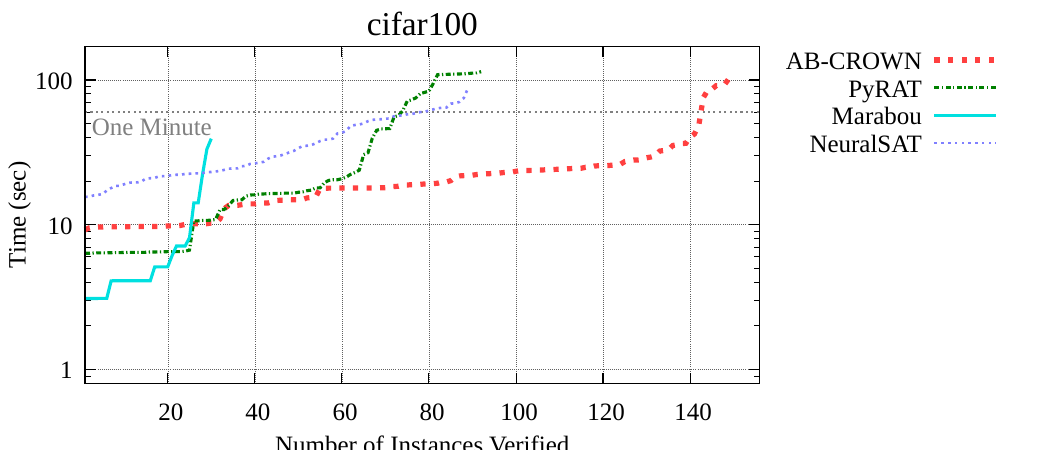}}
\caption{Cactus Plot for cifar100.}
\label{fig:quantPic_cifar}
\end{figure}

\clearpage

\begin{table}[h]
\begin{center}
\caption{Benchmark \texttt{2024-collins-rul-cnn-2023}} \label{tab:cat_collins_rul}
{\setlength{\tabcolsep}{2pt}
\begin{tabular}[h]{@{}llllllrrr@{}}
\toprule
\textbf{\# ~} & \textbf{Tool} & \textbf{Verified} & \textbf{Falsified} & \textbf{Fastest} & \textbf{Penalty} & \textbf{Points} & \textbf{Score} & \textbf{Solved}\\
\midrule
1 & nnenum & 30 & 32 & 0 & 0 & 620 & 100.0 & 100.0\% \\
2 & NNV & 30 & 32 & 0 & 0 & 620 & 100.0 & 100.0\% \\
3 & Marabou & 30 & 32 & 0 & 0 & 620 & 100.0 & 100.0\% \\
4 & $\alpha$-$\beta$-CROWN & 30 & 32 & 0 & 0 & 620 & 100.0 & 100.0\% \\
5 & PyRAT & 30 & 28 & 0 & 0 & 580 & 93.5 & 93.5\% \\
6 & CORA & 0 & 19 & 0 & 11 & -1460 & 0 & 30.6\% \\
7 & NeuralSAT & 30 & 0 & 0 & 32 & -4500 & 0 & 48.4\% \\
\bottomrule
\end{tabular}
}
\end{center}
\end{table}

\begin{figure}[h]
\centerline{\includegraphics[width=\textwidth]{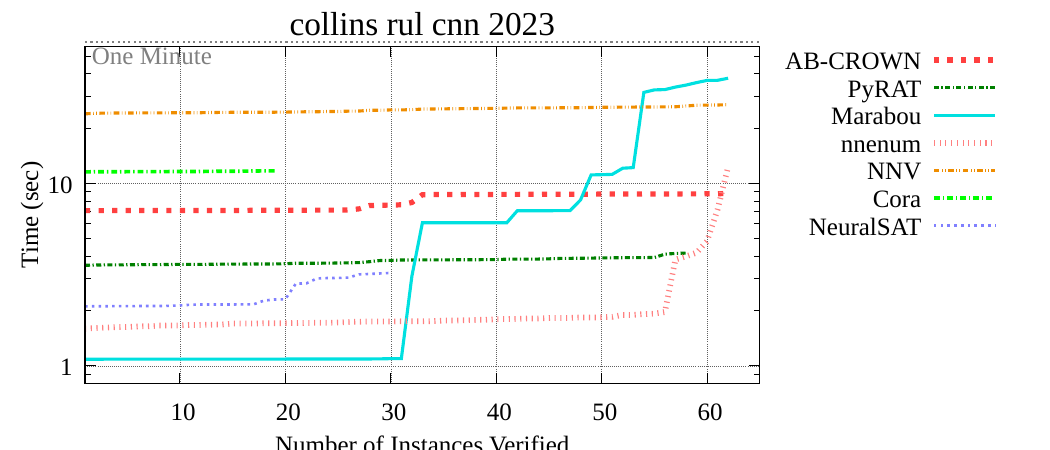}}
\caption{Cactus Plot for collins rul cnn 2023.}
\label{fig:quantPic_collins_rul}
\end{figure}

\clearpage

\begin{table}[h]
\begin{center}
\caption{Benchmark \texttt{2024-cora}} \label{tab:cat_cora}
{\setlength{\tabcolsep}{2pt}
\begin{tabular}[h]{@{}llllllrrr@{}}
\toprule
\textbf{\# ~} & \textbf{Tool} & \textbf{Verified} & \textbf{Falsified} & \textbf{Fastest} & \textbf{Penalty} & \textbf{Points} & \textbf{Score} & \textbf{Solved}\\
\midrule
1 & $\alpha$-$\beta$-CROWN & 24 & 134 & 0 & 0 & 1580 & 100.0 & 87.8\% \\
2 & Marabou & 22 & 134 & 0 & 0 & 1560 & 98.7 & 86.7\% \\
3 & PyRAT & 22 & 128 & 0 & 0 & 1500 & 94.9 & 83.3\% \\
4 & NNV & 15 & 47 & 0 & 0 & 620 & 39.2 & 34.4\% \\
5 & NeVer2 & 17 & 11 & 0 & 0 & 280 & 17.7 & 15.6\% \\
6 & nnenum & 20 & 6 & 0 & 0 & 260 & 16.5 & 14.4\% \\
7 & CORA & 21 & 5 & 0 & 54 & -7840 & 0 & 14.4\% \\
8 & NeuralSAT & 23 & 0 & 0 & 134 & -19870 & 0 & 12.8\% \\
\bottomrule
\end{tabular}
}
\end{center}
\end{table}

\begin{figure}[h]
\centerline{\includegraphics[width=\textwidth]{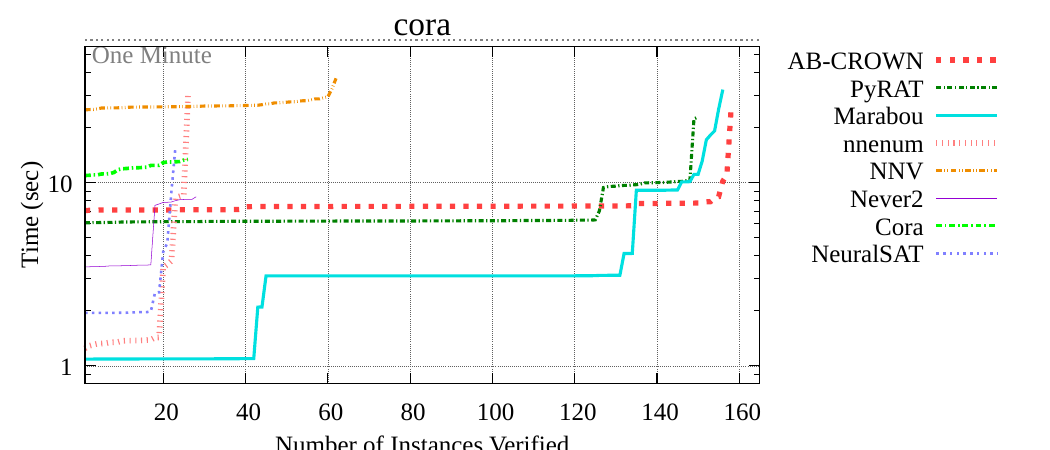}}
\caption{Cactus Plot for cora.}
\label{fig:quantPic_cora}
\end{figure}

\clearpage

\begin{table}[h]
\begin{center}
\caption{Benchmark \texttt{2024-dist-shift-2023}} \label{tab:cat_dist}
{\setlength{\tabcolsep}{2pt}
\begin{tabular}[h]{@{}llllllrrr@{}}
\toprule
\textbf{\# ~} & \textbf{Tool} & \textbf{Verified} & \textbf{Falsified} & \textbf{Fastest} & \textbf{Penalty} & \textbf{Points} & \textbf{Score} & \textbf{Solved}\\
\midrule
1 & PyRAT & 63 & 8 & 0 & 0 & 710 & 100.0 & 98.6\% \\
2 & CORA & 63 & 8 & 0 & 0 & 710 & 100.0 & 98.6\% \\
3 & $\alpha$-$\beta$-CROWN & 63 & 8 & 0 & 0 & 710 & 100.0 & 98.6\% \\
4 & Marabou & 62 & 7 & 0 & 0 & 690 & 97.2 & 95.8\% \\
5 & NNV & 51 & 5 & 0 & 0 & 560 & 78.9 & 77.8\% \\
\bottomrule
\end{tabular}
}
\end{center}
\end{table}

\begin{figure}[h]
\centerline{\includegraphics[width=\textwidth]{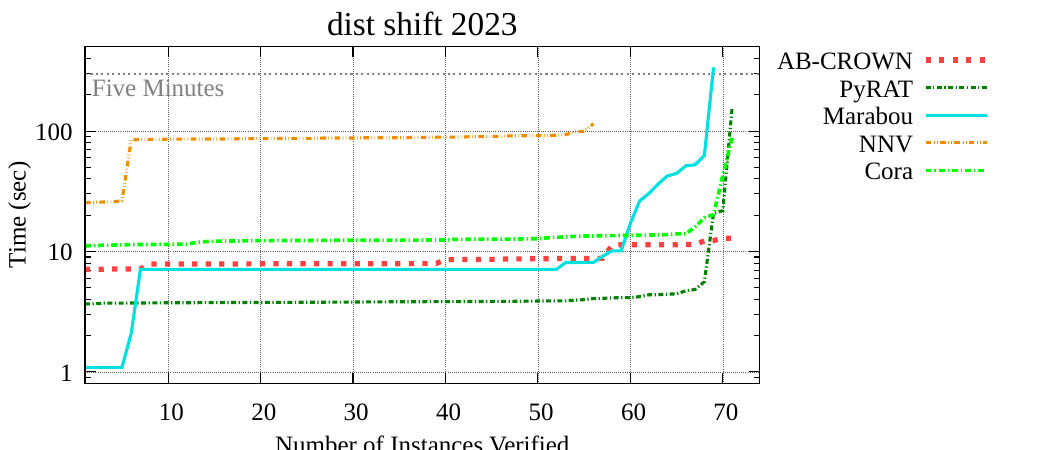}}
\caption{Cactus Plot for dist shift 2023.}
\label{fig:quantPic_dist}
\end{figure}

\clearpage

\begin{table}[h]
\begin{center}
\caption{Benchmark \texttt{2024-linearizenn}} \label{tab:cat_linearize}
{\setlength{\tabcolsep}{2pt}
\begin{tabular}[h]{@{}llllllrrr@{}}
\toprule
\textbf{\# ~} & \textbf{Tool} & \textbf{Verified} & \textbf{Falsified} & \textbf{Fastest} & \textbf{Penalty} & \textbf{Points} & \textbf{Score} & \textbf{Solved}\\
\midrule
1 & PyRAT & 59 & 1 & 0 & 0 & 600 & 100.0 & 100.0\% \\
2 & Marabou & 59 & 1 & 0 & 0 & 600 & 100.0 & 100.0\% \\
3 & $\alpha$-$\beta$-CROWN & 59 & 1 & 0 & 0 & 600 & 100.0 & 100.0\% \\
4 & nnenum & 59 & 0 & 0 & 1 & 440 & 73.3 & 98.3\% \\
\bottomrule
\end{tabular}
}
\end{center}
\end{table}

\begin{figure}[h]
\centerline{\includegraphics[width=\textwidth]{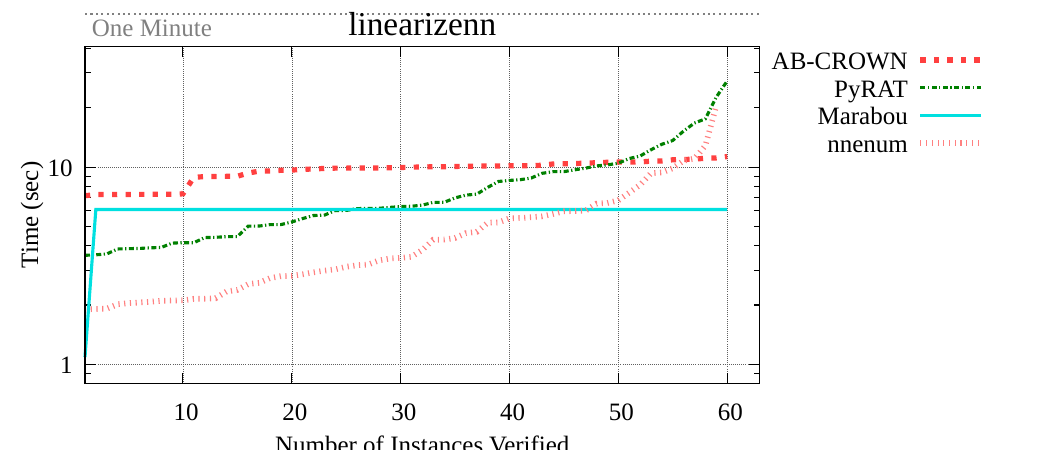}}
\caption{Cactus Plot for linearizenn.}
\label{fig:quantPic_linearize}
\end{figure}

\clearpage

\begin{table}[h]
\begin{center}
\caption{Benchmark \texttt{2024-metaroom-2023}} \label{tab:cat_metaroom}
{\setlength{\tabcolsep}{2pt}
\begin{tabular}[h]{@{}llllllrrr@{}}
\toprule
\textbf{\# ~} & \textbf{Tool} & \textbf{Verified} & \textbf{Falsified} & \textbf{Fastest} & \textbf{Penalty} & \textbf{Points} & \textbf{Score} & \textbf{Solved}\\
\midrule
1 & $\alpha$-$\beta$-CROWN & 91 & 7 & 0 & 0 & 980 & 100.0 & 98.0\% \\
2 & PyRAT & 91 & 6 & 0 & 0 & 970 & 99.0 & 97.0\% \\
3 & NNV & 90 & 2 & 0 & 0 & 920 & 93.9 & 92.0\% \\
4 & Marabou & 46 & 7 & 0 & 0 & 530 & 54.1 & 53.0\% \\
5 & nnenum & 44 & 2 & 0 & 0 & 460 & 46.9 & 46.0\% \\
6 & NeuralSAT & 91 & 0 & 0 & 7 & -140 & 0 & 91.0\% \\
\bottomrule
\end{tabular}
}
\end{center}
\end{table}

\begin{figure}[h]
\centerline{\includegraphics[width=\textwidth]{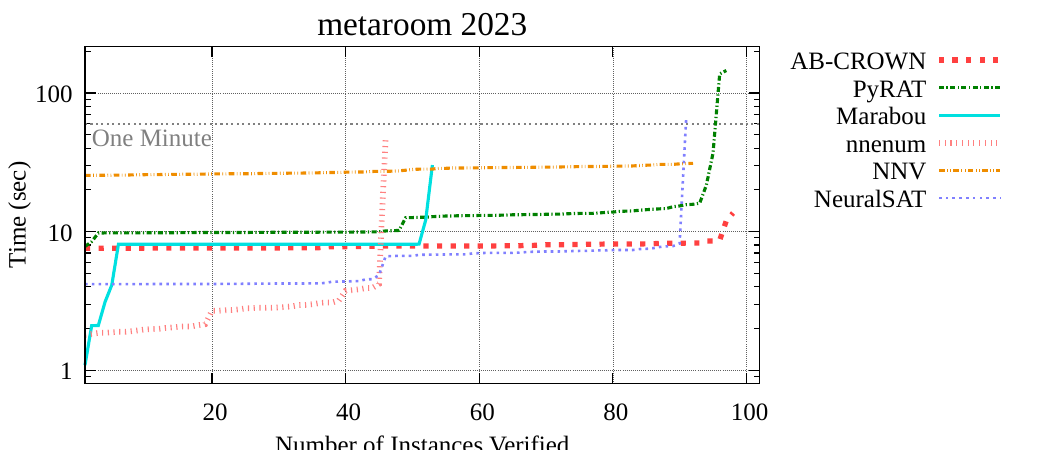}}
\caption{Cactus Plot for metaroom 2023.}
\label{fig:quantPic_metaroom}
\end{figure}

\clearpage

\begin{table}[h]
\begin{center}
\caption{Benchmark \texttt{2024-nn4sys-2023}} \label{tab:cat_nn4sys}
{\setlength{\tabcolsep}{2pt}
\begin{tabular}[h]{@{}llllllrrr@{}}
\toprule
\textbf{\# ~} & \textbf{Tool} & \textbf{Verified} & \textbf{Falsified} & \textbf{Fastest} & \textbf{Penalty} & \textbf{Points} & \textbf{Score} & \textbf{Solved}\\
\midrule
1 & $\alpha$-$\beta$-CROWN & 194 & 0 & 0 & 0 & 1940 & 100.0 & 100.0\% \\
2 & PyRAT & 53 & 0 & 0 & 0 & 530 & 27.3 & 27.3\% \\
3 & Marabou & 24 & 0 & 0 & 0 & 240 & 12.4 & 12.4\% \\
4 & nnenum & 22 & 0 & 0 & 0 & 220 & 11.3 & 11.3\% \\
5 & CORA & 2 & 0 & 0 & 0 & 20 & 1.0 & 1.0\% \\
\bottomrule
\end{tabular}
}
\end{center}
\end{table}

\begin{figure}[h]
\centerline{\includegraphics[width=\textwidth]{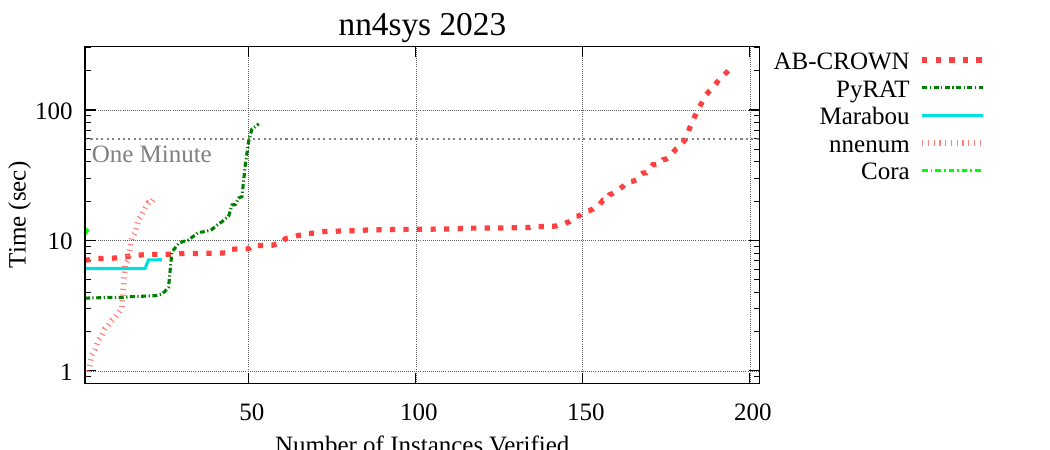}}
\caption{Cactus Plot for nn4sys 2023.}
\label{fig:quantPic_nn4sys}
\end{figure}

\clearpage

\begin{table}[h]
\begin{center}
\caption{Benchmark \texttt{2024-safenlp}} \label{tab:cat_safenlp}
{\setlength{\tabcolsep}{2pt}
\begin{tabular}[h]{@{}llllllrrr@{}}
\toprule
\textbf{\# ~} & \textbf{Tool} & \textbf{Verified} & \textbf{Falsified} & \textbf{Fastest} & \textbf{Penalty} & \textbf{Points} & \textbf{Score} & \textbf{Solved}\\
\midrule
1 & $\alpha$-$\beta$-CROWN & 421 & 659 & 0 & 0 & 10800 & 100.0 & 100.0\% \\
2 & nnenum & 321 & 642 & 0 & 1 & 9480 & 87.8 & 89.2\% \\
3 & PyRAT & 277 & 586 & 0 & 0 & 8630 & 79.9 & 79.9\% \\
4 & NeVer2 & 161 & 466 & 0 & 0 & 6270 & 58.1 & 58.1\% \\
5 & CORA & 266 & 269 & 0 & 0 & 5350 & 49.5 & 49.5\% \\
6 & NNV & 166 & 165 & 0 & 0 & 3310 & 30.6 & 30.6\% \\
7 & Marabou & 300 & 375 & 0 & 402 & -53550 & 0 & 62.5\% \\
\bottomrule
\end{tabular}
}
\end{center}
\end{table}

\begin{figure}[h]
\centerline{\includegraphics[width=\textwidth]{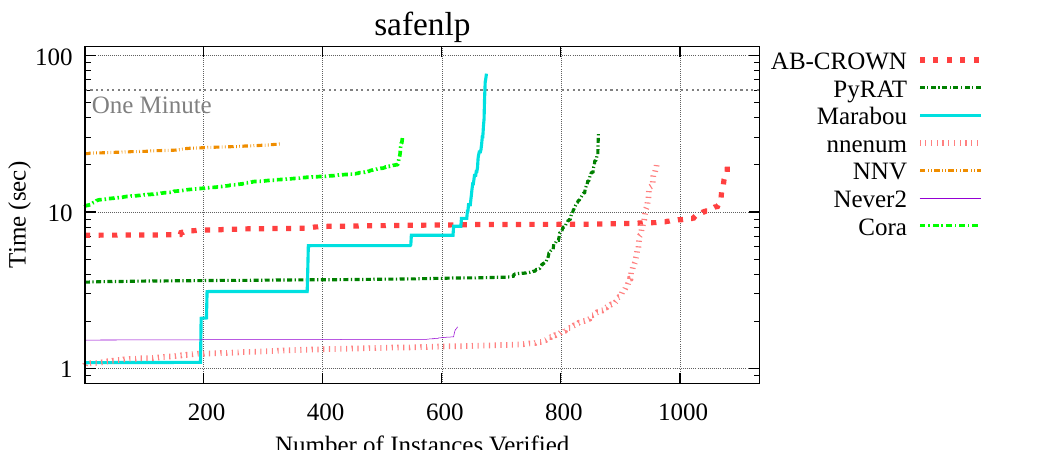}}
\caption{Cactus Plot for safenlp.}
\label{fig:quantPic_safenlp}
\end{figure}

\clearpage

\begin{table}[h]
\begin{center}
\caption{Benchmark \texttt{2024-tinyimagenet}} \label{tab:cat_tinyimagenet}
{\setlength{\tabcolsep}{2pt}
\begin{tabular}[h]{@{}llllllrrr@{}}
\toprule
\textbf{\# ~} & \textbf{Tool} & \textbf{Verified} & \textbf{Falsified} & \textbf{Fastest} & \textbf{Penalty} & \textbf{Points} & \textbf{Score} & \textbf{Solved}\\
\midrule
1 & $\alpha$-$\beta$-CROWN & 131 & 43 & 0 & 0 & 1740 & 100.0 & 87.0\% \\
2 & PyRAT & 48 & 31 & 0 & 0 & 790 & 45.4 & 39.5\% \\
3 & Marabou & 0 & 43 & 0 & 0 & 430 & 24.7 & 21.5\% \\
4 & NNV & 0 & 2 & 0 & 0 & 20 & 1.1 & 1.0\% \\
5 & CORA & 0 & 2 & 0 & 0 & 20 & 1.1 & 1.0\% \\
6 & NeuralSAT & 95 & 0 & 0 & 11 & -700 & 0 & 47.5\% \\
\bottomrule
\end{tabular}
}
\end{center}
\end{table}

\begin{figure}[h]
\centerline{\includegraphics[width=\textwidth]{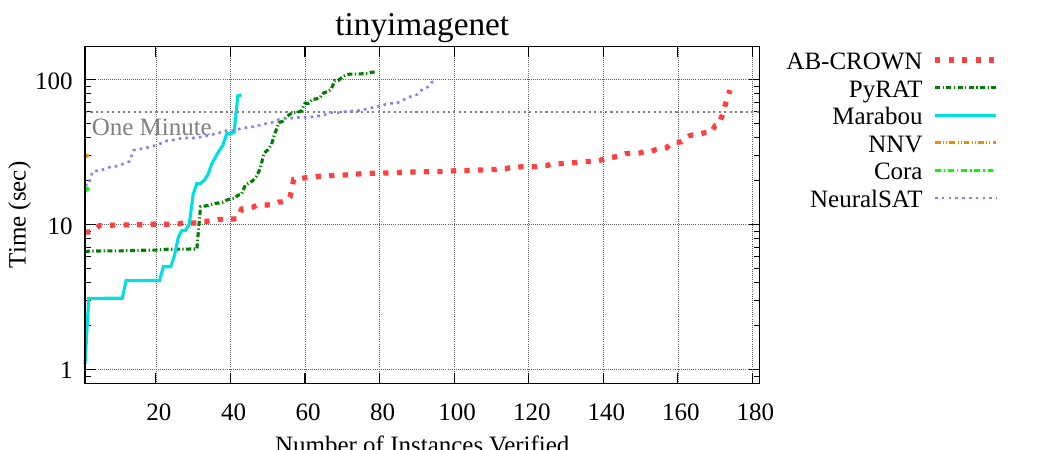}}
\caption{Cactus Plot for tinyimagenet.}
\label{fig:quantPic_tinyimagenet}
\end{figure}

\begin{table}[h]
\begin{center}
\caption{Benchmark \texttt{2024-tllverifybench-2023}} \label{tab:cat_tllverify}
{\setlength{\tabcolsep}{2pt}
\begin{tabular}[h]{@{}llllllrrr@{}}
\toprule
\textbf{\# ~} & \textbf{Tool} & \textbf{Verified} & \textbf{Falsified} & \textbf{Fastest} & \textbf{Penalty} & \textbf{Points} & \textbf{Score} & \textbf{Solved}\\
\midrule
1 & PyRAT & 15 & 17 & 0 & 0 & 320 & 100.0 & 100.0\% \\
2 & NeVer2 & 23 & 9 & 0 & 0 & 320 & 100.0 & 100.0\% \\
3 & CORA & 15 & 17 & 0 & 0 & 320 & 100.0 & 100.0\% \\
4 & $\alpha$-$\beta$-CROWN & 15 & 17 & 0 & 0 & 320 & 100.0 & 100.0\% \\
5 & Marabou & 13 & 17 & 0 & 0 & 300 & 93.8 & 93.8\% \\
6 & nnenum & 2 & 16 & 0 & 0 & 180 & 56.2 & 56.2\% \\
7 & NNV & 0 & 17 & 0 & 0 & 170 & 53.1 & 53.1\% \\
8 & NeuralSAT & 15 & 0 & 0 & 17 & -2400 & 0 & 46.9\% \\
\bottomrule
\end{tabular}
}
\end{center}
\end{table}

\begin{figure}[h]
\centerline{\includegraphics[width=\textwidth]{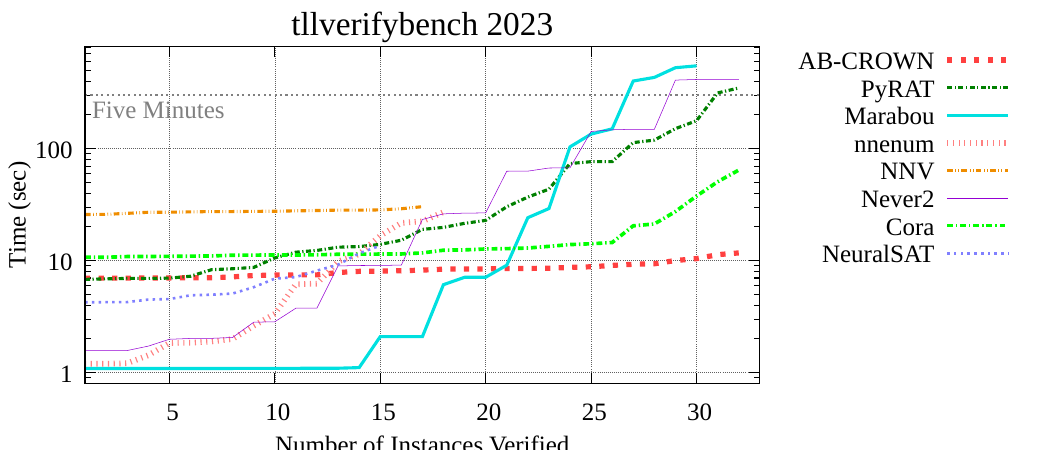}}
\caption{Cactus Plot for tllverifybench 2023.}
\label{fig:quantPic_tllverifybench}
\end{figure}

\clearpage
\subsection{Extended Track}
\label{sec:benchmark_results_extended}

\begin{table}[h]
\begin{center}
\caption{Benchmark \texttt{2024-cctsdb-yolo-2023}} \label{tab:cat_cctsdb}
{\setlength{\tabcolsep}{2pt}
\begin{tabular}[h]{@{}llllllrrr@{}}
\toprule
\textbf{\# ~} & \textbf{Tool} & \textbf{Verified} & \textbf{Falsified} & \textbf{Fastest} & \textbf{Penalty} & \textbf{Points} & \textbf{Score} & \textbf{Solved}\\
\midrule
1 & $\alpha$-$\beta$-CROWN & 11 & 28 & 0 & 0 & 390 & 100.0 & 100.0\% \\
2 & PyRAT & 0 & 2 & 0 & 0 & 20 & 5.1 & 5.1\% \\
\bottomrule
\end{tabular}
}
\end{center}
\end{table}

\begin{figure}[h]
\centerline{\includegraphics[width=\textwidth]{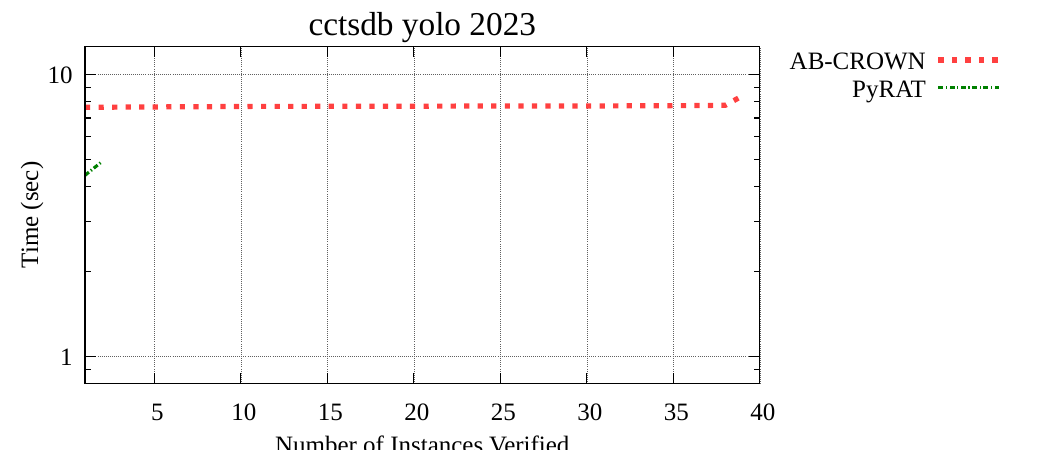}}
\caption{Cactus Plot for cctsdb yolo 2023.}
\label{fig:quantPic_cctsdb}
\end{figure}

\clearpage

\begin{table}[h]
\begin{center}
\caption{Benchmark \texttt{2024-collins-aerospace-benchmark}} \label{tab:cat_collins}
{\setlength{\tabcolsep}{2pt}
\begin{tabular}[h]{@{}llllllrrr@{}}
\toprule
\textbf{\# ~} & \textbf{Tool} & \textbf{Verified} & \textbf{Falsified} & \textbf{Fastest} & \textbf{Penalty} & \textbf{Points} & \textbf{Score} & \textbf{Solved}\\
\midrule
1 & PyRAT & 0 & 6 & 0 & 0 & 60 & 100.0 & 100.0\% \\
2 & $\alpha$-$\beta$-CROWN & 0 & 6 & 0 & 0 & 60 & 100.0 & 100.0\% \\
\bottomrule
\end{tabular}
}
\end{center}
\end{table}

\begin{figure}[h]
\centerline{\includegraphics[width=\textwidth]{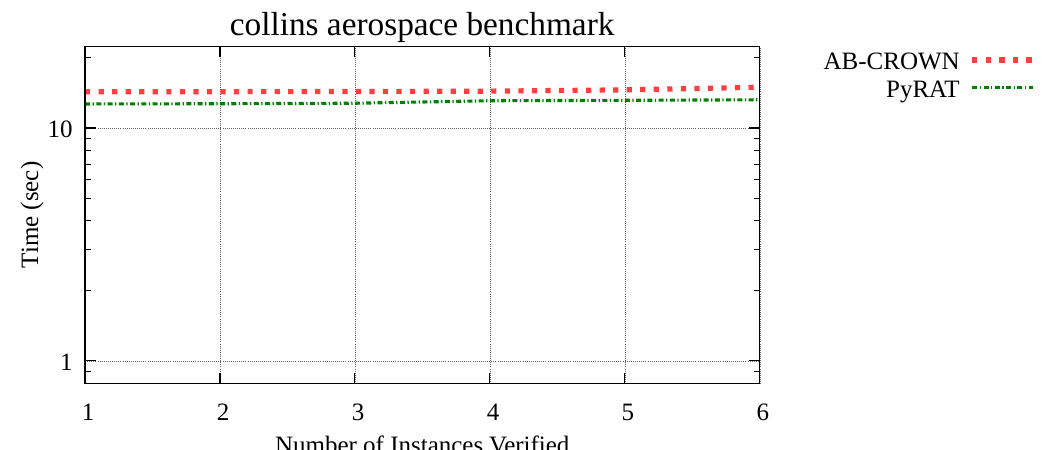}}
\caption{Cactus Plot for collins aerospace benchmark.}
\label{fig:quantPic_collins}
\end{figure}

\clearpage

\begin{table}[h]
\begin{center}
\caption{Benchmark \texttt{2024-lsnc}} \label{tab:cat_lsnc}
{\setlength{\tabcolsep}{2pt}
\begin{tabular}[h]{@{}llllllrrr@{}}
\toprule
\textbf{\# ~} & \textbf{Tool} & \textbf{Verified} & \textbf{Falsified} & \textbf{Fastest} & \textbf{Penalty} & \textbf{Points} & \textbf{Score} & \textbf{Solved}\\
\midrule
1 & $\alpha$-$\beta$-CROWN & 40 & 0 & 0 & 0 & 400 & 100.0 & 100.0\% \\
2 & PyRAT & 15 & 0 & 0 & 0 & 150 & 37.5 & 37.5\% \\
\bottomrule
\end{tabular}
}
\end{center}
\end{table}

\begin{figure}[h]
\centerline{\includegraphics[width=\textwidth]{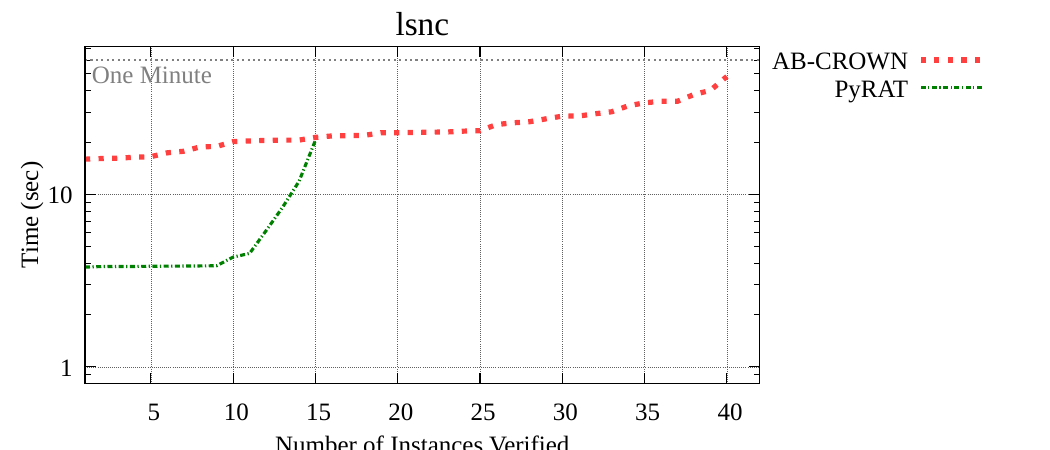}}
\caption{Cactus Plot for lsnc.}
\label{fig:quantPic_lsnc}
\end{figure}

\clearpage

\begin{table}[h]
\begin{center}
\caption{Benchmark \texttt{2024-ml4acopf-2023}} \label{tab:cat_ml4acopf}
{\setlength{\tabcolsep}{2pt}
\begin{tabular}[h]{@{}llllllrrr@{}}
\toprule
\textbf{\# ~} & \textbf{Tool} & \textbf{Verified} & \textbf{Falsified} & \textbf{Fastest} & \textbf{Penalty} & \textbf{Points} & \textbf{Score} & \textbf{Solved}\\
\midrule
1 & $\alpha$-$\beta$-CROWN & 22 & 0 & 0 & 0 & 220 & 100.0 & 95.7\% \\
2 & PyRAT & 15 & 0 & 0 & 0 & 150 & 68.2 & 65.2\% \\
\bottomrule
\end{tabular}
}
\end{center}
\end{table}

\begin{figure}[h]
\centerline{\includegraphics[width=\textwidth]{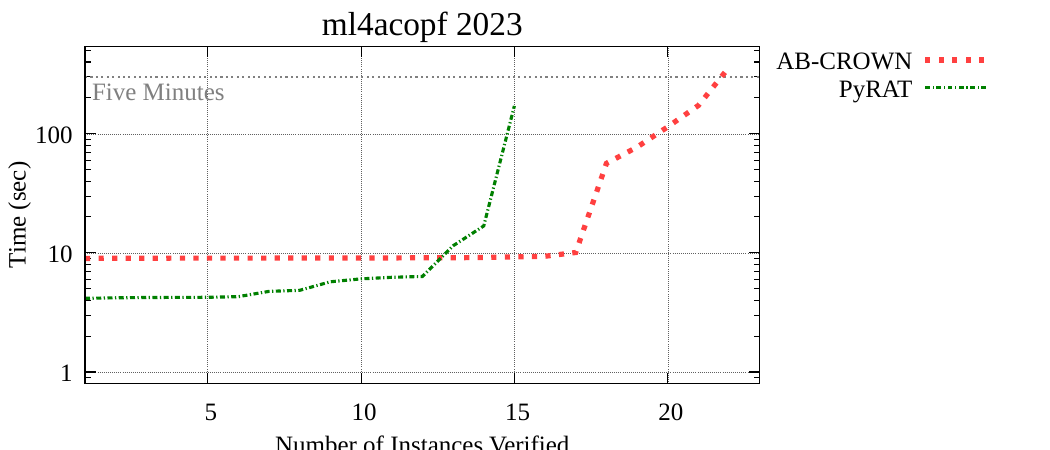}}
\caption{Cactus Plot for ml4acopf 2023.}
\label{fig:quantPic_ml4acopf}
\end{figure}

\clearpage

\begin{table}[h]
\begin{center}
\caption{Benchmark \texttt{2024-ml4acopf-2024}} \label{tab:cat_ml4acopf2024}
{\setlength{\tabcolsep}{2pt}
\begin{tabular}[h]{@{}llllllrrr@{}}
\toprule
\textbf{\# ~} & \textbf{Tool} & \textbf{Verified} & \textbf{Falsified} & \textbf{Fastest} & \textbf{Penalty} & \textbf{Points} & \textbf{Score} & \textbf{Solved}\\
\midrule
1 & $\alpha$-$\beta$-CROWN & 59 & 3 & 0 & 0 & 620 & 100.0 & 89.9\% \\
2 & PyRAT & 29 & 3 & 0 & 0 & 320 & 51.6 & 46.4\% \\
\bottomrule
\end{tabular}
}
\end{center}
\end{table}

\begin{figure}[h]
\centerline{\includegraphics[width=\textwidth]{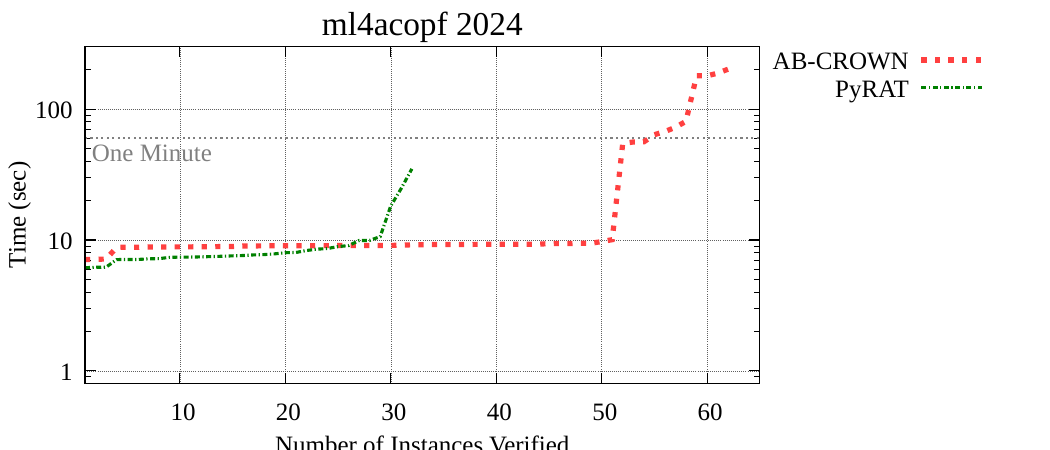}}
\caption{Cactus Plot for ml4acopf 2024.}
\label{fig:quantPic_ml4acopf_2024}
\end{figure}

\clearpage

\begin{table}[h]
\begin{center}
\caption{Benchmark \texttt{2024-traffic-signs-recognition-2023}} \label{tab:cat_traffic}
{\setlength{\tabcolsep}{2pt}
\begin{tabular}[h]{@{}llllllrrr@{}}
\toprule
\textbf{\# ~} & \textbf{Tool} & \textbf{Verified} & \textbf{Falsified} & \textbf{Fastest} & \textbf{Penalty} & \textbf{Points} & \textbf{Score} & \textbf{Solved}\\
\midrule
1 & $\alpha$-$\beta$-CROWN & 0 & 45 & 0 & 0 & 450 & 100.0 & 100.0\% \\
2 & PyRAT & 0 & 4 & 0 & 0 & 40 & 8.9 & 8.9\% \\
3 & NeuralSAT & 0 & 0 & 0 & 11 & -1650 & 0 & 0.0\% \\
\bottomrule
\end{tabular}
}
\end{center}
\end{table}

\begin{figure}[h]
\centerline{\includegraphics[width=\textwidth]{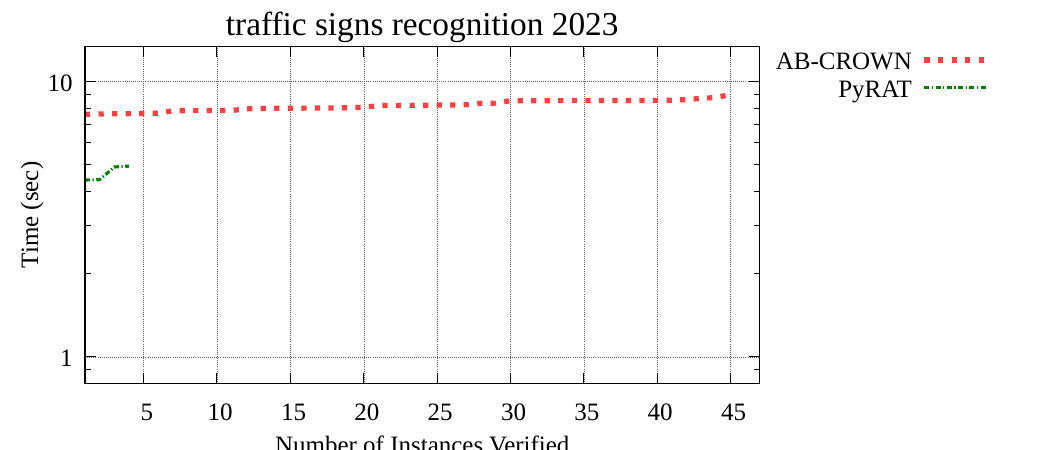}}
\caption{Cactus Plot for traffic signs recognition 2023.}
\label{fig:quantPic_traffic}
\end{figure}

\clearpage

\begin{table}[h]
\begin{center}
\caption{Benchmark \texttt{2024-vggnet16-2023}} \label{tab:cat_vggnet16}
{\setlength{\tabcolsep}{2pt}
\begin{tabular}[h]{@{}llllllrrr@{}}
\toprule
\textbf{\# ~} & \textbf{Tool} & \textbf{Verified} & \textbf{Falsified} & \textbf{Fastest} & \textbf{Penalty} & \textbf{Points} & \textbf{Score} & \textbf{Solved}\\
\midrule
1 & $\alpha$-$\beta$-CROWN & 18 & 0 & 0 & 0 & 180 & 100.0 & 100.0\% \\
2 & nnenum & 13 & 0 & 0 & 0 & 130 & 72.2 & 72.2\% \\
3 & PyRAT & 10 & 0 & 0 & 0 & 100 & 55.6 & 55.6\% \\
4 & NeuralSAT & 6 & 0 & 0 & 0 & 60 & 33.3 & 33.3\% \\
\bottomrule
\end{tabular}
}
\end{center}
\end{table}

\begin{figure}[h]
\centerline{\includegraphics[width=\textwidth]{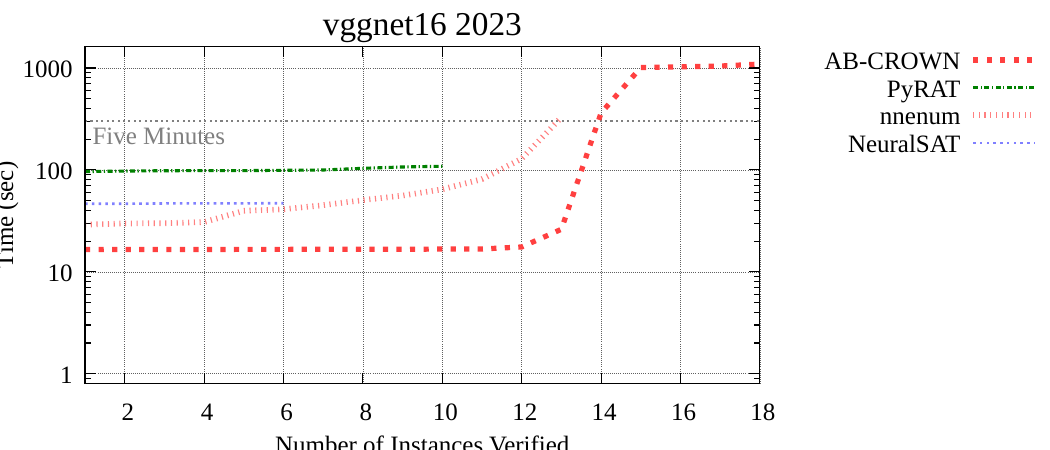}}
\caption{Cactus Plot for vggnet16 2023.}
\label{fig:quantPic_vggnet16}
\end{figure}

\clearpage

\begin{table}[h]
\begin{center}
\caption{Benchmark \texttt{2024-vit-2023}} \label{tab:cat_vit}
{\setlength{\tabcolsep}{2pt}
\begin{tabular}[h]{@{}llllllrrr@{}}
\toprule
\textbf{\# ~} & \textbf{Tool} & \textbf{Verified} & \textbf{Falsified} & \textbf{Fastest} & \textbf{Penalty} & \textbf{Points} & \textbf{Score} & \textbf{Solved}\\
\midrule
1 & $\alpha$-$\beta$-CROWN & 84 & 0 & 0 & 0 & 840 & 100.0 & 42.0\% \\
2 & NeuralSAT & 11 & 0 & 0 & 0 & 110 & 13.1 & 5.5\% \\
\bottomrule
\end{tabular}
}
\end{center}
\end{table}

\begin{figure}[h]
\centerline{\includegraphics[width=\textwidth]{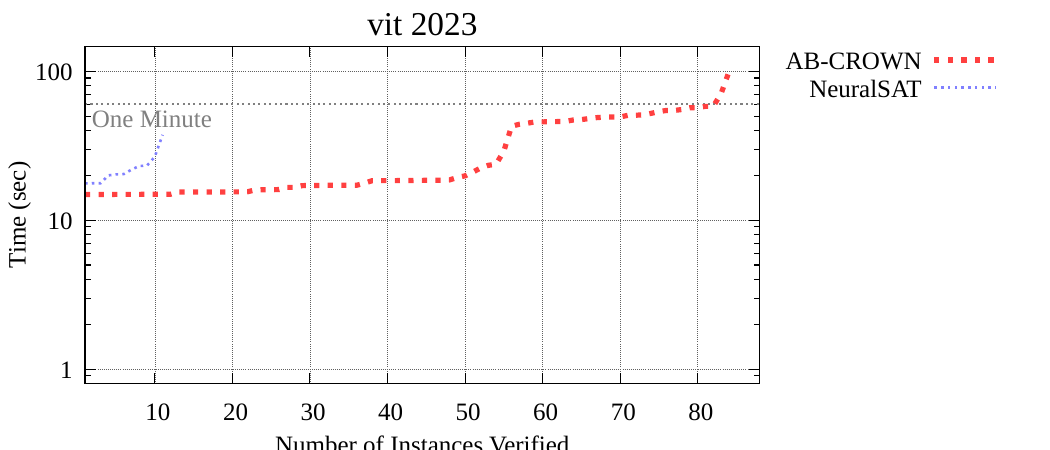}}
\caption{Cactus Plot for vit 2023.}
\label{fig:quantPic_vit}
\end{figure}

\clearpage

\begin{table}[h]
\begin{center}
\caption{Benchmark \texttt{2024-yolo-2023}} \label{tab:cat_yolo}
{\setlength{\tabcolsep}{2pt}
\begin{tabular}[h]{@{}llllllrrr@{}}
\toprule
\textbf{\# ~} & \textbf{Tool} & \textbf{Verified} & \textbf{Falsified} & \textbf{Fastest} & \textbf{Penalty} & \textbf{Points} & \textbf{Score} & \textbf{Solved}\\
\midrule
1 & $\alpha$-$\beta$-CROWN & 60 & 0 & 0 & 0 & 600 & 100.0 & 83.3\% \\
2 & PyRAT & 43 & 0 & 0 & 0 & 430 & 71.7 & 59.7\% \\
\bottomrule
\end{tabular}
}
\end{center}
\end{table}

\begin{figure}[h]
\centerline{\includegraphics[width=\textwidth]{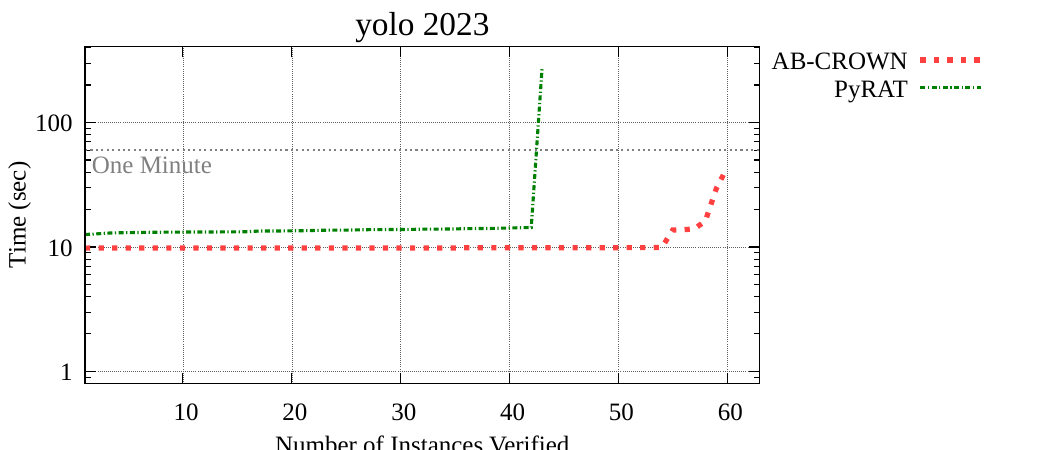}}
\caption{Cactus Plot for yolo 2023.}
\label{fig:quantPic_yolo}
\end{figure}

\clearpage
\section{Results After Tool Updates}
\label{sec:alternative_ranking}
As mentioned in Section~\ref{sec:results}, the two tools NeuralSAT and CORA provided the results in a way that differed from the expected output format.
Even though this difference was minor, it caused the evaluation to penalize them, even though they were able to correctly identify both SAT and UNSAT instances.

While the expected format was described in the VNN-COMP rules, in subsequent VNN-COMP instances, we plan to address this as follows to avoid this issue.
First, we plan to release more details regarding the specification of the format tools should use to report SAT instances and counterexamples.
Second, we plan to execute the aggregate scoring code at submission time, instead of post hoc: tool authors did not detect that there were problems at submission time, as the returned instances as SAT or UNSAT appeared correct, but were aggregated when executing the scoring scripts later for preparation of the scores.
Third, we plan to add an error handling mechanism in the scoring script, so that in the event of a format problem or parsing problem of the results files from the tools, we do not count these as unsound (i.e., wrong) results, but rather count them as an error and exclude results (so just score them as $0$ instead of a negative penalty).
There are some pros/cons of this with respect to scoring, so we will get participant feedback in the next iteration when discussing the scoring mechanisms.

After fixing these issues, the following results were achieved.
\subsection{Regular Track}


\begin{table}[h]
\begin{center}
\caption{Overall Score} \label{tab:score_new}
{\setlength{\tabcolsep}{2pt}
\begin{tabular}[h]{@{}lll@{}}
\toprule
\textbf{\# ~} & \textbf{Tool} & \textbf{Score}\\
\midrule
1 & $\alpha$-$\beta$-CROWN & 1200.0 \\
2 & NeuralSAT & 1113.1 \\
3 & PyRAT & 1000.8 \\
4 & Marabou & 751.0 \\
5 & nnenum & 572.5 \\
6 & NNV & 530.0 \\
7 & CORA & 439.5 \\
8 & NeVer2 & 262.3 \\
\bottomrule
\end{tabular}
}
\end{center}
\end{table}

\begin{figure}[h]
\centerline{\includegraphics[width=\textwidth]{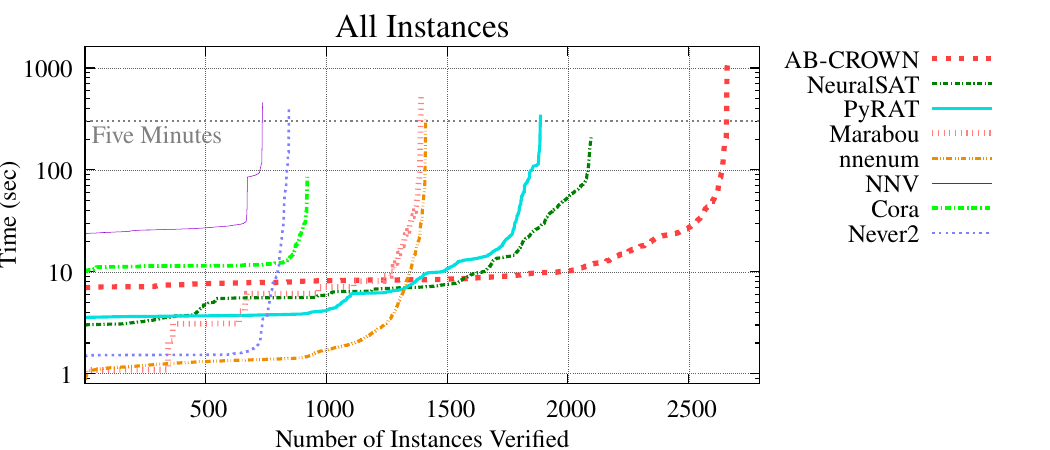}}
\caption{Cactus Plot for All Instances (Regular Track).}
\label{fig:quantPic_new}
\end{figure}

\subsection{Extended Track}


\begin{table}[H]
\begin{center}
\caption{Overall Score} \label{tab:score_extended_new}
{\setlength{\tabcolsep}{2pt}
\begin{tabular}[h]{@{}lll@{}}
\toprule
\textbf{\# ~} & \textbf{Tool} & \textbf{Score}\\
\midrule
1 & $\alpha$-$\beta$-CROWN & 900.0 \\
2 & PyRAT & 398.5 \\
3 & nnenum & 72.2 \\
4 & NeuralSAT & 70.9 \\
\bottomrule
\end{tabular}
}
\end{center}
\end{table}

\begin{figure}[H]
\centerline{\includegraphics[width=\textwidth]{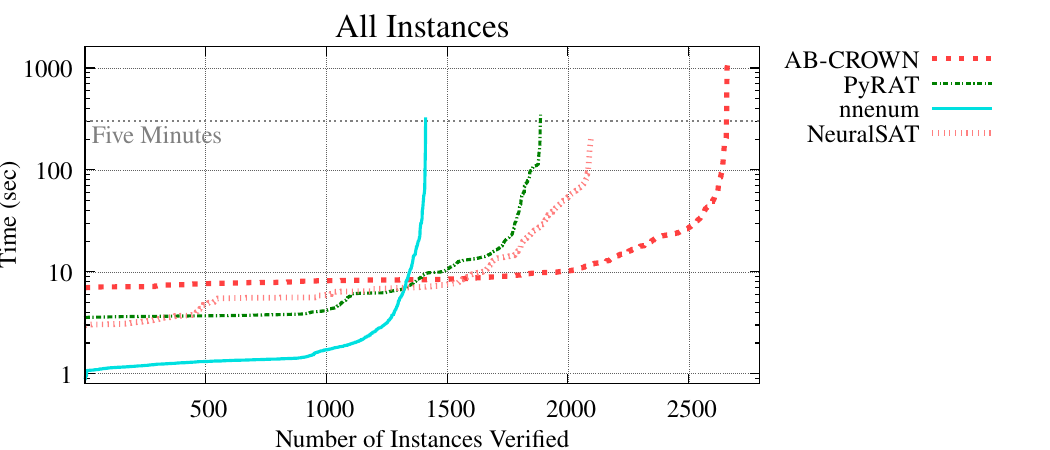}}
\caption{Cactus Plot for All Instances (Extended Track).
}
\label{fig:quantPic_extended_new}
\end{figure}

\subsection{More Results}
The full list of results for this setting can be generated using the scripts in \url{https://github.com/ChristopherBrix/vnncomp2024_results}.

\clearpage
\subsection{Detailed Results}
\label{sec:results_detailed}

\begin{center}
{\setlength{\tabcolsep}{1pt}
\scriptsize
\begin{longtable}{@{}lllllllllll@{}}
\caption{\footnotesize Regular Track Instance Runtimes. Fastest times are \textcolor{blue}{blue}. Second fastest are \textcolor{second}{green}. Penalties are red crosses (\textbf{\textcolor{red}{\ding{55}}}).} \label{tab:all_results} \\
\toprule
\textbf{Category} & \textbf{Id} & \textbf{Result} & \textbf{$\alpha$-$\beta$-C} & \textbf{PyRAT} & \textbf{Marab} & \textbf{NNen} & \textbf{NNV} & \textbf{NeVer2} & \textbf{CORA} & \textbf{NSAT} \\
\midrule
\endhead
Acasxu 2023 & 0 & ~\textsc{unsat} & \textcolor{darkgray}{9.87} & \textcolor{darkgray}{7.87} & \textcolor{darkgray}{8.11} & \textcolor{darkgray}{1.51} & - & \textcolor{darkgray}{3.42} & \textcolor{darkgray}{12.1} & \textcolor{darkgray}{4.61} \\
Acasxu 2023 & 1 & ~\textsc{unsat} & \textcolor{darkgray}{9.03} & \textcolor{darkgray}{9.92} & \textcolor{darkgray}{8.10} & \textcolor{darkgray}{1.42} & - & \textcolor{darkgray}{4.00} & \textcolor{darkgray}{11.1} & \textcolor{darkgray}{4.46} \\
Acasxu 2023 & 2 & ~\textsc{unsat} & \textcolor{darkgray}{9.24} & \textcolor{darkgray}{11.4} & \textcolor{darkgray}{8.10} & \textcolor{darkgray}{1.71} & - & \textcolor{darkgray}{10.7} & \textcolor{darkgray}{11.1} & \textcolor{darkgray}{9.90} \\
Acasxu 2023 & 3 & ~\textsc{unsat} & \textcolor{darkgray}{9.25} & \textcolor{darkgray}{9.29} & \textcolor{darkgray}{8.10} & \textcolor{darkgray}{1.71} & - & \textcolor{darkgray}{8.11} & \textcolor{darkgray}{12.5} & \textcolor{darkgray}{9.83} \\
Acasxu 2023 & 4 & ~\textsc{unsat} & \textcolor{darkgray}{9.05} & \textcolor{darkgray}{7.25} & \textcolor{darkgray}{8.10} & \textcolor{darkgray}{1.42} & - & \textcolor{darkgray}{2.90} & \textcolor{darkgray}{11.6} & \textcolor{darkgray}{4.33} \\
Acasxu 2023 & 5 & ~\textsc{unsat} & \textcolor{darkgray}{9.14} & \textcolor{darkgray}{8.80} & \textcolor{darkgray}{8.10} & \textcolor{darkgray}{1.70} & - & \textcolor{darkgray}{3.86} & \textcolor{darkgray}{12.5} & \textcolor{darkgray}{4.44} \\
Acasxu 2023 & 6 & ~\textsc{unsat} & \textcolor{darkgray}{8.95} & \textcolor{darkgray}{6.90} & \textcolor{darkgray}{8.10} & \textcolor{darkgray}{1.37} & - & \textcolor{darkgray}{2.40} & \textcolor{darkgray}{11.3} & \textcolor{darkgray}{4.26} \\
Acasxu 2023 & 7 & ~\textsc{unsat} & \textcolor{darkgray}{8.95} & \textcolor{darkgray}{6.17} & \textcolor{darkgray}{8.10} & \textcolor{darkgray}{1.50} & - & \textcolor{darkgray}{1.93} & \textcolor{darkgray}{12.4} & \textcolor{darkgray}{4.33} \\
\bottomrule
\end{longtable}
}
\end{center}

\begin{center}
{\setlength{\tabcolsep}{1pt}
\scriptsize
\begin{longtable}{@{}lllllll@{}}
\caption{\footnotesize Extended Track Instance Runtimes. Fastest times are \textcolor{blue}{blue}. Second fastest are \textcolor{second}{green}. Penalties are red crosses (\textbf{\textcolor{red}{\ding{55}}}).} \label{tab:all_results_extended} \\
\toprule
\textbf{Category} & \textbf{Id} & \textbf{Result} & \textbf{$\alpha$-$\beta$-C} & \textbf{PyRAT} & \textbf{NNen} & \textbf{NSAT} \\
\midrule
\endhead
Cctsdb Yolo 2023 & 0 & ~\textsc{unsat} & \textcolor{darkgray}{8.39} & - & - & - \\
Cctsdb Yolo 2023 & 1 & ~\textsc{sat} & \textcolor{darkgray}{7.70} & - & - & - \\
Cctsdb Yolo 2023 & 2 & ~\textsc{sat} & \textcolor{darkgray}{7.76} & - & - & - \\
Cctsdb Yolo 2023 & 3 & ~\textsc{sat} & \textcolor{darkgray}{7.75} & - & - & - \\
Cctsdb Yolo 2023 & 4 & ~\textsc{sat} & \textcolor{darkgray}{7.70} & - & - & - \\
Cctsdb Yolo 2023 & 5 & ~\textsc{unsat} & \textcolor{darkgray}{7.65} & - & - & - \\
Cctsdb Yolo 2023 & 6 & ~\textsc{sat} & \textcolor{darkgray}{7.72} & - & - & - \\
Cctsdb Yolo 2023 & 7 & ~\textsc{unsat} & \textcolor{darkgray}{7.64} & - & - & - \\
Cctsdb Yolo 2023 & 8 & ~\textsc{unsat} & \textcolor{darkgray}{7.69} & - & - & - \\
Cctsdb Yolo 2023 & 9 & ~\textsc{sat} & \textcolor{darkgray}{7.72} & - & - & - \\
Cctsdb Yolo 2023 & 10 & ~\textsc{sat} & \textcolor{darkgray}{7.71} & - & - & - \\
Cctsdb Yolo 2023 & 11 & ~\textsc{sat} & \textcolor{darkgray}{7.72} & \textcolor{darkgray}{4.91} & - & - \\
Cctsdb Yolo 2023 & 12 & ~\textsc{sat} & \textcolor{darkgray}{7.70} & - & - & - \\
Cctsdb Yolo 2023 & 13 & ~\textsc{sat} & \textcolor{darkgray}{7.73} & - & - & - \\
Cctsdb Yolo 2023 & 14 & ~\textsc{unsat} & \textcolor{darkgray}{7.65} & - & - & - \\
Cctsdb Yolo 2023 & 15 & ~\textsc{sat} & \textcolor{darkgray}{7.73} & - & - & - \\
Cctsdb Yolo 2023 & 16 & ~\textsc{sat} & \textcolor{darkgray}{7.75} & - & - & - \\
Cctsdb Yolo 2023 & 17 & ~\textsc{sat} & \textcolor{darkgray}{7.71} & - & - & - \\
Cctsdb Yolo 2023 & 18 & ~\textsc{sat} & \textcolor{darkgray}{7.70} & - & - & - \\
Cctsdb Yolo 2023 & 19 & ~\textsc{sat} & \textcolor{darkgray}{7.72} & - & - & - \\
Cctsdb Yolo 2023 & 20 & ~\textsc{sat} & \textcolor{darkgray}{7.71} & \textcolor{darkgray}{4.38} & - & - \\
Cctsdb Yolo 2023 & 21 & ~\textsc{sat} & \textcolor{darkgray}{7.72} & - & - & - \\
Cctsdb Yolo 2023 & 22 & ~\textsc{sat} & \textcolor{darkgray}{7.76} & - & - & - \\
Cctsdb Yolo 2023 & 23 & ~\textsc{unsat} & \textcolor{darkgray}{7.69} & - & - & - \\
Cctsdb Yolo 2023 & 24 & ~\textsc{sat} & \textcolor{darkgray}{7.70} & - & - & - \\
Cctsdb Yolo 2023 & 25 & ~\textsc{sat} & \textcolor{darkgray}{7.70} & - & - & - \\
Cctsdb Yolo 2023 & 26 & ~\textsc{sat} & \textcolor{darkgray}{7.72} & - & - & - \\
Cctsdb Yolo 2023 & 27 & ~\textsc{unsat} & \textcolor{darkgray}{7.67} & - & - & - \\
Cctsdb Yolo 2023 & 28 & ~\textsc{unsat} & \textcolor{darkgray}{7.65} & - & - & - \\
Cctsdb Yolo 2023 & 29 & ~\textsc{unsat} & \textcolor{darkgray}{7.67} & - & - & - \\
Cctsdb Yolo 2023 & 30 & ~\textsc{sat} & \textcolor{darkgray}{7.73} & - & - & - \\
Cctsdb Yolo 2023 & 31 & ~\textsc{sat} & \textcolor{darkgray}{7.70} & - & - & - \\
Cctsdb Yolo 2023 & 32 & ~\textsc{unsat} & \textcolor{darkgray}{7.64} & - & - & - \\
Cctsdb Yolo 2023 & 33 & ~\textsc{sat} & \textcolor{darkgray}{7.71} & - & - & - \\
Cctsdb Yolo 2023 & 34 & ~\textsc{sat} & \textcolor{darkgray}{7.72} & - & - & - \\
Cctsdb Yolo 2023 & 35 & ~\textsc{unsat} & \textcolor{darkgray}{7.67} & - & - & - \\
Cctsdb Yolo 2023 & 36 & ~\textsc{sat} & \textcolor{darkgray}{7.72} & - & - & - \\
Cctsdb Yolo 2023 & 37 & ~\textsc{sat} & \textcolor{darkgray}{7.73} & - & - & - \\
Cctsdb Yolo 2023 & 38 & ~\textsc{sat} & \textcolor{darkgray}{7.71} & - & - & - \\
\midrule
Collins Aerospace Benchmark & 0 & ~\textsc{sat} & \textcolor{darkgray}{15.0} & \textcolor{darkgray}{12.7} & - & - \\
Collins Aerospace Benchmark & 1 & ~\textsc{sat} & \textcolor{darkgray}{14.6} & \textcolor{darkgray}{12.7} & - & - \\
Collins Aerospace Benchmark & 2 & ~\textsc{sat} & \textcolor{darkgray}{14.3} & \textcolor{darkgray}{13.2} & - & - \\
Collins Aerospace Benchmark & 3 & ~\textsc{sat} & \textcolor{darkgray}{14.3} & \textcolor{darkgray}{12.8} & - & - \\
Collins Aerospace Benchmark & 4 & ~\textsc{sat} & \textcolor{darkgray}{14.3} & \textcolor{darkgray}{13.1} & - & - \\
Collins Aerospace Benchmark & 5 & ~\textsc{sat} & \textcolor{darkgray}{14.4} & \textcolor{darkgray}{13.2} & - & - \\
\midrule
Lsnc & 0 & ~\textsc{unsat} & \textcolor{darkgray}{34.0} & - & - & - \\
Lsnc & 1 & ~\textsc{unsat} & \textcolor{darkgray}{20.6} & - & - & - \\
Lsnc & 2 & ~\textsc{unsat} & \textcolor{darkgray}{16.4} & - & - & - \\
Lsnc & 3 & ~\textsc{unsat} & \textcolor{darkgray}{21.9} & - & - & - \\
Lsnc & 4 & ~\textsc{unsat} & \textcolor{darkgray}{17.8} & - & - & - \\
Lsnc & 5 & ~\textsc{unsat} & \textcolor{darkgray}{16.2} & - & - & - \\
Lsnc & 6 & ~\textsc{unsat} & \textcolor{darkgray}{20.6} & - & - & - \\
Lsnc & 7 & ~\textsc{unsat} & \textcolor{darkgray}{21.4} & - & - & - \\
Lsnc & 8 & ~\textsc{unsat} & \textcolor{darkgray}{18.9} & - & - & - \\
Lsnc & 9 & ~\textsc{unsat} & \textcolor{darkgray}{16.1} & - & - & - \\
Lsnc & 10 & ~\textsc{unsat} & \textcolor{darkgray}{26.0} & - & - & - \\
Lsnc & 11 & ~\textsc{unsat} & \textcolor{darkgray}{20.2} & - & - & - \\
Lsnc & 12 & ~\textsc{unsat} & \textcolor{darkgray}{16.6} & - & - & - \\
Lsnc & 13 & ~\textsc{unsat} & \textcolor{darkgray}{17.4} & - & - & - \\
Lsnc & 14 & ~\textsc{unsat} & \textcolor{darkgray}{20.7} & - & - & - \\
Lsnc & 15 & ~\textsc{unsat} & \textcolor{darkgray}{19.0} & - & - & - \\
Lsnc & 16 & ~\textsc{unsat} & \textcolor{darkgray}{16.0} & - & - & - \\
Lsnc & 17 & ~\textsc{unsat} & \textcolor{darkgray}{32.7} & - & - & - \\
Lsnc & 18 & ~\textsc{unsat} & \textcolor{darkgray}{20.4} & - & - & - \\
Lsnc & 19 & ~\textsc{unsat} & \textcolor{darkgray}{26.4} & - & - & - \\
Lsnc & 20 & ~\textsc{unsat} & \textcolor{darkgray}{30.1} & \textcolor{darkgray}{20.6} & - & - \\
Lsnc & 21 & ~\textsc{unsat} & \textcolor{darkgray}{22.9} & \textcolor{darkgray}{3.85} & - & - \\
Lsnc & 22 & ~\textsc{unsat} & \textcolor{darkgray}{28.5} & \textcolor{darkgray}{4.55} & - & - \\
Lsnc & 23 & ~\textsc{unsat} & \textcolor{darkgray}{22.8} & \textcolor{darkgray}{3.83} & - & - \\
Lsnc & 24 & ~\textsc{unsat} & \textcolor{darkgray}{23.3} & \textcolor{darkgray}{3.82} & - & - \\
Lsnc & 25 & ~\textsc{unsat} & \textcolor{darkgray}{29.4} & \textcolor{darkgray}{6.16} & - & - \\
Lsnc & 26 & ~\textsc{unsat} & \textcolor{darkgray}{48.3} & - & - & - \\
Lsnc & 27 & ~\textsc{unsat} & \textcolor{darkgray}{23.0} & \textcolor{darkgray}{3.82} & - & - \\
Lsnc & 28 & ~\textsc{unsat} & \textcolor{darkgray}{34.7} & - & - & - \\
Lsnc & 29 & ~\textsc{unsat} & \textcolor{darkgray}{40.1} & - & - & - \\
Lsnc & 30 & ~\textsc{unsat} & \textcolor{darkgray}{21.8} & \textcolor{darkgray}{3.84} & - & - \\
Lsnc & 31 & ~\textsc{unsat} & \textcolor{darkgray}{38.0} & - & - & - \\
Lsnc & 32 & ~\textsc{unsat} & \textcolor{darkgray}{22.0} & \textcolor{darkgray}{3.84} & - & - \\
Lsnc & 33 & ~\textsc{unsat} & \textcolor{darkgray}{23.0} & \textcolor{darkgray}{3.79} & - & - \\
Lsnc & 34 & ~\textsc{unsat} & \textcolor{darkgray}{25.4} & \textcolor{darkgray}{11.9} & - & - \\
Lsnc & 35 & ~\textsc{unsat} & \textcolor{darkgray}{22.8} & \textcolor{darkgray}{3.86} & - & - \\
Lsnc & 36 & ~\textsc{unsat} & \textcolor{darkgray}{23.4} & \textcolor{darkgray}{3.82} & - & - \\
Lsnc & 37 & ~\textsc{unsat} & \textcolor{darkgray}{27.4} & \textcolor{darkgray}{8.41} & - & - \\
Lsnc & 38 & ~\textsc{unsat} & \textcolor{darkgray}{28.4} & \textcolor{darkgray}{4.33} & - & - \\
Lsnc & 39 & ~\textsc{unsat} & \textcolor{darkgray}{34.7} & - & - & - \\
\midrule
Ml4acopf 2023 & 0 & ~\textsc{unsat} & \textcolor{darkgray}{77.2} & - & - & - \\
Ml4acopf 2023 & 1 & ~\textsc{unsat} & \textcolor{darkgray}{56.4} & \textcolor{darkgray}{6.23} & - & - \\
Ml4acopf 2023 & 2 & ~\textsc{unsat} & \textcolor{darkgray}{114} & \textcolor{darkgray}{6.06} & - & - \\
Ml4acopf 2023 & 3 & ~\textsc{unsat} & \textcolor{darkgray}{9.08} & - & - & - \\
Ml4acopf 2023 & 4 & ~\textsc{unsat} & \textcolor{darkgray}{9.08} & \textcolor{darkgray}{4.76} & - & - \\
Ml4acopf 2023 & 5 & ~\textsc{unsat} & \textcolor{darkgray}{9.02} & - & - & - \\
Ml4acopf 2023 & 6 & ~\textsc{unsat} & \textcolor{darkgray}{9.01} & - & - & - \\
Ml4acopf 2023 & 7 & ~\textsc{unsat} & \textcolor{darkgray}{9.26} & \textcolor{darkgray}{4.15} & - & - \\
Ml4acopf 2023 & 8 & ~\textsc{unsat} & \textcolor{darkgray}{9.03} & \textcolor{darkgray}{171} & - & - \\
Ml4acopf 2023 & 9 & ~\textsc{unsat} & \textcolor{darkgray}{8.98} & \textcolor{darkgray}{4.28} & - & - \\
Ml4acopf 2023 & 10 & ~\textsc{unsat} & \textcolor{darkgray}{9.03} & \textcolor{darkgray}{5.72} & - & - \\
Ml4acopf 2023 & 11 & ~\textsc{unsat} & \textcolor{darkgray}{9.05} & \textcolor{darkgray}{4.20} & - & - \\
Ml4acopf 2023 & 12 & ~\textsc{unsat} & \textcolor{darkgray}{9.06} & \textcolor{darkgray}{4.85} & - & - \\
Ml4acopf 2023 & 13 & ~\textsc{unsat} & \textcolor{darkgray}{9.03} & \textcolor{darkgray}{4.22} & - & - \\
Ml4acopf 2023 & 14 & ~\textsc{unsat} & \textcolor{darkgray}{9.16} & - & - & - \\
Ml4acopf 2023 & 15 & ~\textsc{unsat} & \textcolor{darkgray}{10.1} & \textcolor{darkgray}{4.24} & - & - \\
Ml4acopf 2023 & 16 & ~\textsc{unsat} & \textcolor{darkgray}{9.07} & - & - & - \\
Ml4acopf 2023 & 17 & ~\textsc{unsat} & \textcolor{darkgray}{9.04} & \textcolor{darkgray}{4.22} & - & - \\
Ml4acopf 2023 & 18 & ~\textsc{unsat} & \textcolor{darkgray}{9.05} & - & - & - \\
Ml4acopf 2023 & 19 & ~\textsc{unsat} & \textcolor{darkgray}{362} & \textcolor{darkgray}{11.4} & - & - \\
Ml4acopf 2023 & 20 & ~\textsc{unsat} & \textcolor{darkgray}{9.37} & \textcolor{darkgray}{6.34} & - & - \\
Ml4acopf 2023 & 21 & ~? & - & - & - & - \\
Ml4acopf 2023 & 22 & ~\textsc{unsat} & \textcolor{darkgray}{173} & \textcolor{darkgray}{16.8} & - & - \\
\midrule
Ml4acopf 2024 & 0 & ~\textsc{unsat} & \textcolor{darkgray}{9.27} & - & - & - \\
Ml4acopf 2024 & 1 & ~\textsc{unsat} & \textcolor{darkgray}{9.28} & - & - & - \\
Ml4acopf 2024 & 2 & ~\textsc{unsat} & \textcolor{darkgray}{9.42} & \textcolor{darkgray}{9.91} & - & - \\
Ml4acopf 2024 & 3 & ~\textsc{unsat} & \textcolor{darkgray}{9.24} & - & - & - \\
Ml4acopf 2024 & 4 & ~\textsc{unsat} & \textcolor{darkgray}{9.13} & \textcolor{darkgray}{9.94} & - & - \\
Ml4acopf 2024 & 5 & ~\textsc{unsat} & \textcolor{darkgray}{9.28} & \textcolor{darkgray}{18.2} & - & - \\
Ml4acopf 2024 & 6 & ~\textsc{unsat} & \textcolor{darkgray}{9.25} & \textcolor{darkgray}{7.48} & - & - \\
Ml4acopf 2024 & 7 & ~\textsc{unsat} & \textcolor{darkgray}{9.24} & \textcolor{darkgray}{8.90} & - & - \\
Ml4acopf 2024 & 8 & ~\textsc{unsat} & \textcolor{darkgray}{9.27} & \textcolor{darkgray}{7.41} & - & - \\
Ml4acopf 2024 & 9 & ~\textsc{sat} & \textcolor{darkgray}{7.13} & \textcolor{darkgray}{6.18} & - & - \\
Ml4acopf 2024 & 10 & ~\textsc{unsat} & \textcolor{darkgray}{9.38} & - & - & - \\
Ml4acopf 2024 & 11 & ~\textsc{unsat} & \textcolor{darkgray}{9.25} & - & - & - \\
Ml4acopf 2024 & 12 & ~\textsc{unsat} & \textcolor{darkgray}{9.28} & \textcolor{darkgray}{7.43} & - & - \\
Ml4acopf 2024 & 13 & ~\textsc{unsat} & \textcolor{darkgray}{9.29} & - & - & - \\
Ml4acopf 2024 & 14 & ~\textsc{unsat} & \textcolor{darkgray}{80.7} & - & - & - \\
Ml4acopf 2024 & 15 & ~\textsc{unsat} & \textcolor{darkgray}{56.1} & \textcolor{darkgray}{8.31} & - & - \\
Ml4acopf 2024 & 16 & ~? & - & - & - & - \\
Ml4acopf 2024 & 17 & ~\textsc{unsat} & \textcolor{darkgray}{9.19} & - & - & - \\
Ml4acopf 2024 & 18 & ~\textsc{unsat} & \textcolor{darkgray}{9.09} & \textcolor{darkgray}{7.36} & - & - \\
Ml4acopf 2024 & 19 & ~\textsc{unsat} & \textcolor{darkgray}{180} & - & - & - \\
Ml4acopf 2024 & 20 & ~\textsc{unsat} & \textcolor{darkgray}{9.21} & \textcolor{darkgray}{7.84} & - & - \\
Ml4acopf 2024 & 21 & ~? & - & - & - & - \\
Ml4acopf 2024 & 22 & ~\textsc{unsat} & \textcolor{darkgray}{180} & - & - & - \\
Ml4acopf 2024 & 23 & ~\textsc{unsat} & \textcolor{darkgray}{8.90} & - & - & - \\
Ml4acopf 2024 & 24 & ~\textsc{unsat} & \textcolor{darkgray}{8.86} & - & - & - \\
Ml4acopf 2024 & 25 & ~\textsc{unsat} & \textcolor{darkgray}{8.99} & \textcolor{darkgray}{8.51} & - & - \\
Ml4acopf 2024 & 26 & ~\textsc{unsat} & \textcolor{darkgray}{8.81} & - & - & - \\
Ml4acopf 2024 & 27 & ~\textsc{unsat} & \textcolor{darkgray}{8.79} & \textcolor{darkgray}{8.60} & - & - \\
Ml4acopf 2024 & 28 & ~\textsc{unsat} & \textcolor{darkgray}{8.90} & \textcolor{darkgray}{35.0} & - & - \\
Ml4acopf 2024 & 29 & ~\textsc{unsat} & \textcolor{darkgray}{8.84} & \textcolor{darkgray}{7.17} & - & - \\
Ml4acopf 2024 & 30 & ~\textsc{unsat} & \textcolor{darkgray}{8.86} & \textcolor{darkgray}{24.9} & - & - \\
Ml4acopf 2024 & 31 & ~\textsc{unsat} & \textcolor{darkgray}{8.87} & \textcolor{darkgray}{10.6} & - & - \\
Ml4acopf 2024 & 32 & ~\textsc{sat} & \textcolor{darkgray}{7.09} & \textcolor{darkgray}{6.14} & - & - \\
Ml4acopf 2024 & 33 & ~\textsc{unsat} & \textcolor{darkgray}{9.04} & - & - & - \\
Ml4acopf 2024 & 34 & ~\textsc{unsat} & \textcolor{darkgray}{8.86} & - & - & - \\
Ml4acopf 2024 & 35 & ~\textsc{unsat} & \textcolor{darkgray}{8.84} & \textcolor{darkgray}{7.21} & - & - \\
Ml4acopf 2024 & 36 & ~\textsc{unsat} & \textcolor{darkgray}{8.81} & - & - & - \\
Ml4acopf 2024 & 37 & ~\textsc{unsat} & \textcolor{darkgray}{63.8} & - & - & - \\
Ml4acopf 2024 & 38 & ~\textsc{unsat} & \textcolor{darkgray}{54.6} & - & - & - \\
Ml4acopf 2024 & 39 & ~\textsc{unsat} & \textcolor{darkgray}{67.5} & - & - & - \\
Ml4acopf 2024 & 40 & ~\textsc{sat} & \textcolor{darkgray}{7.15} & \textcolor{darkgray}{6.20} & - & - \\
Ml4acopf 2024 & 41 & ~\textsc{unsat} & \textcolor{darkgray}{8.91} & \textcolor{darkgray}{7.40} & - & - \\
Ml4acopf 2024 & 42 & ~\textsc{unsat} & \textcolor{darkgray}{9.03} & - & - & - \\
Ml4acopf 2024 & 43 & ~\textsc{unsat} & \textcolor{darkgray}{9.08} & - & - & - \\
Ml4acopf 2024 & 44 & ~\textsc{unsat} & \textcolor{darkgray}{9.24} & \textcolor{darkgray}{7.62} & - & - \\
Ml4acopf 2024 & 45 & ~\textsc{unsat} & \textcolor{darkgray}{9.09} & - & - & - \\
Ml4acopf 2024 & 46 & ~\textsc{unsat} & \textcolor{darkgray}{9.05} & \textcolor{darkgray}{7.69} & - & - \\
Ml4acopf 2024 & 47 & ~\textsc{unsat} & \textcolor{darkgray}{9.04} & \textcolor{darkgray}{8.03} & - & - \\
Ml4acopf 2024 & 48 & ~\textsc{unsat} & \textcolor{darkgray}{9.04} & \textcolor{darkgray}{7.10} & - & - \\
Ml4acopf 2024 & 49 & ~\textsc{unsat} & \textcolor{darkgray}{9.09} & \textcolor{darkgray}{8.00} & - & - \\
Ml4acopf 2024 & 50 & ~\textsc{unsat} & \textcolor{darkgray}{9.06} & \textcolor{darkgray}{7.09} & - & - \\
Ml4acopf 2024 & 51 & ~\textsc{unsat} & \textcolor{darkgray}{9.05} & - & - & - \\
Ml4acopf 2024 & 52 & ~\textsc{unsat} & \textcolor{darkgray}{10.1} & \textcolor{darkgray}{7.74} & - & - \\
Ml4acopf 2024 & 53 & ~\textsc{unsat} & \textcolor{darkgray}{9.08} & - & - & - \\
Ml4acopf 2024 & 54 & ~\textsc{unsat} & \textcolor{darkgray}{9.02} & \textcolor{darkgray}{7.09} & - & - \\
Ml4acopf 2024 & 55 & ~\textsc{unsat} & \textcolor{darkgray}{9.07} & - & - & - \\
Ml4acopf 2024 & 56 & ~\textsc{unsat} & \textcolor{darkgray}{202} & - & - & - \\
Ml4acopf 2024 & 57 & ~\textsc{unsat} & \textcolor{darkgray}{9.73} & \textcolor{darkgray}{7.58} & - & - \\
Ml4acopf 2024 & 58 & ~? & - & - & - & - \\
Ml4acopf 2024 & 59 & ~\textsc{unsat} & \textcolor{darkgray}{188} & - & - & - \\
Ml4acopf 2024 & 60 & ~? & - & - & - & - \\
Ml4acopf 2024 & 61 & ~\textsc{unsat} & \textcolor{darkgray}{56.2} & - & - & - \\
Ml4acopf 2024 & 62 & ~\textsc{unsat} & \textcolor{darkgray}{72.7} & - & - & - \\
Ml4acopf 2024 & 63 & ~\textsc{unsat} & \textcolor{darkgray}{9.44} & - & - & - \\
Ml4acopf 2024 & 64 & ~\textsc{unsat} & \textcolor{darkgray}{9.43} & \textcolor{darkgray}{9.06} & - & - \\
Ml4acopf 2024 & 65 & ~? & - & - & - & - \\
Ml4acopf 2024 & 66 & ~\textsc{unsat} & \textcolor{darkgray}{9.39} & \textcolor{darkgray}{7.51} & - & - \\
Ml4acopf 2024 & 67 & ~? & - & - & - & - \\
Ml4acopf 2024 & 68 & ~? & - & - & - & - \\
\midrule
Traffic Signs Recognition 2023 & 0 & ~\textsc{sat} & \textcolor{darkgray}{8.02} & - & - & - \\
Traffic Signs Recognition 2023 & 1 & ~\textsc{sat} & \textcolor{darkgray}{8.06} & - & - & - \\
Traffic Signs Recognition 2023 & 2 & ~\textsc{sat} & \textcolor{darkgray}{7.96} & - & - & - \\
Traffic Signs Recognition 2023 & 3 & ~\textsc{sat} & \textcolor{darkgray}{7.98} & - & - & - \\
Traffic Signs Recognition 2023 & 4 & ~\textsc{sat} & \textcolor{darkgray}{7.66} & \textcolor{darkgray}{4.39} & - & ~~\textbf{\textcolor{red}{\ding{55}}} \\
Traffic Signs Recognition 2023 & 5 & ~\textsc{sat} & \textcolor{darkgray}{7.99} & - & - & ~~\textbf{\textcolor{red}{\ding{55}}} \\
Traffic Signs Recognition 2023 & 6 & ~\textsc{sat} & \textcolor{darkgray}{7.64} & - & - & - \\
Traffic Signs Recognition 2023 & 7 & ~\textsc{sat} & \textcolor{darkgray}{7.62} & - & - & ~~\textbf{\textcolor{red}{\ding{55}}} \\
Traffic Signs Recognition 2023 & 8 & ~\textsc{sat} & \textcolor{darkgray}{7.60} & - & - & ~~\textbf{\textcolor{red}{\ding{55}}} \\
Traffic Signs Recognition 2023 & 9 & ~\textsc{sat} & \textcolor{darkgray}{7.64} & \textcolor{darkgray}{4.40} & - & ~~\textbf{\textcolor{red}{\ding{55}}} \\
Traffic Signs Recognition 2023 & 10 & ~\textsc{sat} & \textcolor{darkgray}{8.02} & - & - & - \\
Traffic Signs Recognition 2023 & 11 & ~\textsc{sat} & \textcolor{darkgray}{8.00} & - & - & - \\
Traffic Signs Recognition 2023 & 12 & ~\textsc{sat} & \textcolor{darkgray}{7.99} & - & - & - \\
Traffic Signs Recognition 2023 & 13 & ~\textsc{sat} & \textcolor{darkgray}{8.04} & - & - & - \\
Traffic Signs Recognition 2023 & 14 & ~\textsc{sat} & \textcolor{darkgray}{7.66} & - & - & ~~\textbf{\textcolor{red}{\ding{55}}} \\
Traffic Signs Recognition 2023 & 15 & ~\textsc{sat} & \textcolor{darkgray}{8.75} & - & - & - \\
Traffic Signs Recognition 2023 & 16 & ~\textsc{sat} & \textcolor{darkgray}{8.35} & - & - & - \\
Traffic Signs Recognition 2023 & 17 & ~\textsc{sat} & \textcolor{darkgray}{8.24} & - & - & - \\
Traffic Signs Recognition 2023 & 18 & ~\textsc{sat} & \textcolor{darkgray}{8.19} & - & - & - \\
Traffic Signs Recognition 2023 & 19 & ~\textsc{sat} & \textcolor{darkgray}{8.18} & - & - & - \\
Traffic Signs Recognition 2023 & 20 & ~\textsc{sat} & \textcolor{darkgray}{8.33} & - & - & - \\
Traffic Signs Recognition 2023 & 21 & ~\textsc{sat} & \textcolor{darkgray}{8.21} & - & - & - \\
Traffic Signs Recognition 2023 & 22 & ~\textsc{sat} & \textcolor{darkgray}{8.21} & - & - & - \\
Traffic Signs Recognition 2023 & 23 & ~\textsc{sat} & \textcolor{darkgray}{8.22} & - & - & - \\
Traffic Signs Recognition 2023 & 24 & ~\textsc{sat} & \textcolor{darkgray}{7.83} & - & - & ~~\textbf{\textcolor{red}{\ding{55}}} \\
Traffic Signs Recognition 2023 & 25 & ~\textsc{sat} & \textcolor{darkgray}{8.19} & - & - & - \\
Traffic Signs Recognition 2023 & 26 & ~\textsc{sat} & \textcolor{darkgray}{7.85} & \textcolor{darkgray}{4.93} & - & ~~\textbf{\textcolor{red}{\ding{55}}} \\
Traffic Signs Recognition 2023 & 27 & ~\textsc{sat} & \textcolor{darkgray}{7.85} & - & - & ~~\textbf{\textcolor{red}{\ding{55}}} \\
Traffic Signs Recognition 2023 & 28 & ~\textsc{sat} & \textcolor{darkgray}{7.85} & - & - & ~~\textbf{\textcolor{red}{\ding{55}}} \\
Traffic Signs Recognition 2023 & 29 & ~\textsc{sat} & \textcolor{darkgray}{7.84} & \textcolor{darkgray}{4.90} & - & ~~\textbf{\textcolor{red}{\ding{55}}} \\
Traffic Signs Recognition 2023 & 30 & ~\textsc{sat} & \textcolor{darkgray}{8.96} & - & - & - \\
Traffic Signs Recognition 2023 & 31 & ~\textsc{sat} & \textcolor{darkgray}{8.61} & - & - & - \\
Traffic Signs Recognition 2023 & 32 & ~\textsc{sat} & \textcolor{darkgray}{8.53} & - & - & - \\
Traffic Signs Recognition 2023 & 33 & ~\textsc{sat} & \textcolor{darkgray}{8.52} & - & - & - \\
Traffic Signs Recognition 2023 & 34 & ~\textsc{sat} & \textcolor{darkgray}{8.54} & - & - & - \\
Traffic Signs Recognition 2023 & 35 & ~\textsc{sat} & \textcolor{darkgray}{8.68} & - & - & - \\
Traffic Signs Recognition 2023 & 36 & ~\textsc{sat} & \textcolor{darkgray}{8.53} & - & - & - \\
Traffic Signs Recognition 2023 & 37 & ~\textsc{sat} & \textcolor{darkgray}{8.54} & - & - & - \\
Traffic Signs Recognition 2023 & 38 & ~\textsc{sat} & \textcolor{darkgray}{8.52} & - & - & - \\
Traffic Signs Recognition 2023 & 39 & ~\textsc{sat} & \textcolor{darkgray}{8.52} & - & - & - \\
Traffic Signs Recognition 2023 & 40 & ~\textsc{sat} & \textcolor{darkgray}{8.51} & - & - & - \\
Traffic Signs Recognition 2023 & 41 & ~\textsc{sat} & \textcolor{darkgray}{8.55} & - & - & - \\
Traffic Signs Recognition 2023 & 42 & ~\textsc{sat} & \textcolor{darkgray}{8.54} & - & - & - \\
Traffic Signs Recognition 2023 & 43 & ~\textsc{sat} & \textcolor{darkgray}{8.54} & - & - & - \\
Traffic Signs Recognition 2023 & 44 & ~\textsc{sat} & \textcolor{darkgray}{8.54} & - & - & - \\
\midrule
Vggnet16 2023 & 0 & ~\textsc{unsat} & \textcolor{darkgray}{16.5} & \textcolor{darkgray}{99.1} & \textcolor{darkgray}{29.0} & \textcolor{darkgray}{47.0} \\
Vggnet16 2023 & 1 & ~\textsc{unsat} & \textcolor{darkgray}{16.5} & \textcolor{darkgray}{96.2} & \textcolor{darkgray}{29.8} & \textcolor{darkgray}{46.5} \\
Vggnet16 2023 & 2 & ~\textsc{unsat} & \textcolor{darkgray}{16.6} & \textcolor{darkgray}{98.4} & \textcolor{darkgray}{30.8} & - \\
Vggnet16 2023 & 3 & ~\textsc{unsat} & \textcolor{darkgray}{16.6} & \textcolor{darkgray}{97.4} & \textcolor{darkgray}{29.9} & \textcolor{darkgray}{47.0} \\
Vggnet16 2023 & 4 & ~\textsc{unsat} & \textcolor{darkgray}{16.6} & \textcolor{darkgray}{99.8} & \textcolor{darkgray}{45.0} & \textcolor{darkgray}{46.5} \\
Vggnet16 2023 & 5 & ~\textsc{unsat} & \textcolor{darkgray}{16.6} & \textcolor{darkgray}{107} & \textcolor{darkgray}{55.9} & - \\
Vggnet16 2023 & 6 & ~\textsc{unsat} & \textcolor{darkgray}{16.6} & \textcolor{darkgray}{98.4} & \textcolor{darkgray}{39.6} & \textcolor{darkgray}{47.1} \\
Vggnet16 2023 & 7 & ~\textsc{unsat} & \textcolor{darkgray}{16.7} & \textcolor{darkgray}{104} & \textcolor{darkgray}{50.5} & - \\
Vggnet16 2023 & 8 & ~\textsc{unsat} & \textcolor{darkgray}{16.7} & - & \textcolor{darkgray}{81.1} & - \\
Vggnet16 2023 & 9 & ~\textsc{unsat} & \textcolor{darkgray}{16.6} & \textcolor{darkgray}{98.2} & \textcolor{darkgray}{41.1} & \textcolor{darkgray}{46.8} \\
Vggnet16 2023 & 10 & ~\textsc{unsat} & \textcolor{darkgray}{16.6} & \textcolor{darkgray}{109} & \textcolor{darkgray}{64.5} & - \\
Vggnet16 2023 & 11 & ~\textsc{unsat} & \textcolor{darkgray}{17.5} & - & \textcolor{darkgray}{129} & - \\
Vggnet16 2023 & 12 & ~\textsc{unsat} & \textcolor{darkgray}{361} & - & - & - \\
Vggnet16 2023 & 13 & ~\textsc{unsat} & \textcolor{darkgray}{26.1} & - & \textcolor{darkgray}{328} & - \\
Vggnet16 2023 & 14 & ~\textsc{unsat} & \textcolor{darkgray}{1024} & - & - & - \\
Vggnet16 2023 & 15 & ~\textsc{unsat} & \textcolor{darkgray}{1009} & - & - & - \\
Vggnet16 2023 & 16 & ~\textsc{unsat} & \textcolor{darkgray}{1046} & - & - & - \\
Vggnet16 2023 & 17 & ~\textsc{unsat} & \textcolor{darkgray}{1092} & - & - & - \\
\midrule
Vit 2023 & 0 & ~\textsc{unsat} & \textcolor{darkgray}{16.0} & - & - & - \\
Vit 2023 & 1 & ~\textsc{unsat} & \textcolor{darkgray}{17.1} & - & - & - \\
Vit 2023 & 2 & ~\textsc{unsat} & \textcolor{darkgray}{16.0} & - & - & - \\
Vit 2023 & 3 & ~? & - & - & - & - \\
Vit 2023 & 4 & ~? & - & - & - & - \\
Vit 2023 & 5 & ~? & - & - & - & - \\
Vit 2023 & 6 & ~\textsc{unsat} & \textcolor{darkgray}{15.5} & - & - & \textcolor{darkgray}{23.1} \\
Vit 2023 & 7 & ~? & - & - & - & - \\
Vit 2023 & 8 & ~? & - & - & - & - \\
Vit 2023 & 9 & ~? & - & - & - & - \\
Vit 2023 & 10 & ~? & - & - & - & - \\
Vit 2023 & 11 & ~\textsc{unsat} & \textcolor{darkgray}{14.9} & - & - & \textcolor{darkgray}{17.7} \\
Vit 2023 & 12 & ~\textsc{unsat} & \textcolor{darkgray}{16.6} & - & - & - \\
Vit 2023 & 13 & ~\textsc{unsat} & \textcolor{darkgray}{14.9} & - & - & \textcolor{darkgray}{17.7} \\
Vit 2023 & 14 & ~? & - & - & - & - \\
Vit 2023 & 15 & ~? & - & - & - & - \\
Vit 2023 & 16 & ~? & - & - & - & - \\
Vit 2023 & 17 & ~\textsc{unsat} & \textcolor{darkgray}{29.1} & - & - & - \\
Vit 2023 & 18 & ~? & - & - & - & - \\
Vit 2023 & 19 & ~\textsc{unsat} & \textcolor{darkgray}{14.9} & - & - & - \\
Vit 2023 & 20 & ~\textsc{unsat} & \textcolor{darkgray}{48.1} & - & - & - \\
Vit 2023 & 21 & ~\textsc{unsat} & \textcolor{darkgray}{54.6} & - & - & \textcolor{darkgray}{26.6} \\
Vit 2023 & 22 & ~? & - & - & - & - \\
Vit 2023 & 23 & ~? & - & - & - & - \\
Vit 2023 & 24 & ~\textsc{unsat} & \textcolor{darkgray}{14.9} & - & - & \textcolor{darkgray}{17.6} \\
Vit 2023 & 25 & ~? & - & - & - & - \\
Vit 2023 & 26 & ~? & - & - & - & - \\
Vit 2023 & 27 & ~? & - & - & - & - \\
Vit 2023 & 28 & ~? & - & - & - & - \\
Vit 2023 & 29 & ~? & - & - & - & - \\
Vit 2023 & 30 & ~\textsc{unsat} & \textcolor{darkgray}{44.0} & - & - & - \\
Vit 2023 & 31 & ~\textsc{unsat} & \textcolor{darkgray}{14.9} & - & - & - \\
Vit 2023 & 32 & ~\textsc{unsat} & \textcolor{darkgray}{15.5} & - & - & - \\
Vit 2023 & 33 & ~\textsc{unsat} & \textcolor{darkgray}{15.5} & - & - & \textcolor{darkgray}{21.8} \\
Vit 2023 & 34 & ~\textsc{unsat} & \textcolor{darkgray}{57.0} & - & - & - \\
Vit 2023 & 35 & ~? & - & - & - & - \\
Vit 2023 & 36 & ~? & - & - & - & - \\
Vit 2023 & 37 & ~? & - & - & - & - \\
Vit 2023 & 38 & ~\textsc{unsat} & \textcolor{darkgray}{47.1} & - & - & - \\
Vit 2023 & 39 & ~? & - & - & - & - \\
Vit 2023 & 40 & ~\textsc{unsat} & \textcolor{darkgray}{15.5} & - & - & - \\
Vit 2023 & 41 & ~\textsc{unsat} & \textcolor{darkgray}{21.0} & - & - & - \\
Vit 2023 & 42 & ~? & - & - & - & - \\
Vit 2023 & 43 & ~? & - & - & - & - \\
Vit 2023 & 44 & ~? & - & - & - & - \\
Vit 2023 & 45 & ~? & - & - & - & - \\
Vit 2023 & 46 & ~\textsc{unsat} & \textcolor{darkgray}{14.9} & - & - & - \\
Vit 2023 & 47 & ~? & - & - & - & - \\
Vit 2023 & 48 & ~? & - & - & - & - \\
Vit 2023 & 49 & ~? & - & - & - & - \\
Vit 2023 & 50 & ~\textsc{unsat} & \textcolor{darkgray}{14.9} & - & - & - \\
Vit 2023 & 51 & ~? & - & - & - & - \\
Vit 2023 & 52 & ~\textsc{unsat} & \textcolor{darkgray}{48.8} & - & - & - \\
Vit 2023 & 53 & ~? & - & - & - & - \\
Vit 2023 & 54 & ~\textsc{unsat} & \textcolor{darkgray}{15.5} & - & - & - \\
Vit 2023 & 55 & ~? & - & - & - & - \\
Vit 2023 & 56 & ~\textsc{unsat} & \textcolor{darkgray}{14.9} & - & - & - \\
Vit 2023 & 57 & ~? & - & - & - & - \\
Vit 2023 & 58 & ~\textsc{unsat} & \textcolor{darkgray}{15.5} & - & - & \textcolor{darkgray}{20.3} \\
Vit 2023 & 59 & ~? & - & - & - & - \\
Vit 2023 & 60 & ~? & - & - & - & - \\
Vit 2023 & 61 & ~? & - & - & - & - \\
Vit 2023 & 62 & ~? & - & - & - & - \\
Vit 2023 & 63 & ~\textsc{unsat} & \textcolor{darkgray}{15.4} & - & - & - \\
Vit 2023 & 64 & ~\textsc{unsat} & \textcolor{darkgray}{16.0} & - & - & \textcolor{darkgray}{20.0} \\
Vit 2023 & 65 & ~\textsc{unsat} & \textcolor{darkgray}{15.4} & - & - & - \\
Vit 2023 & 66 & ~? & - & - & - & - \\
Vit 2023 & 67 & ~? & - & - & - & - \\
Vit 2023 & 68 & ~? & - & - & - & - \\
Vit 2023 & 69 & ~? & - & - & - & - \\
Vit 2023 & 70 & ~? & - & - & - & - \\
Vit 2023 & 71 & ~\textsc{unsat} & \textcolor{darkgray}{15.5} & - & - & - \\
Vit 2023 & 72 & ~? & - & - & - & - \\
Vit 2023 & 73 & ~\textsc{unsat} & \textcolor{darkgray}{16.6} & - & - & - \\
Vit 2023 & 74 & ~\textsc{unsat} & \textcolor{darkgray}{42.7} & - & - & \textcolor{darkgray}{37.4} \\
Vit 2023 & 75 & ~\textsc{unsat} & \textcolor{darkgray}{15.5} & - & - & - \\
Vit 2023 & 76 & ~? & - & - & - & - \\
Vit 2023 & 77 & ~\textsc{unsat} & \textcolor{darkgray}{16.0} & - & - & \textcolor{darkgray}{23.3} \\
Vit 2023 & 78 & ~\textsc{unsat} & \textcolor{darkgray}{14.9} & - & - & - \\
Vit 2023 & 79 & ~? & - & - & - & - \\
Vit 2023 & 80 & ~\textsc{unsat} & \textcolor{darkgray}{17.7} & - & - & - \\
Vit 2023 & 81 & ~\textsc{unsat} & \textcolor{darkgray}{14.9} & - & - & \textcolor{darkgray}{20.4} \\
Vit 2023 & 82 & ~? & - & - & - & - \\
Vit 2023 & 83 & ~? & - & - & - & - \\
Vit 2023 & 84 & ~? & - & - & - & - \\
Vit 2023 & 85 & ~\textsc{unsat} & \textcolor{darkgray}{19.3} & - & - & - \\
Vit 2023 & 86 & ~? & - & - & - & - \\
Vit 2023 & 87 & ~\textsc{unsat} & \textcolor{darkgray}{23.4} & - & - & - \\
Vit 2023 & 88 & ~? & - & - & - & - \\
Vit 2023 & 89 & ~\textsc{unsat} & \textcolor{darkgray}{14.9} & - & - & - \\
Vit 2023 & 90 & ~? & - & - & - & - \\
Vit 2023 & 91 & ~\textsc{unsat} & \textcolor{darkgray}{14.9} & - & - & - \\
Vit 2023 & 92 & ~? & - & - & - & - \\
Vit 2023 & 93 & ~\textsc{unsat} & \textcolor{darkgray}{44.7} & - & - & - \\
Vit 2023 & 94 & ~? & - & - & - & - \\
Vit 2023 & 95 & ~? & - & - & - & - \\
Vit 2023 & 96 & ~? & - & - & - & - \\
Vit 2023 & 97 & ~? & - & - & - & - \\
Vit 2023 & 98 & ~? & - & - & - & - \\
Vit 2023 & 99 & ~? & - & - & - & - \\
Vit 2023 & 100 & ~? & - & - & - & - \\
Vit 2023 & 101 & ~? & - & - & - & - \\
Vit 2023 & 102 & ~? & - & - & - & - \\
Vit 2023 & 103 & ~\textsc{unsat} & \textcolor{darkgray}{50.5} & - & - & - \\
Vit 2023 & 104 & ~? & - & - & - & - \\
Vit 2023 & 105 & ~? & - & - & - & - \\
Vit 2023 & 106 & ~\textsc{unsat} & \textcolor{darkgray}{17.1} & - & - & - \\
Vit 2023 & 107 & ~? & - & - & - & - \\
Vit 2023 & 108 & ~? & - & - & - & - \\
Vit 2023 & 109 & ~? & - & - & - & - \\
Vit 2023 & 110 & ~? & - & - & - & - \\
Vit 2023 & 111 & ~? & - & - & - & - \\
Vit 2023 & 112 & ~? & - & - & - & - \\
Vit 2023 & 113 & ~\textsc{unsat} & \textcolor{darkgray}{54.6} & - & - & - \\
Vit 2023 & 114 & ~? & - & - & - & - \\
Vit 2023 & 115 & ~? & - & - & - & - \\
Vit 2023 & 116 & ~\textsc{unsat} & \textcolor{darkgray}{19.8} & - & - & - \\
Vit 2023 & 117 & ~\textsc{unsat} & \textcolor{darkgray}{18.5} & - & - & - \\
Vit 2023 & 118 & ~? & - & - & - & - \\
Vit 2023 & 119 & ~? & - & - & - & - \\
Vit 2023 & 120 & ~\textsc{unsat} & \textcolor{darkgray}{18.5} & - & - & - \\
Vit 2023 & 121 & ~\textsc{unsat} & \textcolor{darkgray}{45.8} & - & - & - \\
Vit 2023 & 122 & ~\textsc{unsat} & \textcolor{darkgray}{50.9} & - & - & - \\
Vit 2023 & 123 & ~? & - & - & - & - \\
Vit 2023 & 124 & ~\textsc{unsat} & \textcolor{darkgray}{18.6} & - & - & - \\
Vit 2023 & 125 & ~? & - & - & - & - \\
Vit 2023 & 126 & ~? & - & - & - & - \\
Vit 2023 & 127 & ~? & - & - & - & - \\
Vit 2023 & 128 & ~? & - & - & - & - \\
Vit 2023 & 129 & ~? & - & - & - & - \\
Vit 2023 & 130 & ~? & - & - & - & - \\
Vit 2023 & 131 & ~? & - & - & - & - \\
Vit 2023 & 132 & ~\textsc{unsat} & \textcolor{darkgray}{50.7} & - & - & - \\
Vit 2023 & 133 & ~\textsc{unsat} & \textcolor{darkgray}{18.4} & - & - & - \\
Vit 2023 & 134 & ~\textsc{unsat} & \textcolor{darkgray}{18.5} & - & - & - \\
Vit 2023 & 135 & ~\textsc{unsat} & \textcolor{darkgray}{98.2} & - & - & - \\
Vit 2023 & 136 & ~? & - & - & - & - \\
Vit 2023 & 137 & ~? & - & - & - & - \\
Vit 2023 & 138 & ~\textsc{unsat} & \textcolor{darkgray}{47.3} & - & - & - \\
Vit 2023 & 139 & ~? & - & - & - & - \\
Vit 2023 & 140 & ~? & - & - & - & - \\
Vit 2023 & 141 & ~\textsc{unsat} & \textcolor{darkgray}{17.1} & - & - & - \\
Vit 2023 & 142 & ~? & - & - & - & - \\
Vit 2023 & 143 & ~? & - & - & - & - \\
Vit 2023 & 144 & ~? & - & - & - & - \\
Vit 2023 & 145 & ~? & - & - & - & - \\
Vit 2023 & 146 & ~\textsc{unsat} & \textcolor{darkgray}{49.3} & - & - & - \\
Vit 2023 & 147 & ~\textsc{unsat} & \textcolor{darkgray}{58.0} & - & - & - \\
Vit 2023 & 148 & ~\textsc{unsat} & \textcolor{darkgray}{17.1} & - & - & - \\
Vit 2023 & 149 & ~\textsc{unsat} & \textcolor{darkgray}{53.6} & - & - & - \\
Vit 2023 & 150 & ~\textsc{unsat} & \textcolor{darkgray}{49.2} & - & - & - \\
Vit 2023 & 151 & ~\textsc{unsat} & \textcolor{darkgray}{45.7} & - & - & - \\
Vit 2023 & 152 & ~? & - & - & - & - \\
Vit 2023 & 153 & ~? & - & - & - & - \\
Vit 2023 & 154 & ~\textsc{unsat} & \textcolor{darkgray}{70.5} & - & - & - \\
Vit 2023 & 155 & ~? & - & - & - & - \\
Vit 2023 & 156 & ~? & - & - & - & - \\
Vit 2023 & 157 & ~\textsc{unsat} & \textcolor{darkgray}{49.2} & - & - & - \\
Vit 2023 & 158 & ~? & - & - & - & - \\
Vit 2023 & 159 & ~? & - & - & - & - \\
Vit 2023 & 160 & ~? & - & - & - & - \\
Vit 2023 & 161 & ~? & - & - & - & - \\
Vit 2023 & 162 & ~? & - & - & - & - \\
Vit 2023 & 163 & ~\textsc{unsat} & \textcolor{darkgray}{46.1} & - & - & - \\
Vit 2023 & 164 & ~\textsc{unsat} & \textcolor{darkgray}{17.1} & - & - & - \\
Vit 2023 & 165 & ~? & - & - & - & - \\
Vit 2023 & 166 & ~\textsc{unsat} & \textcolor{darkgray}{17.1} & - & - & - \\
Vit 2023 & 167 & ~? & - & - & - & - \\
Vit 2023 & 168 & ~? & - & - & - & - \\
Vit 2023 & 169 & ~? & - & - & - & - \\
Vit 2023 & 170 & ~? & - & - & - & - \\
Vit 2023 & 171 & ~? & - & - & - & - \\
Vit 2023 & 172 & ~? & - & - & - & - \\
Vit 2023 & 173 & ~\textsc{unsat} & \textcolor{darkgray}{22.4} & - & - & - \\
Vit 2023 & 174 & ~? & - & - & - & - \\
Vit 2023 & 175 & ~\textsc{unsat} & \textcolor{darkgray}{18.4} & - & - & - \\
Vit 2023 & 176 & ~\textsc{unsat} & \textcolor{darkgray}{18.5} & - & - & - \\
Vit 2023 & 177 & ~? & - & - & - & - \\
Vit 2023 & 178 & ~\textsc{unsat} & \textcolor{darkgray}{18.5} & - & - & - \\
Vit 2023 & 179 & ~? & - & - & - & - \\
Vit 2023 & 180 & ~\textsc{unsat} & \textcolor{darkgray}{18.4} & - & - & - \\
Vit 2023 & 181 & ~? & - & - & - & - \\
Vit 2023 & 182 & ~\textsc{unsat} & \textcolor{darkgray}{45.6} & - & - & - \\
Vit 2023 & 183 & ~\textsc{unsat} & \textcolor{darkgray}{18.4} & - & - & - \\
Vit 2023 & 184 & ~\textsc{unsat} & \textcolor{darkgray}{56.8} & - & - & - \\
Vit 2023 & 185 & ~\textsc{unsat} & \textcolor{darkgray}{17.1} & - & - & - \\
Vit 2023 & 186 & ~? & - & - & - & - \\
Vit 2023 & 187 & ~? & - & - & - & - \\
Vit 2023 & 188 & ~? & - & - & - & - \\
Vit 2023 & 189 & ~\textsc{unsat} & \textcolor{darkgray}{23.7} & - & - & - \\
Vit 2023 & 190 & ~\textsc{unsat} & \textcolor{darkgray}{58.0} & - & - & - \\
Vit 2023 & 191 & ~\textsc{unsat} & \textcolor{darkgray}{52.2} & - & - & - \\
Vit 2023 & 192 & ~\textsc{unsat} & \textcolor{darkgray}{18.4} & - & - & - \\
Vit 2023 & 193 & ~? & - & - & - & - \\
Vit 2023 & 194 & ~\textsc{unsat} & \textcolor{darkgray}{55.3} & - & - & - \\
Vit 2023 & 195 & ~? & - & - & - & - \\
Vit 2023 & 196 & ~? & - & - & - & - \\
Vit 2023 & 197 & ~\textsc{unsat} & \textcolor{darkgray}{17.1} & - & - & - \\
Vit 2023 & 198 & ~? & - & - & - & - \\
Vit 2023 & 199 & ~\textsc{unsat} & \textcolor{darkgray}{45.9} & - & - & - \\
\midrule
Yolo 2023 & 0 & ~\textsc{unsat} & \textcolor{darkgray}{9.83} & \textcolor{darkgray}{14.3} & - & - \\
Yolo 2023 & 1 & ~\textsc{unsat} & \textcolor{darkgray}{9.81} & \textcolor{darkgray}{13.3} & - & - \\
Yolo 2023 & 2 & ~? & - & - & - & - \\
Yolo 2023 & 3 & ~\textsc{unsat} & \textcolor{darkgray}{9.82} & \textcolor{darkgray}{13.8} & - & - \\
Yolo 2023 & 4 & ~\textsc{unsat} & \textcolor{darkgray}{9.81} & \textcolor{darkgray}{13.8} & - & - \\
Yolo 2023 & 5 & ~\textsc{unsat} & \textcolor{darkgray}{13.7} & - & - & - \\
Yolo 2023 & 6 & ~\textsc{unsat} & \textcolor{darkgray}{9.83} & \textcolor{darkgray}{14.0} & - & - \\
Yolo 2023 & 7 & ~\textsc{unsat} & \textcolor{darkgray}{9.84} & - & - & - \\
Yolo 2023 & 8 & ~\textsc{unsat} & \textcolor{darkgray}{9.84} & \textcolor{darkgray}{13.9} & - & - \\
Yolo 2023 & 9 & ~\textsc{unsat} & \textcolor{darkgray}{9.82} & - & - & - \\
Yolo 2023 & 10 & ~\textsc{unsat} & \textcolor{darkgray}{9.86} & - & - & - \\
Yolo 2023 & 11 & ~\textsc{unsat} & \textcolor{darkgray}{9.84} & \textcolor{darkgray}{274} & - & - \\
Yolo 2023 & 12 & ~\textsc{unsat} & \textcolor{darkgray}{9.86} & - & - & - \\
Yolo 2023 & 13 & ~\textsc{unsat} & \textcolor{darkgray}{9.83} & \textcolor{darkgray}{13.2} & - & - \\
Yolo 2023 & 14 & ~\textsc{unsat} & \textcolor{darkgray}{16.2} & - & - & - \\
Yolo 2023 & 15 & ~\textsc{unsat} & \textcolor{darkgray}{9.83} & - & - & - \\
Yolo 2023 & 16 & ~\textsc{unsat} & \textcolor{darkgray}{14.0} & - & - & - \\
Yolo 2023 & 17 & ~? & - & - & - & - \\
Yolo 2023 & 18 & ~? & - & - & - & - \\
Yolo 2023 & 19 & ~\textsc{unsat} & \textcolor{darkgray}{9.84} & - & - & - \\
Yolo 2023 & 20 & ~\textsc{unsat} & \textcolor{darkgray}{9.84} & \textcolor{darkgray}{13.6} & - & - \\
Yolo 2023 & 21 & ~? & - & - & - & - \\
Yolo 2023 & 22 & ~\textsc{unsat} & \textcolor{darkgray}{9.83} & \textcolor{darkgray}{13.5} & - & - \\
Yolo 2023 & 23 & ~\textsc{unsat} & \textcolor{darkgray}{9.80} & - & - & - \\
Yolo 2023 & 24 & ~\textsc{unsat} & \textcolor{darkgray}{9.80} & \textcolor{darkgray}{14.3} & - & - \\
Yolo 2023 & 25 & ~? & - & - & - & - \\
Yolo 2023 & 26 & ~? & - & - & - & - \\
Yolo 2023 & 27 & ~? & - & - & - & - \\
Yolo 2023 & 28 & ~\textsc{unsat} & \textcolor{darkgray}{9.82} & \textcolor{darkgray}{14.1} & - & - \\
Yolo 2023 & 29 & ~\textsc{unsat} & \textcolor{darkgray}{9.82} & \textcolor{darkgray}{13.7} & - & - \\
Yolo 2023 & 30 & ~\textsc{unsat} & \textcolor{darkgray}{9.89} & \textcolor{darkgray}{13.6} & - & - \\
Yolo 2023 & 31 & ~\textsc{unsat} & \textcolor{darkgray}{9.85} & \textcolor{darkgray}{13.2} & - & - \\
Yolo 2023 & 32 & ~\textsc{unsat} & \textcolor{darkgray}{13.8} & - & - & - \\
Yolo 2023 & 33 & ~? & - & - & - & - \\
Yolo 2023 & 34 & ~\textsc{unsat} & \textcolor{darkgray}{29.4} & - & - & - \\
Yolo 2023 & 35 & ~\textsc{unsat} & \textcolor{darkgray}{9.82} & \textcolor{darkgray}{13.0} & - & - \\
Yolo 2023 & 36 & ~\textsc{unsat} & \textcolor{darkgray}{9.86} & \textcolor{darkgray}{14.1} & - & - \\
Yolo 2023 & 37 & ~\textsc{unsat} & \textcolor{darkgray}{9.85} & \textcolor{darkgray}{13.2} & - & - \\
Yolo 2023 & 38 & ~\textsc{unsat} & \textcolor{darkgray}{9.84} & \textcolor{darkgray}{13.4} & - & - \\
Yolo 2023 & 39 & ~\textsc{unsat} & \textcolor{darkgray}{9.84} & - & - & - \\
Yolo 2023 & 40 & ~\textsc{unsat} & \textcolor{darkgray}{9.83} & \textcolor{darkgray}{13.2} & - & - \\
Yolo 2023 & 41 & ~\textsc{unsat} & \textcolor{darkgray}{43.5} & - & - & - \\
Yolo 2023 & 42 & ~\textsc{unsat} & \textcolor{darkgray}{9.83} & \textcolor{darkgray}{13.9} & - & - \\
Yolo 2023 & 43 & ~\textsc{unsat} & \textcolor{darkgray}{9.80} & \textcolor{darkgray}{13.2} & - & - \\
Yolo 2023 & 44 & ~\textsc{unsat} & \textcolor{darkgray}{9.88} & \textcolor{darkgray}{12.7} & - & - \\
Yolo 2023 & 45 & ~\textsc{unsat} & \textcolor{darkgray}{9.84} & \textcolor{darkgray}{12.6} & - & - \\
Yolo 2023 & 46 & ~\textsc{unsat} & \textcolor{darkgray}{9.82} & \textcolor{darkgray}{13.2} & - & - \\
Yolo 2023 & 47 & ~\textsc{unsat} & \textcolor{darkgray}{9.82} & \textcolor{darkgray}{13.0} & - & - \\
Yolo 2023 & 48 & ~\textsc{unsat} & \textcolor{darkgray}{9.85} & \textcolor{darkgray}{13.2} & - & - \\
Yolo 2023 & 49 & ~\textsc{unsat} & \textcolor{darkgray}{9.82} & \textcolor{darkgray}{13.5} & - & - \\
Yolo 2023 & 50 & ~\textsc{unsat} & \textcolor{darkgray}{9.83} & \textcolor{darkgray}{14.0} & - & - \\
Yolo 2023 & 51 & ~? & - & - & - & - \\
Yolo 2023 & 52 & ~\textsc{unsat} & \textcolor{darkgray}{9.81} & \textcolor{darkgray}{14.1} & - & - \\
Yolo 2023 & 53 & ~\textsc{unsat} & \textcolor{darkgray}{9.88} & \textcolor{darkgray}{13.7} & - & - \\
Yolo 2023 & 54 & ~\textsc{unsat} & \textcolor{darkgray}{9.85} & \textcolor{darkgray}{13.2} & - & - \\
Yolo 2023 & 55 & ~\textsc{unsat} & \textcolor{darkgray}{9.85} & \textcolor{darkgray}{13.7} & - & - \\
Yolo 2023 & 56 & ~\textsc{unsat} & \textcolor{darkgray}{9.82} & \textcolor{darkgray}{13.5} & - & - \\
Yolo 2023 & 57 & ~\textsc{unsat} & \textcolor{darkgray}{9.82} & \textcolor{darkgray}{13.8} & - & - \\
Yolo 2023 & 58 & ~\textsc{unsat} & \textcolor{darkgray}{9.84} & \textcolor{darkgray}{13.1} & - & - \\
Yolo 2023 & 59 & ~\textsc{unsat} & \textcolor{darkgray}{9.90} & \textcolor{darkgray}{13.1} & - & - \\
Yolo 2023 & 60 & ~\textsc{unsat} & \textcolor{darkgray}{9.86} & \textcolor{darkgray}{14.2} & - & - \\
Yolo 2023 & 61 & ~\textsc{unsat} & \textcolor{darkgray}{9.82} & - & - & - \\
Yolo 2023 & 62 & ~\textsc{unsat} & \textcolor{darkgray}{9.87} & - & - & - \\
Yolo 2023 & 63 & ~\textsc{unsat} & \textcolor{darkgray}{9.82} & - & - & - \\
Yolo 2023 & 64 & ~? & - & - & - & - \\
Yolo 2023 & 65 & ~\textsc{unsat} & \textcolor{darkgray}{9.83} & \textcolor{darkgray}{13.9} & - & - \\
Yolo 2023 & 66 & ~\textsc{unsat} & \textcolor{darkgray}{9.82} & \textcolor{darkgray}{13.9} & - & - \\
Yolo 2023 & 67 & ~? & - & - & - & - \\
Yolo 2023 & 68 & ~\textsc{unsat} & \textcolor{darkgray}{9.83} & \textcolor{darkgray}{13.8} & - & - \\
Yolo 2023 & 69 & ~? & - & - & - & - \\
Yolo 2023 & 70 & ~\textsc{unsat} & \textcolor{darkgray}{9.85} & \textcolor{darkgray}{12.9} & - & - \\
Yolo 2023 & 71 & ~\textsc{unsat} & \textcolor{darkgray}{9.83} & \textcolor{darkgray}{13.5} & - & - \\
\bottomrule
\end{longtable}
}
\end{center}


\end{document}